\documentclass[mnsc,nonblindrev]{informs3} 

\OneAndAHalfSpacedXI

\usepackage{natbib}
 \bibpunct[, ]{(}{)}{,}{a}{}{,}%
 \def\bibfont{\small}%
\TheoremsNumberedThrough     
\ECRepeatTheorems

\EquationsNumberedThrough    

\MANUSCRIPTNO{}

\usepackage{hyperref}
\usepackage{url}

\usepackage{algorithmic}
\usepackage{color,xcolor}

\usepackage[mathscr]{eucal}
\usepackage{amsbsy}

\usepackage[linesnumbered,ruled]{algorithm2e}

\usepackage{fancyhdr}

\usepackage{microtype}

\usepackage{amsmath,amssymb,amsfonts}

\usepackage[inline]{enumitem}

\usepackage{comment}

\usepackage{xifthen}

\usepackage[noabbrev,nameinlink]{cleveref}
\crefname{ineq}{inequality}{inequalities}
\creflabelformat{ineq}{#2{\upshape(#1)}#3} 

\usepackage{subcaption}

\DontPrintSemicolon

\usepackage{basic_latex_macros}

\usepackage{your_macros}

\usepackage{bbm,amssymb}

\usepackage{amsmath,amsfonts,bm}

\def\eqref#1{(\ref{#1})}

\def\floor#1{\lfloor #1 \rfloor}
\def\1{\bm{1}}

\newcommand{\kernel}{\mathscr{K}}
\newcommand{\cx}{\mathbbm{x}}

\DeclareMathAlphabet{\mathsfit}{\encodingdefault}{\sfdefault}{m}{sl}
\SetMathAlphabet{\mathsfit}{bold}{\encodingdefault}{\sfdefault}{bx}{n}

\newcommand{\otil}{\ensuremath{\tilde{\order}}}
\newcommand{\order}{\ensuremath{\mathcal{O}}}
\newcommand{\tracer}[2]{\ensuremath{\langle \!\langle {#1}, \; {#2}
    \rangle \!\rangle}}

\newcommand{\bThetaStar}{\ensuremath{\bTheta^*}}

\newcommand{\price}{\ensuremath{p}}

\newcommand{\tinit}{\ensuremath{\smallsub{t}{init}}}

\newcommand{\Jevent}{\ensuremath{\mathcal{J}}}
\newcommand{\JeventTil}{\ensuremath{\widetilde{\Jevent}}}

\newcommand{\Gevent}{\ensuremath{\mathcal{G}}}
\newcommand{\Oevent}{\ensuremath{\mathcal{U}}}
\newcommand{\cOevent}{\ensuremath{\check{\mathcal{U}}}}
\newcommand{\cU}{\check{U}}

\usepackage{scalerel}

\newcommand{\Smat}{\ensuremath{\mathbf{S}}}
\newcommand{\tSmat}{\ensuremath{\mathbf{\tilde{S}}}}
\newcommand{\tvarepsilon}{\tilde{\varepsilon}}
\newcommand{\tbx}{\mathbf{\tilde{{{\boldsymbol{x}}}}}}
\newcommand{\tbdelta} {\tilde{{\boldsymbol{\delta}}}}
\newcommand{\cbx}{\mathbf{\check{{{\boldsymbol{x}}}}}}
\newcommand{\cbdelta} {\check{{\boldsymbol{\delta}}}}

\newcommand{\dterm}{\ensuremath{\rho}}
\newcommand{\ErrTerm}{\ensuremath{e}}
\newcommand{\tErrTerm}{\tilde{\ensuremath{e}}}
\newcommand{\tdterm}{\tilde{\ensuremath{\rho}}}
\newcommand{\cdterm}{\check{\ensuremath{\rho}}}
\newcommand{\interterm}{\mathfrak{D_1}}
\newcommand{\fluctterm}{\mathfrak{D_2}}
\newcommand{\tinterterm}{\tilde{\mathfrak{D}}_1}

\newcommand{\cfluctterm}{\check{\mathfrak{D}}_2}
\newcommand{\Nc}{\mathbf{N}_{\eta}}
\newcommand{\CoveringSet}{\mathbbm{F}}
\newcommand{\tdTheta}{\tilde{\boldsymbol{\Delta}}}
\newcommand{\rat}{\mathfrak{S}}
\newcommand{\bzeta}{\boldsymbol{\zeta}}
\newcommand{\dd}{\mathrm{d}}
\newcommand{\simplifykappa}[1]{\frac{10(h+h^2)}{{#1}^{2/3}} \Big( (d_a+d_x)(1+\frac{1}{2}\log{\log{#1}}) + 6\log{#1} + 3\log{L} \Big)^2 \log{#1} }

\newcommand{\figdir}{figs}

\newcommand{\bThetaTrue}{\ensuremath{\bTheta^*}}
\newcommand{\bThetaHat}{\ensuremath{\widehat{\bTheta}}}

\newcommand{\preone}[1]{\ensuremath{\phi_1(#1)}}
\newcommand{\pretwo}[1]{\ensuremath{\phi_2(#1)}}

\makeatletter

\def\underbracex#1#2{\mathop{\vtop{\m@th\ialign{##\crcr
   $\hfil\displaystyle{#2}\hfil$\crcr
   \noalign{\kern3\p@\nointerlineskip}%
   #1\crcr\noalign{\kern3\p@}}}}\limits}

\def\underbracea{\underbracex\upbracefilla}

\def\upbracefilla{$\m@th \setbox\z@\hbox{$\braceld$}%
  \bracelu\leaders\vrule \@height\ht\z@ \@depth\z@\hfill 
\kern\p@\vrule \@width\p@\kern\p@\vrule \@width\p@\kern\p@\vrule \@width\p@
$}

\def\upbracefillb{$\m@th \setbox\z@\hbox{$\braceld$}%
\vrule \@width\p@\kern\p@\vrule \@width\p@\kern\p@\vrule \@width\p@\kern\p@
 \leaders\vrule \@height\ht\z@ \@depth\z@\hfill\bracerd
  \braceld\leaders\vrule \@height\ht\z@ \@depth\z@\hfill
\kern\p@\vrule \@width\p@\kern\p@\vrule \@width\p@\kern\p@\vrule \@width\p@
$}

\def\upbracefillc{$\m@th \setbox\z@\hbox{$\braceld$}%
\vrule \@width\p@\kern\p@\vrule \@width\p@\kern\p@\vrule \@width\p@\kern\p@
\leaders\vrule \@height\ht\z@ \@depth\z@\hfill
\kern\p@\vrule \@width\p@\kern\p@\vrule \@width\p@\kern\p@\vrule \@width\p@
$}

\def\upbracefilld{$\m@th \setbox\z@\hbox{$\braceld$}%
\vrule \@width\p@\kern\p@\vrule \@width\p@\kern\p@\vrule \@width\p@\kern\p@
 \leaders\vrule \@height\ht\z@ \@depth\z@\hfill\braceru$}

\def\underbracebd{\underbracex\upbracefillbd}
\def\upbracefillbd{$\m@th \setbox\z@\hbox{$\braceld$}%
\vrule \@width\p@\kern\p@\vrule \@width\p@\kern\p@\vrule \@width\p@\kern\p@
\bracerd\braceld
 \leaders\vrule \@height\ht\z@ \@depth\z@\hfill\braceru$}
\makeatother

\begin{document}



\RUNAUTHOR{Cai, Chen, Wainwright, and Zhao}

\RUNTITLE{Doubly High-Dimensional Contextual Bandits}

\TITLE{Doubly High-Dimensional Contextual Bandits: \\ An Interpretable
  Model for Joint Assortment-Pricing}

\ARTICLEAUTHORS{%
\AUTHOR{Junhui Cai}
\AFF{Department of Information Technology, Analytics, and Operations,
University of Notre Dame,
\EMAIL{jcai2@nd.edu}}
\AUTHOR{Ran Chen}
\AFF{Laboratory for Information and Decision Systems, 
Massachusetts Institute of Technology,
\EMAIL{ran1chen@mit.edu}}
\AUTHOR{Martin J. Wainwright}
\AFF{Laboratory for Information and Decision Systems, Statistics and Data Science Center, \\
EECS \& Mathematics,
Massachusetts Institute of Technology,
\EMAIL{wainwrigwork@gmail.com}}
\AUTHOR{Linda Zhao}
\AFF{Department of Statistics and Data Science,
University of Pennsylvania,
\EMAIL{lzhao@wharton.upenn.edu}}
} 

\ABSTRACT{Key challenges in running a retail business include how to
  select products to present to consumers (the assortment problem),
  and how to price products (the pricing problem) to maximize revenue
  or profit.  Instead of considering these problems in isolation, we
  propose a joint approach to assortment-pricing based on contextual
  bandits. Our model is doubly high-dimensional, in that both context
  vectors and actions are allowed to take values in high-dimensional
  spaces.  In order to circumvent the curse of dimensionality, we
  propose a simple yet flexible model that captures the interactions
  between covariates and actions via a (near) low-rank representation
  matrix.  The resulting class of models is reasonably expressive
  while remaining interpretable through latent factors, and includes
  various structured linear bandit and pricing models as particular
  cases.  We propose a computationally tractable procedure that
  combines an exploration/exploitation protocol with an efficient
  low-rank matrix estimator, and we prove bounds on its regret.
  Simulation results show that this method has lower regret than
  state-of-the-art methods applied to various standard bandit and
  pricing models.  
  Real-world case studies on the assortment-pricing problem, from an industry-leading instant noodles company to an emerging beauty start-up, underscore the gains achievable using our
  method. In each case, we show at least three-fold gains in
  revenue or profit by our bandit method, as well as the
  interpretability of the latent factor models that are learned.
  }


\maketitle

\pagebreak


\section{Introduction}
\label{Sec-Intro}


In the modern business and healthcare landscape, it is now \emph{status quo} to make
use of online decision-making algorithms that incorporate individual
characteristics as well as micro- and macro-economic factors. For
example, on-line retailers determine product offerings and pricing
based on customer demographics, browsing, and purchasing history;
business managers allocate resources, such as staff and equipment, 
based on current operational conditions; and medical providers prescribe treatment 
and therapy combinations based on the patient's medical records.

In these settings, bandit algorithms are often deployed to learn the
reward structure while optimizing performance by strategically
``exploring'' and ``exploiting'' potential actions.  In order to make
optimal decisions, the decision-maker must take into account two
factors: the reward (e.g., revenue or profit) depends on both a space
of possible actions, as well as individual features or covariates,
also known as the context.  Many bandit algorithms are limited to
finite or relatively low-dimensional action and context spaces, but in
practice, both can be high-dimensional in nature.  For instance, the
action vector for an online retailer may include pricing and
assortment information for thousands of products.  Thus, we are led to
consider the following question: can we develop useful models and
efficient learning procedures for contextual bandits that are
high-dimensional in both actions and covariates?  Providing one
affirmative answer to this open question---and demonstrating the
utility of the resulting model and algorithms for two real-world
motivating case studies---are the primary contributions of our work.

\subsection{Background and Our Approach}
\label{Subsec-background}

The primary application that motivates our work is dynamic assortment
and pricing.  It is a central challenge for on-line retailers, and
using bandits for this problem is natural given the sequential nature
of the decision-making.  The assortment problem amounts to choosing
the selection of products to be offered while satisfying capacity
constraints, whereas the pricing problem is to set selling prices for
these products.  Both assortment and pricing decisions share a common
goal: maximizing a specific objective function, such as revenue or
profit.  Although both dynamic assortment optimization and pricing
problems have been separately studied extensively in the literature,
the joint assortment-pricing problem has received comparatively less
attention.

The key to a successful assortment and pricing strategy lies in
understanding market response to the assortment-pricing decisions.  A major challenge in
modern assortment-pricing is the explosion in dimensionality of both
the action and covariate spaces.  Companies typically consider large
numbers ($\gg 100$) of products simultaneously.  From the universe of
products, they consider large collections of possible product subsets
to display, and the associated price takes values in continuous space.
Thus, the action space becomes high-dimensional with a mixture of
discrete and continuous elements.  The problem is further complicated
by the high-dimensional covariates: fueled by the rise of e-commerce,
it is possible to measure many customer-specific or industry-specific
features that can be relevant to modeling demand and price
sensitivity.  As the action-covariate dimensions grow, without some
kind of structure, there are ``no-free-lunch'' theorems showing that
it is prohibitively costly, both in terms of samples and computation,
to learn an optimal policy~\citep{lattimore2020bandit}.  Thus, it
becomes essential to develop models with ``low-dimensional structure''
that explain important features of the data, while being amenable to
statistically and computationally efficient algorithms.

A fortunate fact---and the starting point for our modeling---is that
there often exist low-dimensional latent factors that explain the bulk
of the reward structure.  In the retail context, the demands for
products, one deciding quantity for the reward, that share similar 
features/attributes are influenced in common ways by underlying market
features. And usually only a handful of the underlying product factors matter. For instance, 
there exists ``color psychology'' in marketing \citep{singh2006impact} 
and customers' color preference in basic colors such as white, black, blue, and red \citep{madden2000managing}.
Similarly, the covariate
vectors relevant for assortment-pricing can be explained by a few
latent factors.  For instance, at the individual level, much of the
variance in consumer buying power can be captured by a mixture of
demographic (e.g., income, education level) and geographic traits~(see \cite{pol1991demographic} and references therein);
at the macro level, population purchasing preference, usually
indicated by season, region, and other macroeconomic indices,
significantly impact the overall demand \citep{estelami2001macro,
  gordon2013does, kumar2014assessing}.  As a result, the interaction
effects between the action and covariates---a major source
contributing to revenue generation---can be characterized by a few
latent factors.  For example, customers with lower buying power tend
be more conservative in their buying behavior
(e.g.,~\citep{wakefield2003situational}).

In summary, the low-dimensional structure, therefore, captures the
essence of the effect of actions and covariates on the reward function
and often aligns with intuitive or interpretable factors.  Accounting
for the common latent factors further speeds up the reward learning
per product and low-dimensional models often improve computational
efficiency.  The interpretability and computational efficiency using
latent factors turn the ``curse of dimensionality'' into a ``blessing
of dimensionality''~\citep{li2018embracing}.

With these insights, we tackle the joint assortment-pricing problem by
casting it as a doubly high-dimensional bandit problem and propose a
new model that captures interactions between the high-dimensional
actions and covariates via an (approximately) low-rank matrix
representation.  Our goal is to offer a sequence of assortment and
pricing decisions, which can be represented as a sequence of action
vectors $\{ \ba_t \}_{t=1}^T$ that take values in (some subset of)
$\Real^{d_a}$, under the contexts, which can be represented as a
sequence of covariate vectors $\{ \bx_t \}_{t=1}^T$ taking values in
$\Real^{d_x}$, such that the cumulative expected revenue over the time
horizon $T$ is maximized.  Since both the action dimension $d_a$ and
the covariate dimension $d_x$ can be large, our proposed model uses a
low-rank matrix to take advantage of the low-dimensional latent
factors. Specifically, our reward model takes the bilinear form: given
an action vector $\ba \in \mathbb{R}^{d_a}$ and a covariate vector
$\bx \in \mathbb{R}^{d_x}$, we observe a noisy reward $\Reward$ with
conditional mean
\begin{align*}
  \Exs[\Reward \mid \bx, \ba] = \ba^T \bThetaTrue \bx,
\end{align*}
where $\bThetaTrue\in\mathbb{R}^{d_a \times d_x}$ is an unknown
representation matrix that is relatively low-rank---say with rank
$\myrank \ll \min\{d_a, d_x\}$---or more generally, well-approximated
by a matrix with low rank.

The representation matrix $\bThetaTrue$ captures important interactions
between actions and covariate pairs via its spectral structure.
Performing a singular value decomposition (SVD) on the representation
matrix yields the latent structure, with the left (respectively right)
singular vectors corresponding to the action (respectively covariate)
space structure.  In this way, our model implicitly performs a form of
dimension reduction in how the actions and covariates interact to
determine the reward function.

Given this structure, we also propose a new algorithm (\hicoab) that
combines low-rank estimation with an exploration/exploitation
strategy.  It is a computationally efficient approach, involving only
the solution of simple convex programs in all phases.  We prove
non-asymptotic bound on its expected regret, one that shows that it is
also statistically efficient in terms of problem dimension and
structure.  We also show our method not only can solve the joint
assortment-pricing problem, but is actually general enough to
encompass various bandit models as special cases.

\subsection{Main Contributions}
\label{Subsec-results}

Let us summarize some of our main contributions:

\begin{enumerate}[leftmargin=*, itemsep=0ex, topsep=0pt]
    \item \textbf{A general and interpretable model for joint
      assortment-pricing:}
We propose a new model for doubly high-dimensional contextual bandits,
in which both covariates and actions can be high-dimensional and
continuous, by leveraging the low-dimensional latent factors using a
low-rank representation matrix.  We then adopt our new model to tackle
the joint dynamic assortment-pricing problem, which allows for
feature-dependent and context-specific demand heterogeneity, while
previous literature mostly studies the assortment and pricing problems
separately. In addition, our model can account for new products as opposed to
a predetermined set of available products in most existing assortment models.

As we argue, an advantage of this low-rank model is its combination of
a high degree of interpretability with prediction power.  The low-rank
matrix encapsulates the interaction between action-covariate pairs via
its singular vectors, providing a form of dimension reduction and
interpretability.  On the other hand, given the covariate, our model
is able to predict the reward of an unseen arm.  Both interpretability
and predictive power can be tremendously useful for decision-makers.

Our model is general. It unifies a number of structured bandit and
pricing models studied in past work; it can capture complex
relationships between variables; and it is applicable to an array of
applications involving multiple decision-making.

\item \textbf{An computationally efficient and adaptive algorithm.}
  We propose an efficient algorithm for online learning, referred to
  as the \textbf{Hi}gh-dimensional \textbf{C}ontextual and
  \textbf{Hi}gh-dimensional \textbf{C}ontinumm \textbf{A}rmed
  \textbf{B}andit (\hicoab).  It interleaves estimation steps, in which
  the low-rank representation matrix is estimated based on data
  observed thus far, with exploration/exploitation steps, in which new
  actions are selected.  Unlike explore-then-commit type algorithms,
  \hicoab~is adaptive with respect to the time horizon $T$: in
  particular, there are no $T$-dependent tuning parameters (e.g.,
  exploration period). Moreover, it updates the representation matrix
  across the entire time horizon, making it more suitable for online
  learning. On a separate note, it is also adaptive to the rank $r$ (i.e., the number of latent factors) in that it does not require prior knowledge of $r$.

\item \textbf{A non-asymptotic and instance-dependent upper bound.}
  We measure the performance of our algorithm using the standard
  notion of expected regret, which is the average expected deficit in
  reward achieved by \hicoab~compared with an oracle that knows the
  low-rank representation matrix.  We provide a non-asymptotic upper
  bound on the expected regret of \hicoab.  The technical challenge is
  that samples are not i.i.d because the bandit protocol during the
  exploitation step adaptively collects samples, making classic matrix
  theory results for i.i.d. data inapplicable.  We overcome this
  challenge by proving a new tail bound for the low-rank matrix
  estimator by carefully leveraging empirical process and martingales
  concentration results, and thereby a non-asymptotic upper bound on
  the expected regret.  We further note that the non-asymptotic bounds
  hold for all $T$ while the algorithm does not require prior
  knowledge of $T$. This adaptivity is of both theoretical interest
  and practical importance.
\item \textbf{Take-away insights for assortment-pricing practice.}  We
  compare the performance of \hicoab~against existing algorithms under
  various standard bandit and pricing models and applications to
  real-world retail problems. Simulations demonstrate that
  \hicoab~outperforms state-of-the-art methods in expected regret.  We
  further apply \hicoab~to two real-world retail problems involving
  joint assortment and pricing, one for a leading instant noodle
  producer and the other for a manicure start-up.  The results
  demonstrate the effectiveness of this joint approach for revenue
  maximization.  Both case studies involve a large number of products
  and covariates and no existing methods are applicable.  \hicoab~is
  shown to be capable of managing such doubly high dimensionalities
  and providing simultaneous assortment and pricing decisions. The
  assortment-pricing policy based on \hicoab~yields sales almost
  four times as high as the strategies in practice. Moreover, our
  model reveals insights for assortment and pricing such as the
  popularity of flavor (noodles) or color (manicure) under different
  contexts such as locations and seasons.  Finally, our model is able
  to predict the revenue of a new product, which can guide new product
  designs.
\end{enumerate}


\subsection{Related Literature}
\label{Subsec-lit}

So as to situate our work more broadly, let discuss and summarize some
related literature.

\paragraph{Dynamic Assortment and Pricing.}

In the field of operations research and revenue management, assortment
and pricing are key decisions to be made by any firm; accordingly,
there is a substantial body of past work on dynamic assortment and
dynamic pricing.

Beginning with dynamic assortment, \cite{caro2007dynamic} was an early
approach to formulate it as a multi-armed bandit problem, but assuming
independent demand for each product. A popular alternative demand
model is the multinomial logit (MNL) choice model, which uses a
logistic model to estimate demand
parameters~\citep{rusmevichientong2010dynamic, saure2013optimal}; more
recent work has adapted multi-arm bandit techniques to the MNL
model~\citep{chen2017note, agrawal2019mnl, chen2021dynamic, chen2023robust}.  The MNL
model can be further extended for personalization dynamic assortment
by integrating personal information~\citep{cheung2017thompson,
  chen2020dynamic, miao2022online}.  Another MNL variant accounts for
heterogeneity via customer segmentation~\citep{bernstein2019dynamic,
  kallus2020dynamic}.  In particular, \cite{kallus2020dynamic} also adopt a low-rank matrix to model
the interaction between product and customer types, but they only
consider finite types of products and customers and does not account for product/customer attributes.
Apart from estimating the optimal set, recently \cite{shen2023combinatorial} propose the first inferential framework for testing optimal assortment.
It is worth mentioning two limitations on the MNL model.
First, the unit of the MNL model is limited to one customer arrival, so does
not capture the dynamic nature of customer arrivals.  These dynamics
have significant impacts on decisions taken over extended periods like
weeks, months, and years.  Our model, on the other hand, directly
models the revenue itself and can be adopted at any granularity.  At
the customer level, we do not assume each customer only purchases one
of the products offered, which is usually not the case in
practice. 
Second, MNL models treat each product in isolation, ignoring any inherent product similarities. 
Our model leverages product attributes to capture product similarities.
As a result, our model can improve predictive power and provide deeper insights.
Moreover, it can further handle new products while the set of
available products is predetermined in the MNL models.

Dynamic pricing has been another important stream in revenue
management and the price-demand curve is often assumed to be
linear~\citep{kleinberg2003value, araman2009dynamic,
  besbes2009dynamic, broder2012dynamic, den2014simultaneously,
  keskin2014dynamic}; the paper by~\cite{den2015dynamic} provides a
helpful survey.  Recent work has turned towards dynamic pricing based
on customer characteristics (e.g.,~\cite{ban2021personalized,
  chen2021nonparametric, bastani2022meta}) and/or product features
(e.g.,~\cite{qiang2016dynamic, javanmard2019dynamic,
  cohen2020feature, miao2022context, fan2022policy}).  Much of the pricing literature focuses on the
single-product setting, while sellers usually need to price multiple
products simultaneously.  The MNL model has also been used in
multi-product pricing problem~\citep{akccay2010joint,
  gallego2014multiproduct} but they do not consider product feature or
customer characteristics.  The recent work of \cite{bastani2022meta}
proposes a meta-dynamic pricing algorithm, which uses an empirical
Bayes approach to learn the demand function across products by
assuming that the demand parameters of each product follow a common
unknown prior to leveraging the similarities in demand among related
products. However, their algorithm learns products sequentially, while
our algorithm provides pricing for multi-products simultaneously.

While dynamic assortment and pricing problems have been studied
extensively in isolation, research addressing the joint
assortment-pricing problem is relatively sparse.
\cite{chen2022statistical} engaged with this issue in an offline
setting.  More recently, \citet{miao2021dynamic} offer a solution
using the MNL choice model.  However, their approach is hampered by
the limitations inherent to the MNL model, namely that the time
horizon corresponds to the number of customers rather than actual time
units; products are assumed to be independent of one another; the
number of possible products cannot be large; and the model is unable
to propose new products. Furthermore, their model does not integrate
contextual information. In contrast, our work takes into account
product features and contexts using a new model that can address these
limitations.


\paragraph{Multi-Arm and Continuum Armed Contextual Bandits.}

Most online decision-making problems, including dynamic assortment and
pricing, can be modeled as particular instances of a bandit problem,
with the latter dating back to the seminal work
of~\citet{robbins1952some}.  At each round, a decision-maker chooses
an action (arm) and then observes a reward.  The goal is to act
strategically so as to determine a near-optimal policy without
incurring large regret.  There is now a very well-developed literature
on the bandit problem, and its extension to the contextual bandits; we
refer the reader to the comprehensive book
by~\citet{lattimore2020bandit} and references therein for more
background.

More recently, the literature on high-dimensional bandit problems has
been an active area; it exploits a relatively mature body of
statistical tools for high-dimensional problems (e.g., see the
book~\citep{ wainwright2019high} and references therein).  There is a
line of work on contextual bandits with high-dimensional covariates,
including the LASSO bandit problem~\citep{abbasi2012online,
  kim2019doubly, bastani2020online, hao2020high, papini2021leveraging,
  xu2021learning,chen2022nearly}, in which the mean reward is assumed to be a linear
function of a sparse unknown parameter vector.  As we describe in the
sequel, these high-dimensional bandit models are special cases of the
high-dimensional low-rank model studied in this paper.  Other work
exploit non-parametric methods---among them boosting, random forests,
or neural networks---to estimate the reward
function~\citep{feraud2016random, zhou2020neural, ban2021ee,
  chen2022interconnected, xu2022neural}.  Such approaches are quite
different in flavor from our model, and we compare to one such method
in our experimental results.

There are various other models and problems that have connections to
but differ from the setup in this paper.  For example, one line of
research focuses on representation learning in linear bandits,
specifically for low-rank bandit models and multi-task learning where
several bandits are played concurrently. The actions for each task are
embedded in the same space and share a common low-dimensional
representation~\citep{kveton2017stochastic, lale2019stochastic,
  yang2020impact, hu2021near, lu2021low, kang2022efficient}.  However,
this line of research does not consider contextual information, and
often imposes case-specific assumptions on the action space.  Among
such papers, \citet{kang2022efficient} study a trace inner product
bandit with a matrix of known (low) rank $r$, in which the action is
matrix-valued.

Our algorithm and theory, in contrast, are designed explicitly for
contextual problems, and we do not need to know the rank $r$ of the
target matrix.  Our reward model is connected to but different from
other papers that propose bilinear-type reward models
(e.g.,~\citet{jun2019bilinear,kim2021scheduling,rizk2021best}) in
which \emph{both} arguments of the bilinear function are part of the
action.  Such models can be understood as a structured linear bandit
of a particular type, and unlike our models, do not capture the
interaction between the covariate and action at each time step.

One class of models for continuum-action bandits takes the reward
function to be ``smooth'' over the action space, with Lipschitz or
H\"{o}lder smoothness (e.g.,~\cite{agrawal1995continuum,
  kleinberg2004nearly, Kleinberg2019}) being typical examples.
Researchers have taken different approaches to such models, including
reducing the problem to a finite action space via discretization, or
using non-parametric methods estimate the reward function; both
approaches lead to different procedures from those that we study.
Other work on contextual bandits with continuous states-action spaces
imposes Lipchitz-type conditions on the reward function jointly over
the action-covariate space~\citep{lu2010contextual,
  slivkins2011contextual, krishnamurthy2020contextual}; for these
reasons, it is limited to relatively low-dimensional settings.  There
is also other work on high-dimensional models for contextual
bandits~\citep{turugay2020exploiting}.
 Yet, these are rooted in different models, cater to different settings, 
 employ disparate techniques, and lack the interpretability inherent in our low-rank bilinear model.


\paragraph{Factor Models and Low-rank Matrix Estimation.}
Factor models and methods for low-rank matrix estimation and
prediction have been studied extensively in both statistics and
machine learning
(e.g.,~\citet{SreAloJaa05,RecFazPar10,candes2010matrix,
  negahban2011estimation,udell2016generalized,
  cai2018rate,chen2022statistical}) with a wide variety of practical
applications ranging from psychology \cite{hotelling1933analysis},
finance and economics~\citep{fan2021recent}, recommendation
system~\citep{bennett2007netflix}, and electronic health
records~\citep{schuler2016discovering}.  Our \hicoab~algorithm uses
least-squares with nuclear norm regularization, which is a well-known
approach (cf. Chapter 10 in~\citet{wainwright2019high});
however, our analysis requires a number of technical innovations to deal with the highly adaptive nature of bandit data collection.


\subsection{Notation}
\label{Subsec-notation}
We use bold lowercase for vectors and bold uppercase for matrices.  We
use $\|\ba\|_2$ to denote the $\ell_2$-norm of vector $\ba$.  For a
matrix $\bA$, we define its Frobenius norm $\fronorm{\bA} \defn
\sqrt{\sum_{i, j} a_{ij}^2}$; its $\ell_2$-spectral norm
$\opnorm{\bA} \defn \sup_{\|\bx\|_2 = 1} \|\bA
\bx\|_2$; and its nuclear norm $\mynucnorm{\bA} \defn \sum_{k=1}^d
\sigma_k(\bA)$ be its nuclear norm, where $d$ is the rank and
$\sigma_k(\bA)$ are the singular values of $\bA$. We use
$\inprod{\ba}{\bb} \defn \ba^\top \bb$ to denote the Euclidean inner
product between two vectors, and $\tracer{\bA}{\bB} \defn
\text{trace}(\bA^\top \bB)$ the trace inner product between two
matrices.  We use the standard notation $\order(\cdot)$ and
$\Omega(\cdot)$ to characterize the asymptotic growth rate of a
function.


\subsection{Outline}
\label{Subsec-outline}

The remainder of the paper is organized as follows.  We begin
in~\Cref{Sec-Formulation} by motivating our model with our real-world
case study of an online retailer that seeks to perform joint
assortment-pricing. Equipped with this motivation, we then formally
describe the doubly high-dimensional contextual bandit model.
\Cref{Sec-Algo} presents the \hicoab~algorithm for representation
learning and regret minimization, whereas~\Cref{Sec-Theory} provides a
non-asymptotic instance-dependent bound on the expected regret.
Finally, \Cref{Sec-Sim} describes a suite of empirical results on
simulated data and real-world case studies.  We compare the
performance of \hicoab~with other pricing and bandit algorithms as
well as two case studies on real sales data from one of the largest
online retailers and a start-up.  We conclude with a summary and
discussion of future research in~\Cref{Sec-Conclusion}.  Proofs and
additional empirical results are provided in the online appendices in the
supplemental material.


\section{Problem Motivation and Formulation}
\label{Sec-Formulation}

In this section, we begin by motivating the class of problems studied
in this paper with a concrete example.  We then provide a more precise
formulation of the problem.


\subsection{A Real-world Instance of a Doubly High-dimensional Bandit}
\label{Subsec-example}

Let us provide some concrete motivation for the model that we develop
by discussing the e-commerce assortment-pricing problem faced by a
market-leading producer of instant noodles in China.  This company has
a total of $176$ products in their portfolio, but can display no more
than $30$ at any time on their main website page.  Consequently, the
total number of product combinations available to them is ${176
  \choose 30} \approx 6.4\times 10^{33}$---an extremely large number!
In addition to this assortment decision, they need to decide on the
prices. All together, the vector characterizing their
possible actions at each time step is high-dimensional and involves a
mixture of both discrete and continuous values.

At the same time, they also have at their disposal a rich array of
contextual information, including macro-environmental information such
as season, location, and specific holidays.  In addition, in certain
cases, additional micro-level information is also available, such as
users' profile information and historical data.
Encoding this side information as a context vector also leads
to a high-dimensional state.

The company makes decisions in a sequential fashion, jointly choosing
the assortment and pricing in each round.  After doing so, they
observe the revenue/profit at the end of each period, which we refer
to as the reward.  The firm needs to learn the reward function with
respect to different assortment and pricing given the contextual
information on the fly and make the optimal assortment and pricing
decision that maximizes the cumulative reward across the time horizon.
This exploration-exploitation problem can be modeled as a bandit
problem and both the arm and contextual vectors take continuous values
in high-dimensional spaces.

Models with high-dimensional actions and covariate arise in many
applications of dynamic assortment-pricing.  As noted previously, some
past work focuses on settings where the numbers of products and slots
are relatively small, and utility functions of each product are taken
to be different and unrelated to each other. At the same time, without
imposing additional structure on high-dimensional bandit problems, one
cannot expect to obtain non-trivial guarantees (due to
``no-free-lunch'' theorems).  Accordingly, it is essential to impose
structure, and in this paper, we posit low-dimensional structure in
the form of a small number of latent factors that control interactions
between actions and covariates in determining the expected reward.

\subsection{Formalizing the Model}
\label{Subsec-model}

With this intuition in place, let us formalize the class of models
that we study in this paper.  We consider a firm that makes
assortment-pricing decisions over a period of $T$ rounds, indexed by
$t \in [T] \defn \{1, 2, \ldots, T \}$.  Each can be of any
predetermined granularity (e.g., by day, week, month, or by the
arrival of one customer).  There are a total of $K$ products to be
sold, indexed by $k \in [K] \defn \{1, \ldots, K\}$.

\subsubsection{Action and Context Vectors.}

At each time $t \in [T]$, the product with index $k \in [K]$ is
associated with an $m$-dimensional attribute vector $\tbp_{t,k} \in
\real^{m}$ along with a non-negative price $p_{t, k} \in [0,
  \infty)$.  Features encoded by the vector $\tbp_{t, k}$ depend on
  the product, but might include color, flavor, material, and
  technical specifications, etc.  Collecting together all the
  attribute vectors and prices across the $K$ products, we obtain the
  \emph{action vector} at time $t \in [T]$, given by
\begin{align}
\label{eq:action}
  \ba_t & \defn \big( \tbp_{t,1},p_{t,1},\,\tbp_{t,2}, p_{t,2}, \,
  \cdots,\,\tbp_{t,k}, p_{t,k}, \cdots,\,\tbp_{t,K},p_{t,K},1 \big),
\end{align}
where $(\tbp_{t,k}, p_{t,k})$ denotes the product feature and price
for slot $k$ at time $t$.  The special notation $(\tbp_{t,k} = \bm 0,
p_{t,k} = 0)$ indicates that the slot for product $k$ being empty.
Note that this action vector $\ba_t$ has $d_a \defn K \,(m+1) + 1$
components in total, and is thus high-dimensional for the values of
$(K, m)$ typical in practice.

For each period $t \in [T]$, the firm also observes side-information
in the form of a \emph{context vector} \mbox{$\bx_t \in \real^{d_x}$.}
This context vector $\bx_t$ can include individual or aggregated
customer information depending on the granularity or
macro-environmental factors and thus the dimension $d_x$ can be large.
We assume that $\bx_t$ is independent of the firm's decision prior to
$t$.

\subsubsection{Reward Structure.}
The goal of the firm is to make assortment-pricing decisions, via
their choice of the action vector $\ba_t$ at each time $t$, so as to
maximize revenue.  At time $t$, we model the revenue in terms of
its conditional expectation given a context-action pair $(\bx_t,
\ba_t)$.  In particular, we assume that the observed reward
$\Reward_t$ has a conditional mean function of the bilinear form
\begin{align}
\label{eq:reward-l}
  \Exs[\Reward \mid \bx_t, \ba_t] = \ba_t^T \bThetaTrue \bx_t,
\end{align}
where $\bThetaTrue \in \mathbb{R}^{d_a \times d_x}$ is an unknown
representation matrix.  The matrix $\bThetaTrue$ captures interactions
between the action vector $\ba_t$ and the covariate vector $\bx_t$ in
determining the expected reward.

To provide intuition, consider traditional assortment models where $K$
products are chosen from a total of $N$ predetermined available
products, each of ${N \choose K}$ possible assortments can be used to
define an action in our high-dimensional contextual bandit.  In
particular, suppose that the action vector $\ba_t \in \real^{d_a}$ is
a standard basis vector, where a single entry in position $j$
indicates the $j^{th}$ action to be taken, and the representation
\mbox{$\bThetaTrue=(\bbeta_1,\bbeta_2,\cdots,\bbeta_{N \choose K })^\top$}
and $\bbeta_i$ is the parameter vector corresponding to $i$-th
assortment.  In this case, each assortment is assumed to be different
and ignores the similarity between assortments and products.  On the
other hand, formulating the action vector by concatenating the product
attribute vectors (cf. equation~\eqref{eq:action}) can take advantage
of the similarity between different assortment as products with
similar features often have similar rewards.

Our model can also be generalized to a collection of $L \geq 2$ of
different targets---say different platforms or geographic locations.
Indexing the targets by $\ell \in [L] \defn \{1, 2, \ldots, L \}$, at
each time $t$, we observe a collection of contexts of covariate
vectors $ \{\bx_{t,\ell} \}_{\ell=1}^L$ and apply the same decision
$\ba_t$ to all targets. We then observe a batch of rewards
$\{y_{t,\ell}\}_{\ell=1}^L$, which we model as follows
\begin{align}
\label{eq:reward}
\Exs[\Reward_\ell \mid \bx_{t, \ell}, \ba_t] & = \ba_{t}^\top \bThetaTrue
\bx_{t,\ell} \quad \mbox{for $\ell = 1, 2, \cdots, L$}.
\end{align}
For simplicity, we assume that the reward function of each target
location is independent, but note that it is possible to extend our
model to account for dependency.

\subsection{Firm's Objective and  Regret}
\label{Subsec-regret}

The objective of the firm is to design a policy $\pi$ that chooses a
sequence of history-dependent actions $(\ba_1, \ba_2, \ldots, \ba_T)$
so as to maximize the \emph{expected cumulative revenue}
\begin{align}
\label{eq:cum-reward}
\E_\pi  \left [ \sum_{t=1}^T \sum_{\ell=1}^L \ba_t^\top \bThetaTrue \bx_{t,\ell} \right].
\end{align}
If the representation matrix $\bThetaTrue$ is known a priori, then the
firm can choose an optimal decision $\ba^*_t \in \ActionSpace$ that
maximizes the sum of the reward functions \eqref{eq:reward} across $L$
targets, i.e., $\ba_{t}^{*} \defn \sup_{\ba \in \ActionSpace_t}
\sum_{\ell=1}^L \ba^{\top} \bThetaTrue \bx_{t,\ell}$.  We call this
optimal solution a \emph{clairvoyant solution} and the clairvoyant
revenue over the time horizon is given by $\sum_{t=1}^T
\sum_{\ell=1}^L \ba_{t}^{* \top} \bThetaTrue \bx_{t,\ell}$.  Of course,
this clairvoyant value is not attainable because $\bThetaTrue$ is unknown
in practice, but it serves as a useful benchmark for performance
of any algorithm.

With this benchmark in place, we analyze procedures design to learn
policies $\pi$ that minimize the \emph{cumulative regret}---that is,
the gap between the expected cumulative revenue over the time horizon
$T$ between the revenue earned by implementing policy $\pi$, and the
clairvoyant solution.  Equivalently, we seek to minimize the
\emph{time-averaged regret}\footnote{The time-averaged form of regret
is rescaled by $1/T$ relative to the cumulative regret; we do so with the intent that
our bounds can be stated in the form of standard consistency
guarantees, with the error decreasing to zero as $T$
increases.}---that is, the quantity
\begin{align}
\label{eq:regret}
\Regret{T}{\pi} & \defn \frac{1}{T} \E_\pi \left[ \sum_{t=1}^T
  \sum_{\ell=1}^L \ba_t^{*\top} \bThetaTrue \bx_{t,\ell} - \ba_{t}^\top
  \bThetaTrue \bx_{t,\ell} \right] = \frac{1}{T} \E_\pi \left[
  \sum_{t=1}^T \sum_{\ell=1}^L (\ba_t^* - \ba_{t})^\top \bThetaTrue
  \bx_{t,\ell} \right].
\end{align}
Since the representation matrix $\bThetaTrue$ is unknown to us, we need to
design an algorithm that simultaneously learns the representation
matrix on the fly (exploration) and maximizes the total revenue
(exploitation). This exploration--exploitation problem with
high-dimensional action and covariate spaces, to which we refer as a
\emph{doubly high-dimensional contextual bandit}, is the focus of this
paper.


\subsection[Low-Rank Structure]{Low-Rank Structure of $\bThetaTrue$ and its implications}
\label{Subsec-lowrank}

As argued previously, although actions and covariates are
high-dimensional, the demand and sales are often driven by certain
latent factors.  Therefore, it is reasonable to impose a low-rank
assumption on the representation matrix $\bThetaTrue$.

To understand the meaning of such a low-rank condition, consider a
matrix $\bThetaTrue$ that is of rank $\myrank \ll \min \{d_a, d_x \}$.  It
has a singular value decomposition of the form $\bThetaTrue = \bU \bS
\bV^T$, where $\bS = \operatorname{diag}\{s_1, \ldots, s_\myrank \}$
is a diagonal matrix with the ordered singular values $s_1 \geq s_2
\geq \cdots \geq s_\myrank > 0$, and both $\bU \in \real^{d_a \times
  \myrank}$ and $\bV \in \real^{d_x \times \myrank}$ are matrices with
orthonormal columns, corresponding to the left
$\{\bu_j\}_{j=1}^\myrank$ and right singular vectors
$\{\bv_j\}_{j=1}^\myrank$, respectively.  With this notation, the
reward function~\eqref{eq:reward-l} can be decomposed as
\begin{align}
\E[\Reward \mid \ba_t, \bx_t] = \ba_t^\top\bThetaTrue\bx_t =
\sum_{j=1}^\myrank s_j \inprod{\ba_t}{\bu_j} \cdot
\inprod{\bv_j}{\bx_t}
\end{align}
In other words, the mean reward is the summation of the products
between the action projected on the left singular vector and the
covariates projected on the right singular vector, weighted by the
singular values.  The low-rank condition on $\bThetaTrue$ dictates that
the expected reward is governed by a relatively small number of
interactions between linear combinations of the action features and
covariates. In this way, our model automatically explores the
low-dimensional structure of the action and context vectors in terms
of their effects on the reward via the left and right singular
vectors; consequently, we can draw conceptual and modeling insights
from the spectral structures of both the action and covariates.  In the
context of joint assortment-pricing, the left singular vectors $\bu_j$
(respectively, the right singular vectors $\bv_j$) can be thought of
as weights associated with the latent product factor $j$
(respectively, the latent covariate factor $j$).

We note that our empirical studies provide evidence for the
suitability of the low-rank structure of $\bThetaTrue$.  For instance, in
our instant noodle example, there are $13$ possible flavors along with
the price, so a total of $14$ attributes.
In addition to including these attributes themselves, we also include
their squares (so that we can model non-linear effects), for a total
of $30$ meta-attributes.  We include these $30$ meta-attributes for
each of $K = 30$ possible product slots considered, leading to an
action vector of dimension
\begin{align*}
  d_\action & = 30 K + 1 \; = \; 841,
\end{align*}
where the additional one accounts for the presence of a constant
offset term.  In terms of covariates, we include $31$ provinces, the
year 2021/2022, $12$ months, weekdays, an indicator of the annual sale
events and an additional one, leading to a covariate vector of
dimension $d_\state = 50$.  For this pair $(d_\action, d_\state) =
(841, 50)$, our procedure learns a matrix $\bThetaHat$ with rank $4$.
See~\Cref{Subsec-mk} for further discussion of the latent factors, and
their real-world significance.

\subsection{Relation to Other Models}
\label{Subsec-comp-gen}

In this section, we discuss how our model is related to other known
bandit models and approaches to dynamic pricing (see
Sections~\ref{sec:compatibility_bandit} and~\ref{sec:compatibility_pricing},
respectively).

\subsubsection{Connection with Other Bandit Models.}
\label{sec:compatibility_bandit}

Let us summarize some connections to other bandit models that can be
re-expressed as special cases of our reward model~\eqref{eq:reward}.
In this section, we recycle the notation $K$, using it to represent
the number of actions in the multi-action bandit by
convention.\footnote{Please note, this should not be confused with the
maximum number of slots in our general model set-up.}

\begin{enumerate}
    \item A \emph{multi-arm bandit} is defined by $K$ independent
      actions~\citep{robbins1952some}.  The $i$-th action can be
      represented by the unit basis vector \mbox{$\ba_i = (0,0,\cdots,
        1, \cdots, 0)$}, where the single $1$ appears in the $i$-th
      entry. By setting $\bx = 1$ and $\bThetaTrue \in \Real^{K \times 1}$ be the
      rank one matrix with entries $\bThetaTrue_{ii}=\mu_i$, we have
      $\ba_i^\top \bThetaTrue \bx = \mu_i$ as a special case of our model.
      The \emph{linear bandit}
      (e.g.,~\citep{rusmevichientong2010linearly,dani2008stochastic,auer2002using,abbasi2011improved})
      is a natural generalization of the multi-arm bandit, in which
      each of the $K$ possible actions is associated with an arbitrary
      vector $\ba_k$, and the reward function is a mapping $\ba
      \mapsto \mu(\ba) = \inprod{\btheta}{\ba}$. Augmenting $\ba$ with
      $\bx = 1$ as the ``context'', we can write this model in the
      form $\bThetaTrue = \btheta$, again leading to a rank one setting.

    \item In a \emph{high-dimensional contextual extension} of the
      multi-arm bandit~\citep{bastani2020online}, in addition to the
      $K$ arms, each represented with action vector $\ba_i$ as above,
      we also have a (possibly high-dimensional) context vector $\bx
      \in \real^{d_x}$.  The reward associated with arm $i$ is given
      by $\inprod{\bbeta_i}{\bx}$.  By defining the matrix
      \mbox{$\bThetaTrue = (\bbeta_1, \bbeta_2, \cdots, \bbeta_K)^\top \in
         \Real^{K \times d_x}$,} we can represent this model in our bilinear
      form.
    \item \emph{Continuum-action bandits (without context)}.  Given a
      continuous action $b \in \real$, these models
      (e.g.,~\citep{Kleinberg2019}) use a general non-parametric
      reward function $b \mapsto \mu(b)$.  Such models are actually
      non-parametric in nature, but can be approximated by linear
      bandits by lifting the action space.  More precisely, since all
      continuous functions on a bounded interval can be approximated
      by polynomial functions to arbitrary precision, we can
      approximate the reward function using a polynomial of order at
      most $N$.  Defining the augmented action vector \mbox{$\ba = (1,
        b, b^2, b^3, \cdots, b^N)$,} we then have a linear bandit in
      dimension $N+1$.
\end{enumerate}

\subsubsection{Compatibility with Pricing Models.}
\label{sec:compatibility_pricing}

The linear price-demand model plays a central role in the literature
on dynamic pricing.  This linear demand model is a special case of our
bilinear reward model, and can also be extended (by augmenting the
state-action vectors) to incorporate nonlinear demand curves.

We focus on a recent extension to the linear price-demand curve
proposed by~\citet{ban2021personalized} which considers the
personalized pricing problem and assumes a personalized demand model
whose parameters depend on the context vector. Specifically, they
assume the demand model as
\begin{align}
\label{eq:ban-demand}
D_t = \balpha^T \bx_t + (\bbeta^T \bx_t) p_t + \epsilon_t
\end{align}
where $\balpha,\bbeta \in \Real^{d_x}$ are the unknown demand
parameter vectors, $p_t \in \Real^{+}$ is the price, $\bx_t \in
\Real^{d_x}$ customer characteristics, and $\epsilon_t$ is the
noise. In this model, the inner product $\inprod{\balpha}{\bx_t}$
captures the ``context-dependent customer taste and potential market
size'', whereas the inner product $\inprod{\bbeta}{\bx_t}$ captures the
``context-dependent price sensitivity''.  Therefore, the expected revenue
at time $t$ is
\begin{align}
\label{eq:ban-revenue}
\E[Y \, \mid \bx_t, p_t] = p_t \big[ \inprod{\balpha}{\bx_t} +
  \inprod{\bbeta}{ \bx_t} p_t \big].
\end{align}
Note that the mean reward \eqref{eq:ban-revenue} is a special case of
our model with action $\ba_t = (p_t, p_t^2)$, covariate $\bx_t$ is the same as in the demand model \eqref{eq:ban-demand}, and unknown parameter matrix \mbox{$\bThetaTrue =
  (\balpha; \bbeta) \in \Real^{2 \times d_x}$.}



\section{Algorithm}
\label{Sec-Algo}

In this section, we describe our learning algorithm for the doubly
high-dimensional contextual bandit problem.  It involves two phases at
each time period and is thus modular and generalizable.  The first
phase is devoted to learning a low-rank representation, whereas the
assortment-pricing decisions are made in the second phase.  In the
first phase, the algorithm constructs an estimate $\hTheta_t$ using a
penalized form of least-squares regression with covariates $(\ba_i,
\bx_{i, \ell})$ and responses $\reward_{i, \ell}$ for $i=1, \ldots, t$
and $\ell = 1,\ldots, L$.  In the second phase, we use the estimated
bilinear reward induced by $\hTheta_t$ to choose assortment-price
actions within the action space
$\mathcal{A}_t$. See~\Cref{algo:high-arm} for the full details.

\begin{algorithm}[tb]
   \caption{The \hicoab~Algorithm.}
    \label{algo:high-arm}
\begin{algorithmic}
    \STATE {\bfseries Result:} Actions $\ba_{\tinit + 1}, \ldots, \ba_T$.
    \STATE {\bfseries Input:} Initial step number $\tinit$; set of
    possible actions $\mathcal{A}_{\tinit}$, action vectors based on
    domain knowledge $\{\ba_i\}_{i=1}^{\tinit}$, covariate vectors
    $\{\bx_{i,\ell}\}_{i=1}^{\tinit}$, rewards $\reward_{i,\ell}$ for
    $\ell = 1, \ldots, L$, and exploration parameter $h$.  \STATE
    \textbf{Initialization:} $\lambda_0 \leftarrow \frac{2}{ \tinit
      L} \opnorm{\sum_{i=1}^{\tinit} \sum_{\ell=1}^L \big
      |\ba_i^\top\hTheta_{\tinit} \bx_{i, \ell} -\reward_{i,\ell}
      \big| \; \bx_{i,\ell} \ba_i^\top }$, \quad \mbox{and} \quad $t
    \leftarrow \tinit +1$.  \WHILE{$t < T$} \STATE $\lambda_{t}
    \leftarrow \lambda_0 / \sqrt{t} $; \STATE \textbf{Step 1: Low-rank
      representation learning:}\\ \STATE $\hTheta_t \leftarrow \arg
    \min_{\bTheta} \Big \{ \frac{1}{2t \;L} \sum_{i=1}^{t}
    \sum_{\ell=1}^L ( \ba_i^\top \bTheta \bx_{i,\ell} -
    \reward_{i,\ell} )^2 + \lambda_{t} \mynucnorm{\bTheta} \Big \} $;
    \STATE \textbf{Step 2: Policy learning:}\\ \STATE \quad $\ha_{t+1}
    \leftarrow \arg \max_{\ba \in \mathcal{A}_t } \Big \{
    \sum_{\ell=1}^L \ba^\top \hTheta_t \bx_{t+1,\ell} \Big \}$ (take the one with largest norm if the solution is not unique); \IF{
      $t \notin \{ \floor{w^{3\over 2}} : w\in \mathbb{Z}_+ \}$ }
    \STATE \textit{Exploitation:} $\ba_{t+1} \leftarrow \ha_{t+1}$;
    \ELSE \STATE \textit{Exploration:} $\ba_{t+1} \leftarrow \ha_{t+1}
    + \bdelta_{t+1}$ where $\bdelta_{t+1} \sim N(\bm 0_{d_a},
    h\bI_{d_a})$, update action space $\mathcal{A}_{t+1}$; \ENDIF
    \STATE Apply action $\ba_{t+1}$ and observe reward
    $\reward_{t+1,\ell}$ for $\ell = 1, \ldots, L$; \STATE $t
    \leftarrow t + 1$; \ENDWHILE
\end{algorithmic}
\end{algorithm}

\subsection{Step 1: Low-rank Representation Learning.}

The first step of the algorithm is to estimate the low-rank
representation matrix $\bThetaTrue$.  As motivated in
\Cref{Sec-Formulation}, it is reasonable to impose a low-rank
condition on $\bThetaTrue$.  Disregarding computational issues, one might
imagine estimating $\bThetaTrue$ by imposing a rank constraint, or a
penalty involving the rank.  However, rank penalization is a non-convex
problem with associated computational challenges, so that it is
standard to replace it with the nuclear norm so as to obtain a convex
problem.  Doing so in our context yields the nuclear-norm regularized
estimator
\begin{align}
\label{eq:rank-pen2}
\hTheta_t \defn \argmin_{\bTheta} \left \{ \frac{1}{2 L t}\sum_{i=1}^t
\sum_{\ell=1}^L \left( \ba_i^\top \bTheta \bx_{i,\ell} -
\reward_{i,\ell} \right )^2 + \lambda_{t} \cdot \mynucnorm{\bTheta}
\right \},
\end{align}
where $\lambda_t > 0$ is a regularization parameter. We update the
parameter $\lambda_t$ over the time periods with $\lambda_t =
\tfrac{\lambda_0}{\sqrt{t}}$, where $\lambda_0 > 0$ is an initial
choice, specified by cross-validation. The decay rate $1/\sqrt{t}$ is
chosen to match the typical standard deviation of the first
data-dependent term: with $L$ being constant, it is the sample average of
$t$ terms.

\subsection{Step 2: Policy Learning.}

Given an estimate of the low-rank matrix $\bThetaTrue$, we can proceed to
the action step, i.e., to select the assortment and pricing for time
$t$.  The goal of the action step is to \emph{exploit} the knowledge
we have learned, i.e., $\hTheta_t$, so as to decide on the next action
$\ba_{t+1}$ that maximizes the reward, and at the same time to
\emph{explore} actions that better inform the true $\bThetaTrue$, which in
turn will help make better decisions to achieve higher long-term
rewards.  Specifically, given the estimate $\hTheta_t$ and the
covariate vectors $\bx_{t+1,\ell}$ for $\ell \in [L]$, we look for an
action $\ha_{t+1}$ in the action space $\mathcal{A}_t$ that maximizes
the total rewards across $L$ objects:
\begin{align}
\label{eq:max-reward}
\ha_{t+1} & \defn \arg \max_{\ba \in \mathcal{A}_t } \left \{
\sum_{\ell=1}^L \ba^\top \hTheta_t \bx_{t+1,\ell} \right \}.
\end{align}
At a subset of times, we further perturb $\ha_{t+1}$ for the purpose
of exploration by adding random noise to each coordinate as follows:
$\ba_{t+1} = \ha_{t+1} + \bdelta_{t+1}$ where $\bdelta_{t+1} \sim
N(\bm 0_{d_a}, h \bI_{d_a})$ and $h$ is a tuning parameter.  In our
current algorithm, we perform this perturbation at times $t \in \{
\floor{w^{3\over 2}} : w \in \mathbb{Z}_+ \}$.

The intuition for this particular choice ($\floor{w^{3\over 2}}$) is
to explore more in the initial stage and exploit less in the later
stage of the algorithm.  To be specific, there are approximately
$T^{2\over 3}$ steps for exploration before time $T$. The density of
exploration at a small time frame around $T$ is $T^{-{1\over 3}}$,
which goes to zero as $T\to \infty$.  Note that the exponent need not
be $\frac{3}{2}$, but can be any number strictly larger than $1$; this
choice affects trade-offs between different terms in the regret, as
discussed later in~\Cref{rmk:rate,rmk:burn}.

The form of randomness used in the exploration step is another design
parameter of the algorithm.  For each exploration step, one can also
let \mbox{$\bdelta_{t+1} \sim N(\bm 0_{d_a}, \diag(\htau_t))$} where
each element of $\htau_t$ is the coordinate-wise standard error of the
previous actions $\{\ba_{i}\}_{i=1}^t$. This choice serves to set an
appropriate scale for exploration while avoiding more complicated
procedures for tuning the parameter $h$. Finally, we update the action
space $\mathcal{A}_{t+1}$ according to $\ba_{t+1}$.  For example, if
the action space $\mathcal{A}_t \in \mathbb{R}^{d_a}$ can be defined
by an upper limit $\ua_t$ and a lower limit $\la_t$, then we simply
expand the action space by pushing the boundary of each coordinate to
$\ba_{t+1,j}$ if $\ba_{t+1,j} \notin [\la_{t,j}, \ua_{t,j}]$ for
$j=1,\ldots,d_{a}$.

\begin{remark}[Initialization]
The initial step number $\tinit$ and actions $\{\ba_i\}_{i=1}^{\tinit}$
depends on the availability of historical data. 
When there exists historical data, $\tinit$ is the number of steps 
in the historical data and the actions are corresponding real actions. 
The real actions are often guided by ``domain knowledge'', such as market research, past experience, and heuristics. Historical data is often available in real applications so we design our algorithm to take advantage of all the data available.
In the absence of historical data, actions can either be domain-informed or randomly selected within the action set for reasonably small number of steps. 
\end{remark}

\begin{remark}[Adaptivity]
\label{rmk:adapt-robust}
We note that our algorithm is adaptive (w.r.t. $T$), and also robust,
especially compared with explore-then-commit type algorithms that
sample arms for a period of time and then use the estimates for the
remaining horizon. Our algorithm does not require specifying the
$T$-dependent tuning parameters, and it updates the representation
matrix across the entire time horizon, making it more suitable for
online learning. Moreover, it does not require knowing or
pre-specifying the target rank $r$.

This adaptivity is of practical importance as in practice for the
assortment and pricing, retailers would like to consistently achieve
good performances for periods of any length, instead of just a
specific fixed period of time. This is why we keep exploring ---
though at a decreasing frequency --- and update the parameters
throughout.
\end{remark}

\begin{remark}[Interpretability]
To take advantage of the interpretability of our model, we can further
explore the structure of the $\hTheta_t$.  Specifically, we can apply
singular value decomposition (SVD) on $\hTheta_t$ to explore the
underlying latent structure of the covariates from the right singular
vectors and the latent structure of the arms from the left singular
vectors.  One can further rotate the singular vectors using techniques
in factor analysis such as Varimax \citep{kaiser1958varimax,
  rohe2020vintage} so as to obtain a sparse/simplified loading
structure for easier interpretation.
\end{remark}

\section{Regret Analysis}
\label{Sec-Theory}

We now turn to some theoretical analysis of our procedure, beginning
in~\Cref{SecThm} with the statement of our main theorem, and with the
following~\Cref{Subsec-proof} devoted to proofs.

\subsection{Instance-dependent Regret Bound}
\label{SecThm}

We begin by stating a non-asymptotic instance-dependent bound on the
expected time-averaged regret incurred by~\Cref{algo:high-arm}.  It
shows that in for any problem and for any dimensions, the expected
time-averaged regret decays to zero at least as fast
$\otil(T^{-1/6})$.

Our analysis applies to an instantiation of~\Cref{algo:high-arm} with
actions chosen randomly according to the exploration protocol $\ba_{t}
= \hat{\ba}_{t} + \bdelta_{t}$, where $\bdelta_{t} \sim
N(\mathbf{0}_{d_a}, h \bI_{d_a} )$, implemented at each time instant
\vspace{-.06in}
\begin{align}
  \label{EqnExploreFreq}
  t \in \{ \floor{w^{3\over 2}} \, \mid \, w = 1, 2, 3, \ldots \}.
\end{align}
The constraint set for this instance is $\mathcal{A}_{\tinit} =\{\ba \in \R^{d_a}:\|\ba\|\leq 1\}$ with $\tinit=1$ and a random selected first action.
As our analysis involves an additional assumption on the reward error,
we introduce a short-hand notation for the reward error,
\vspace{-.06in}
\begin{align}
\label{eq:epsilon}
\varepsilon_{t,\ell} = y_{t,\ell} - \ba_{t}^T \bThetaTrue \bx_{t,\ell}.
\end{align}
Finally, our statement involves a burn-in period \mbox{$\burnin =
  C_{h,L,\lambda_0} (d_x + d_a)^6 \left(\log(d_x + d_a) \right)^4$.}
 \vspace{-.04in}
\begin{theorem}
\label{thm:main}
Suppose that the ground truth $\bThetaTrue$ has rank $\myrank$, we observe
covariates $\bx_{t,\ell} \overset{i.i.d}{\sim} N(\mathbf{0}_{d_x} ,
\bI_{d_x})$, and the reward errors $\varepsilon_{t,\ell}
\overset{i.i.d}{\sim} N(0, \sigma^2)$.
Then there are universal constants $\{c_j\}_{j=1}^4$ such that for all
$T \geq \burnin$, the expected time-averaged regret is bounded as
\begin{multline}
\label{eq:thmmain}
\Regret{T}{\pi} \leq \frac{c_1}{T}{\sqrt{Ld_x} \opnorm{\bThetaTrue}
  \burnin } + \frac{c_2 \log{T}}{T} \sqrt{Ld_x} \opnorm{\bThetaTrue} +
\frac{c_3}{T^{1/6}} \frac{\lambda_0 \sqrt{ 2 \myrank d_x L}}{h^2} +\\
\frac{c_4\log{T}}{T^{1/6}} \frac{ d_x(\sqrt{L}+\sqrt{L\log{L}})+ 3\sqrt{d_xL}\log{L}+2\sqrt{Ld_x} }{h^2}\sigma.
\end{multline}
\end{theorem}

\begin{remark}[Comments on $T$ and dimension dependence]
\label{rmk:rate}
In rough terms, \Cref{thm:main} guarantees that the expected regret
converges to zero at least as quickly as $\frac{\log{T}}{T^{1/6}}$ as
$T$ tends to infinity.
The convergence rate depends on the frequency of the exploration which
depends on the exponent $3\over 2$ in the exploration set, $\{\lfloor
w^{3\over 2} \rfloor: w \in \mathbb{Z}_+\}$.  It could be possible to
further tune this exponent for a faster convergence rate, and we leave an optimal choice for future
work.

What is most important about our convergence guarantee is that the
product $d_x d_a$ of the state and action dimensions \emph{does not}
appear in the bound: rather, any dimension factors are multiplied only
by the rank $r$, which we expect to be far lower than the dimension.
Thus, best of our knowledge, our result stands as the first
convergence result with non-trivial dimension scaling (i.e., $\ll d_x d_a$) for
doubly high-dimensional contextual bandits.

As we argued, our model
is more general and expressive than many existing bandit models,
making it intrinsically more complex, and requires analysis from
scratch. That being said, we provide a non-asymptotic,
instance-dependent bound.  Moreover, our algorithm does not require
prior knowledge of $T$ and $r$ as mentioned in
\Cref{rmk:adapt-robust}, and our bound also holds consistently for all
$T$ and $r$, which we will further discuss in
\Cref{rmk:rate-adaptivity}.
Establishing tighter bounds in various specific (well-studied) settings, i.e., special cases of our model, requires separate analyses, which deviates from our main goal. Nevertheless, our simulation shows that our method 
outperforms the state-of-the-art methods for these specific settings in Sections \ref{Subsec-price}-\ref{Subsec-bandit}.
\end{remark}

\vspace{.02in}
\begin{remark}[Burn-in term]
  \label{rmk:burn}
The first term in the bound~\eqref{eq:thmmain} is a burn-in term,
where the algorithm is gaining knowledge of $\bThetaTrue$ from
scratch. We do not impose any assumptions on these starting steps so
that we have a relatively conservative burn-in term.  In practice, we
can leverage historical data to obtain an initial estimate of
$\bThetaTrue$ so that the burn-in term can be much smaller.

The order of the burn-in term depends on the exponent---currently
$3/2$---used to specify the exploration
frequency~\eqref{EqnExploreFreq}.  Smaller values of this exponent
lead to more exploration, and hence a smaller burn-in term.  As noted, it would be interesting to determine optimal choices of the
exponent.
\end{remark}

\vspace{.02in}
\begin{remark}[Constant $C_{h,L,\lambda_0}$ of $\burnin$]
\label{rmk:lambda0}
While constant $C_{h,L,\lambda_0}$ depends on $h,L,\lambda_0$, the
primary dependency is actually on $h$ and $L$. The order of
$\lambda_0$ in terms of dimensions and noise level is
$\sigma\sqrt{d_x}$.  We do not assume the order of $\lambda_0$ or
bound it with a high probability bound in order to show its role in
time-averaged expected cumulative regret. If we utilize the order
$\sigma \sqrt{d_x}$, then $C_{h,L,\lambda_0}$ can be replaced by a
constant depending on $h$ and $L$ only.
\end{remark}

\begin{remark}[Dependence on dimensions $d_a, d_x$ and rank $\myrank$]
When $T$ is small, the first ``burn-in'' term dominates.  It depends
on $T$ and the dimensions but not the rank.  As $T$ grows, the last
two terms dominate. Recall from \Cref{rmk:lambda0} that $\lambda_0$ is
of order $\sigma\sqrt{d_x}$, so the third term depends on $T, d_x$ and
$\myrank$ but not $d_a$; it has the order $\Omega({d_x\sqrt{r}
  T^{-{1\over 6}}})$. The last term is of the order $\Omega({d_x
  T^{-{1\over 6}}} \log{T})$. In terms of $T$, these last two terms
are of the same order up to $\log T$.
\end{remark}
%
\vspace{.02in}
\begin{remark}[Adaptivity]
\label{rmk:rate-adaptivity}
\Cref{algo:high-arm} is adaptive as it does not require
 knowing $T$ a priori (except the ending point) as mentioned in
 \Cref{rmk:adapt-robust}; moreover, the non-asymptotic bounds
 hold for all $T$.  This adaptivity is of both theoretical interest and
 practical importance.  Adaptivity overcomes the limitations of the
 traditional bandit framework, which possibly favors good performance
 at a specific $T$ at the expense of other values.
 These limitations lead to
 algorithms involving $T$-dependent tuning parameters. In practice, it
 is preferable to have algorithms that do not require such tuning yet
 consistently perform well across all $T$. This important and desirable adaptivity property, unfortunately, often comes at the cost of the rate, as underscored by \citet{cai2017confidence}.
\end{remark}
\vspace{0.02in}
\begin{remark}[Assumptions]
To convey the main idea in a simple way, we have chosen to enforce
relatively stringent assumptions.  However, neither the normality
assumption nor the shape of the constraint set are essential to the
core structure of the proof.
\end{remark}

\subsection{Proof Sketch}
\label{Subsec-proof}

At a high level, the proof of~\Cref{thm:main} consists of two major
steps:
\begin{itemize}
\item \Cref{Sec-theta-bound} provides the high-probability bound on
  the estimation error of the low-rank representation matrix estimator
  $\hTheta_t$;
\item
  \Cref{Sec-regret-bound} provides a non-asymptotic upper bound for
  the expected regret $\Regret{T}{\pi}$.
\end{itemize}
\noindent Here we provide a sketch of each of the steps, referring the
reader to Appendix~\ref{proof:thm} for all the technical details.

\subsubsection{Bounding the Estimation Error}
\label{Sec-theta-bound}

An accurate estimate of the matrix $\bThetaStar$ is required to obtain
good actions, so that our first step is to bound this estimation
error.  We introduce the shorthand $\dTheta_t \defn \hTheta_t -
\bThetaStar$ for the error of the estimate $\hTheta_t$ at round
$t$. Our first auxiliary result provides a high-probability bound on
the Frobenius norm error $\fronorm{\dTheta_t}$.
\begin{proposition}
\label{prop:theta_bound}
For any time $t \geq \burnin \defn C_{h,L,\lambda_0} (d_x+d_a)^6
\left(\log(d_x+d_a)\right)^4$, we have
\begin{subequations}
\begin{align}
  \fronorm{\dTheta_t} \leq { 9t^{1\over 3}(2+ \sqrt{t}) (
    \sqrt{d_x\log{tL}}+2\log{tL} )\sigma \over th^2} + 6\lambda_0 {\sqrt{2r} \over
    h^2 t^{1\over 6}}
\end{align}
with probability at least
\begin{align}
  \label{EqnDefnFailure}
\phi(t) & \defn 1 - {4\over t} - {1\over t^2} - {3\over Lt} - {2\over
  L^3t^3 } - {1\over t^2L}.
\end{align}
\end{subequations}
\end{proposition}
In this statement, the quantity $C_{h,L,\lambda_0}$ depends on $L, h$
and $\lambda_0$, but not other problem parameters. \\

The technical challenge in establishing this result lies in the fact
that the actions taken are based on past data, and also affect future
data, resulting in a highly non-i.i.d. dataset.  For this reason, the
summands in the empirical loss function are strongly dependent, so
that known results for matrix completion, based on i.i.d. or weakly
dependent data, are no longer applicable.  Herein lies the need for
care and technical innovation to handle the adaptive nature of bandit
data collection.

The proof can be roughly separated into three steps, which we describe
at a high-level below.  In the first step, we use the optimality
conditions defining the estimate to derive a basic inequality, which
we then re-arrange via a Taylor series into a more amenable form.  In
Steps 2 and 3, we use empirical process theory and concentration of
measure to derive high-probability upper bounds on different
components of this inequality.  We conclude by combining each of these
steps.  

\begin{enumerate}
\item First, we observe that since $\bThetaHat_t$ minimizes the
  function $\bTheta \mapsto \LL[t]{\bTheta} + \lambda_t
  \mynucnorm{\bTheta}$, we have the basic inequality
 \begin{align*}
    \LL[t]{{\hTheta}_t} + \lambda_t \mynucnorm{{\hTheta}_t} \le
    \LL[t]{\bThetaStar} + \lambda_t \mynucnorm{\bThetaStar},
 \end{align*}
By performing a first-order Taylor series expansion of the loss
function around $\bThetaStar$, this inequality implies that
\begin{align}
\label[ineq]{eq:first}
\ErrTerm_t(\dTheta_t) & \leq - \tracer{\nabla
  \LL[t]{\bThetaStar}}{\dTheta_t} + \lambda_t\left(
\mynucnorm{\bThetaStar} - \mynucnorm{\bThetaStar + \dTheta_t} \right),
\end{align} 
where we have defined the Taylor series error function
\begin{align*}
  \ErrTerm_t(\dTheta) \defn \LL[t]{\bThetaStar + \dTheta} -
  \LL[t]{\bThetaStar} - \tracer{ \nabla \LL[t]{\bThetaStar}}{\dTheta}.
\end{align*}
The remainder of our analysis focuses on the three terms in the
\Cref{eq:first}.  We need to establish a lower bound on
the left-hand side term $\ErrTerm_t(\dTheta_t)$, and upper bounds on
the two terms on the right-hand side.

\item Beginning with the left-hand side, we prove the following lower
  bound:
\begin{lemma}
\label{lem:main}
Under the assumptions of \Cref{thm:main}, for any $ t \geq 2$, we have
\begin{align}
  \ErrTerm_{t}(\dTheta) & \geq { \floor{t^{2\over 3}} \over 2t} h^2
  \fronorm{\dTheta}^2 - \simplifykappa{t}\fronorm{\dTheta}^2
\end{align}
with probability at least $1-{1\over Lt} -{3\over t} $.
\end{lemma}
Note that the first term on the right-hand side scales as $t^{-1/3}
\fronorm{\dTheta}^2$, whereas the second term scales as
$t^{-2/3}\log{t} \opnorm{\dTheta}^2$. Since $\opnorm{\dTheta}^2 \leq
\fronorm{\dTheta}^2$, we have established a lower bound on
$\ErrTerm_{t}(\dTheta)$ that scales as $t^{-1/3} \fronorm{\dTheta}^2$
for large $t$, along with a pre-factor that depends on $(h, d_x, d_a,
L)$.

\item Our next lemma provides a high-probability bound on the quantity
  $\big| \tracer{\nabla \LL[t]{\bThetaStar}}{\dTheta} \big|$, which
  appears as the first term on the right-hand side of the
  \Cref{eq:first}.  It involves the two pre-factors:
\begin{align*}
\preone{t} & \defn { \sigma (6+ 3\sqrt{t}) ( \sqrt{d_x\log{tL}}+2\log{tL} ) \over t} \quad \mbox{and} \\
\pretwo{t} & \defn 2h\sigma t^{-2/3}\log{t}\sqrt{ \max\{d_a,d_x\} \log{(d_a+d_x)}
  \over L}  \\ 
  & \qquad + \frac{8h\sigma}{t}
\sqrt{\log{(tL)}}\sqrt{(d_x+3\log(Lt))( d_a+3\log{t} )}( \log(d_x+d_a)
+2 \log{t} ) 
\end{align*}
\vspace{-.3in}
\begin{lemma}
\label{lem:nabla}
Under the assumptions of~\Cref{thm:main}, uniformly over all matrices
$\dTheta \in \R^{d_a \times d_x}$, we have
\begin{align}
 \big| \tracer{\nabla \LL[t]{\bThetaStar}}{\dTheta} \big| \leq &
 \preone{t} \; \fronorm{\dTheta} + \pretwo{t} \; \mynucnorm{\dTheta}.
\end{align}
with probability at least $1- \frac{1}{t^2L}-{2\over L^3t^3} - {2\over
  Lt} -{1\over t} -{1\over t^2}$.
\end{lemma}
\noindent By examining the prefactors $\preone{t}$ and $\pretwo{t}$
and considering their scaling in the pair $(t, \dTheta)$, we see that
$|\tracer{\nabla \LL[t]{\bThetaStar}}{\dTheta}|$ is upper bounded by a
quantity scaling $\tfrac{\log(t)}{\sqrt{t}} \fronorm{\dTheta}$ for
sufficiently large $t$.
\end{enumerate}

With these two lemmas in place, let us sketch out the remainder of the
proof, deferring the full argument to Appendix~\ref{app:proof-prop1}.
For a rank $r$ matrix $\bThetaStar$, a spectral decomposition argument
can be used to show that
\begin{align}
\label{eq:delta-nuclear}
\mynucnorm{\bThetaStar} - \mynucnorm{\bThetaStar + \dTheta_t} & \leq
\sqrt{2r} \fronorm{\dTheta_t}.
\end{align}
We use this inequality to control the remaining term in the
bound~\eqref{eq:first}.

As noted in our discussion following Steps 2 and 3, for sufficiently
large $t$, we have established the scaling relations
$\ErrTerm_t(\dTheta_t) \gtrsim \tfrac{1}{t^{1/3}}
\fronorm{\dTheta_t}^2$, and $|\tracer{\nabla
  \LL[t]{\bThetaStar}}{\dTheta}| \precsim \tfrac{\log(t)}{\sqrt{t}}
\fronorm{\dTheta}$ for sufficiently large $t$.  Combining these
scaling relations with the bound~\eqref{eq:delta-nuclear}, our choice
$\lambda_t \sim \tfrac{1}{\sqrt{t}}$, and substituting into the
\Cref{eq:first}, we find that
\begin{align*}
\frac{1}{t^{1/3}} \fronorm{\dTheta_t}^2 \precsim
\tfrac{\log(t)}{\sqrt{t}} \fronorm{\dTheta_t} + \frac{1}{\sqrt{t}}
\sqrt{2r} \fronorm{\dTheta_t}.
\end{align*}
Consequently, we conclude that $\fronorm{\dTheta_t} \lesssim
\tfrac{1}{t^{1/6}} \log(t)$ with high probability.  Again, we refer
the reader to Appendix~\ref{app:proof-prop1} for all the technical
details, including careful tracking of the lower order terms.

\subsubsection{Bounding the Expected Regret:} 
\label{Sec-regret-bound}

At each round $t$, we define the event
\begin{align}
  \Event_t & \defn \Big \{ \fronorm{\dTheta_t} \leq \frac{ 9t^{1/3}(2+
    \sqrt{t}) ( \sqrt{d_x\log{tL}}+2\log{tL} )\sigma }{ th^2} + 6\lambda_0 {\sqrt{2r}
    \over h^2 t^{1/6}} \Big \}.
\end{align}
\Cref{lem:main} guarantees that for large $t$, $\Prob(\Event_t^c) \le
     {4\over t} +{1\over t^2} + {3\over Lt} + {2\over L^3t^3 } +
     {1\over t^2L^2}$. Considering the expectation of the regret on
     $\Event_t$ and $\Event_t^c$ separately, we show that both terms
     vanish with $t$ at a polynomial rate.

\section{Experimental Studies}
\label{Sec-Sim}

This section is devoted to some experimental studies of the behavior
of the proposed algorithm in different settings, both via controlled
simulations and applications to two real-world datasets.

In Sections~\ref{Subsec-price} through~\ref{Subsec-bandit}, we compare
the performance of \hicoab~with other bandit and pricing algorithms.
In all cases, we assume the reward error~\eqref{eq:epsilon}
follow a normal distribution with mean zero and variance $\sigma^2$,
i.e.,
\begin{align}
\label{eq:epsilon-normal}
\varepsilon_{t, \ell} \overset{i.i.d.}{\sim} N (0,\sigma^2).
\end{align}
We then revisit the instant noodle joint assortment-pricing case study
in~\Cref{Subsec-mk}.  In this context, we find that \hicoab~can boost
cumulative sales by a factor larger than $4$; moreover, examination of
the learned representation matrix $\bThetaHat$ provides insight into
the latent factors of actions and covariates that influence revenue.
Finally, in~\Cref{Subsec-manime}, we provide a real-world case study
analysis of the assortment-pricing problem faced by a manicure
start-up.  We defer some more technical details of the discussion to
Appendix~\ref{app:sim-case-study} in the supplementary material.

\subsection{Simulation Experiment I: Pricing Models}
\label{Subsec-price}

\begin{figure}
\centering
\widgraph{0.6\linewidth}{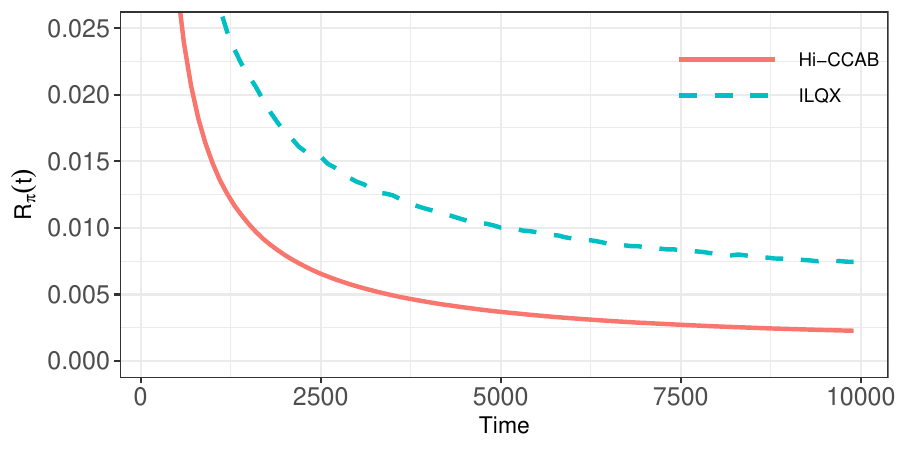}
  \caption{Time-averaged regret for \hicoab~and the iterated
    Lasso-regularized quasi-likelihood regression (ILQX) proposed by
    \cite{ban2021personalized}.}
\label{fig:pricing-regret}
\end{figure}
 
\sloppy We follow the simulation set-up introduced
by~\cite{ban2021personalized}: in particular, generate demands and
rewards according to the model~\eqref{eq:ban-demand}--\eqref{eq:ban-revenue}, with parameter vectors
\begin{align*}
  \alpha & \defn [1.1, -0.1, 0, 0.1, 0, 0.2, 0, 0.1, -0.1, 0, 0, 0.1, -0.1,
    0.2, -0.2], \quad \mbox{and} \\
\beta & \defn (-1) × [0.5, 0.1, -0.1, 0, 0, 0, 0, 0.2, 0.1, 0.2, 0,
  0.2, -0.1, -0.2, 0],
\end{align*}
and the variable $\epsilon_t$ follows a normal distribution with mean
zero and standard deviation $0.01$.

We compare the time-averaged regret $\Regret{t}{\pi}$ of ILQX (iterated lasso-regularized quasi-likelihood regression with price experimentation) proposed by \cite{ban2021personalized} with \hicoab. 
The basic idea of ILQX is to use LASSO to estimate the unknown $\alpha$ and $\beta$,
and at the same time to conduct price experiments for at least an order of $\sqrt{t}$ times.
As shown in \Cref{sec:compatibility_pricing}, their revenue model \eqref{eq:ban-revenue} is a special case of our model \eqref{eq:reward-l}.

\Cref{fig:pricing-regret} compares the performance, measured by the
time-averaged regret, of \hicoab~and ILQX. It is clear that
\hicoab~converges faster than ILQX.  As shown in
\cite{ban2021personalized}, ILQX converges faster than the greedy
iterated least squares \citep{keskin2014dynamic, qiang2016dynamic},
which decides the price based on the least square estimate of the
unknown $\alpha$ and $\beta$ at each iteration without experiments.
We see that \hicoab~has better performance than various dynamic
pricing algorithms, and is competitive in a continuum armed bandit
problem.

\subsection{Simulation Experiment II: Bandit Models}
\label{Subsec-bandit}

\begin{figure*}
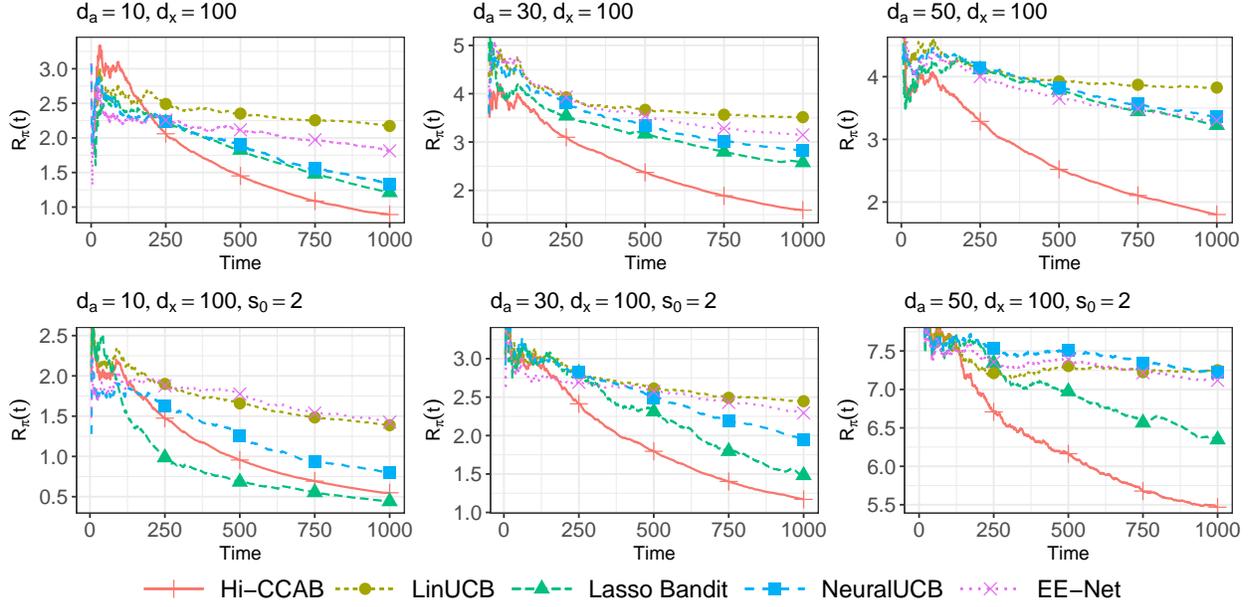

  \centering
  \widgraph{1\linewidth}{\figdir/fig_lin-non-sparse-sparse_avg_regret}
  \caption{Time-averaged regret in a multi-armed linear bandit
    setup for \hicoab~and other bandit algorithms. The upper row
    corresponds to a non-sparse case and the lower one corresponds to
    a sparse case.}
  \label{fig:sim}
\end{figure*}

In this simulation study, we consider a multi-armed linear bandit,
which (as discussed previously) corresponds to a special case of our
model with $\bThetaTrue = (\bbeta_1, \bbeta_2, \cdots,
\bbeta_m)^\top$.  In particular, each row of $\bThetaTrue$ is the
parameter of each arm for the multi-arm contextual bandit.
Specifically, we set the number of arms $d_a = \{10, 30, 50\}$ and the
dimension of covariates $d_x = 100$.  For the representation matrix
$\bThetaStar$, we consider both non-sparse and sparse cases.  For the
non-sparse case, we generate $\bThetaStar = \bU \bD \bV^\top$ where
$\bU \in \mathbb{R}^{d_a\times r}, \bV \in \mathbb{R}^{d_x\times r}$
($r=5$), and $\bD$ is a diagonal matrix with $(1, 0.9, 0.9, 0.8, 0.5)$
as the diagonal entries.  All entries of $\bU$ and $\bV$ are first
generated from i.i.d. $N(0,1)$, and then applied Gram–Schmidt to make
each column orthogonal. The matrix $\bU$ is scaled to have $\ell_2$-norm 
$\sqrt{d_a}$ for each column so that the rewards are comparable across different
$d_a$'s.  For the sparse case, each row of $\bThetaTrue$ is set as
zero except for $s_0 = 2$ randomly selected elements that are drawn
from $N(0,1)$.  We generate the covariate $\bx \overset{i.i.d}{\sim}
N(0, \bI_{d_x})$ and the rewards from our model \eqref{eq:reward} with
variance of the reward error~\eqref{eq:epsilon-normal} $\sigma^2 =
0.01$.

We compare \hicoab~against 
\begin{enumerate*} [label=\itshape\alph*\upshape)]
\item the LinUCB \citep{li2010contextual}, which is an extension of
  the traditional Upper Confidence Bound (UCB) algorithm to the
  contextual multi-armed bandit settings;
\item a Lasso Bandit for high-dimensional contextual bandits
  \citep{bastani2020online};
\item NeuralUCB, a neural-network-based method for contextual bandits
  \citep{zhou2020neural} and
\item EE-Net \citep{ban2021ee}, which uses two separate neural
  networks for exploration and exploitation.
\end {enumerate*} 
Details of the tuning parameters of each algorithm are provided in
Appendix~\ref{app:sim-imp}.

\Cref{fig:sim} shows the regret averaged over 50 simulations. For the
non-sparse case in the upper row, \hicoab~converges faster than all
other methods.  The advantage of \hicoab~is more pronounced when the
dimension of arms becomes larger.  This phenomenon demonstrates the
advantages of leveraging the low-rank structure, especially when the
dimension is high.  It is expected that when the dimensions of the
action and covariates continue to grow, the gap between \hicoab~and
the alternatives will further enlarge.  For the sparse case in the
lower row---a setting not to the advantage of \hicoab---when the dimension
of arms is relatively small ($d_a=10$), Lasso Bandit converges faster
but the margin between \hicoab~and Lasso Bandit is small.  As the
number of arms increases, \hicoab~outperforms all other methods.  This
phenomenon comes as a surprise since in this sparse setup, the matrix
$\bThetaStar$ is close to full rank.  This surprising result may be explained
as follows.  As the number of arms increases, we observe that the top
singular vectors explain most variability and thus approximate
$\bThetaStar$ well enough.  In particular, when $d_a = 10$, the top
20\% singular vectors account for around 40\% of the variance, while
when $d_a = 50$, they account for almost 60\%.  For the top 50\%
singular vectors, the percentage of variance explained is around 70\%
when $d_a = 10$, while it is almost 90\% when $d_a = 50$.  In
addition, \hicoab~is adaptive since it does not require prior
knowledge on the rank of $\bThetaTrue$ and it penalizes more when $t$
is small and the penalization gradually decreases as $t$ increases.
In sum, our penalized estimator is robust for such sparse settings,
especially for high dimensional settings.

\subsection{Case Study I: Instant Noodle Company}
\label{Subsec-mk}

In this section, we revisit the instant noodle case study first
introduced in~\Cref{Subsec-example}.  We begin with a more detailed
descriptions of the data and experiment setup and results.  Our method
can provide assortment and pricing decisions simultaneously, which
increases the cumulative sales by a factor of $4$ times, and provide
insightful interpretations of customer behavior through the latent
factors of the estimated representation matrix $\bThetaHat$.

\paragraph{Data Description.}
The original data contains daily sales of $N = 176$ products (SKUs)
across $369$ cities over the time period from March 1st, 2021 to May 31st,
2022, for a total of $T = 456$ days.  The sales are split across 31
Chinese provinces.  Each product consists of either a single flavor of
noodles (13 possible choices), or an assortment of flavors with
varying counts/flavor.  The assortment and price of each product
change daily.  The assortment and prices were the same across
locations.  The maximum number of products to be shown on the homepage
is $K=30$.  Thus, the total possible of combinations is ${N \choose K}
= {176 \choose 30}$; if each such combination is associated with an
action, the action space is extremely high-dimensional, and standard
multi-arm bandit algorithms are computationally prohibitive.


\paragraph{Experiment Setup and Results.}

To apply \hicoab, we specify the action vectors $\ba_t$ and the
covariate vectors $\{\bx_{t,\ell}\}_{\ell=1}^{L}$ with $L = 31$ at
given time $t$ following the setup in~\Cref{Subsec-example}.  The
action vector takes the form
\begin{align*}
\ba = (\tbp_1,\tbp_1^2, \price_1,p_1^2, \cdots,
\tbp_{K}, \tbp_{K}^2,p_{K},p_{K}^2,1) \in
\real^{2(m+1)K+1=841},
\end{align*}
where $\tbp_k=(f_{k,1}, \cdots, f_{k,m})$ is a vector of non-negative
integers to denote the counts of $m=13$ flavors, $p_k$ is the price,
and $\tbp_k^2$ denotes the vector formed by squaring each component of
$\tbp_k$.  The context vector $\bx_{\ell} \in \mathbb{R}^{50}$ for
location $\ell$ includes dummy variables of 31 provinces, the year
2021/2022, 12 months, weekdays, and an indicator of the annual sales
event on Jun 18 and Nov 11. See Appendix~\ref{app:sim} for complete
details.

In order to run simulations using the dataset, we first create a
pseudo-ground-truth model by estimating $\bThetaTrue$ and the variance
of the reward error $\sigma^2$ as in
equation~\eqref{eq:epsilon-normal} using the full dataset.  The pair
$(\bThetaTrue, \sigma)$ define what we refer to as the
\emph{pseudo-ground-truth} for the problem.

Using this fitted model, we assess the validity of bilinear reward
assumption~\eqref{eq:reward} via out-of-sample sales prediction.  In
particular, we do so via a leave-one-out (LOO) approach: recursively
over the index $i$, we compute an estimate $\bThetaHat$ of the true
representation matrix with the $i^{th}$ sample removed, and then use
this fit to predict the $i^{th}$ reward.  We then measure the
performance of these LOO predictions relative to the actual rewards;
doing so yields an out-of-sample prediction error rate of approximately
7\%.  See Appendix~\ref{app:sim} for details.

Initializing \hicoab~with the initial step number $\tinit = 100$ and
$\lambda_0$ according to~\Cref{algo:high-arm}, we then run $100$
trials, in each of which we iterate from $\tinit = 100$ to $T = 456$.
At each iteration $t$, we first estimate $\hTheta_t$ and make an
assortment-pricing decision $\ba_{t+1}$ that maximizes the total sales
given the covariate $\bx_{t+1}$ according to~\Cref{algo:high-arm}, and
then generate a reward based on the pseudo-ground-truth model with
$(\bThetaTrue, \sigma)$.

We evaluate the performance of \hicoab~in terms of the time-averaged
regret~\eqref{eq:regret} and the percentage gain of the cumulative
sales by comparing with the original actions, since no existing bandit
algorithms are applicable to this joint assortment-pricing problem
with contextual information.

\begin{figure}
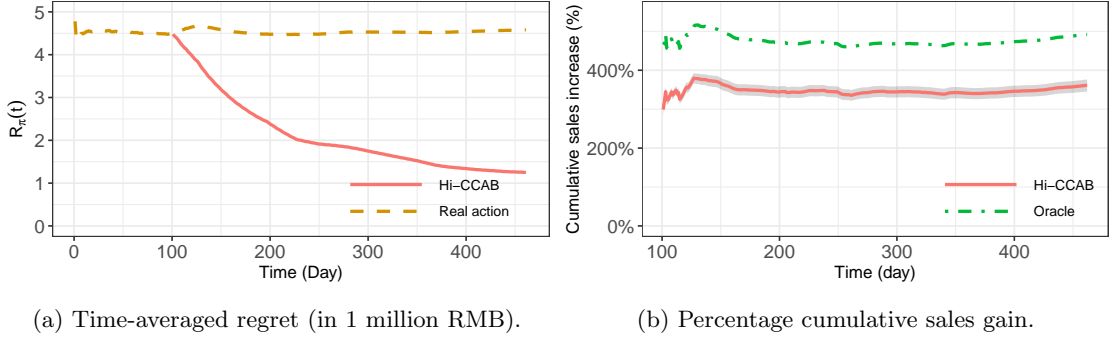

\centering
\begin{subfigure}{.45\textwidth}
  \centering
  \widgraph{\linewidth}{\figdir/fig_time_avg_regret_with_main_nopromo}
  \caption{Time-averaged regret (in 1 million RMB).}
  \label{fig:regret}
\end{subfigure}%
\begin{subfigure}{.45\textwidth}
\begin{center}
  \widgraph{\linewidth}{\figdir/fig_cum_sales_increase_main_band_nopromo}
  \caption{Percentage cumulative sales gain.}
  \label{fig:sales-increase}
\end{center}
\end{subfigure}
\caption{Performance of \hicoab~compared with real actions in terms of the time-averaged regret and the percentage cumulative gain in revenue. The boundaries of the shadow are the 5th and 95th quantiles. }
\label{fig:hicoab-perf}
\end{figure}

\Cref{fig:regret} shows the time-averaged regret and
\Cref{fig:sales-increase} shows the percentage gain in cumulative
sales compared to the real sales (averaged over 100 simulations). The
expected average regret of \hicoab~converges to zero while that of
original actions remains flat.  In terms of percentage gain in
cumulative sales, \hicoab~boosts cumulative sales by almost 4
times. The cumulative sales by \hicoab~is also converging to the
clairvoyant sales obtained by the oracle solution.

\paragraph{Interpretation of the Representation Matrix $\bThetaStar$.}
One advantage of our model is the interpretability which allows us to
gain insights from the representation matrix $\bThetaStar$.
Specifically, our model is able to discover the underlying factors of
the effect of arm-covariate pairs on the reward.  In the following, we
examine the pseudo ground truth $\bThetaStar$ we obtained using all
the data.

The rank of $\bThetaTrue$ is 4 with the singular values being $(1.9,
0.2, 0.02, 0.00003)$. The leading singular value dominates the rest
and thus the leading singular vectors are the most important ones in
explaining the effect on the reward, which we will focus on in what
follows.

\begin{figure}
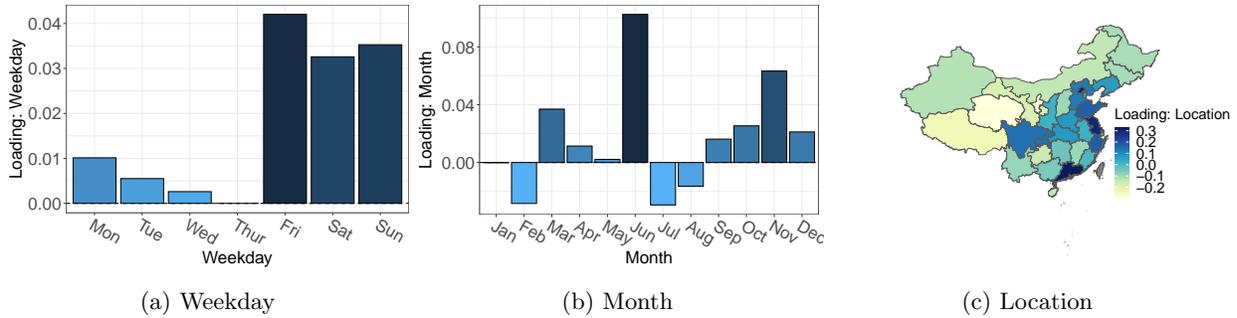

\centering
\begin{subfigure}{.33\textwidth}
  \centering \widgraph{\linewidth}{\figdir/fig_weekday_slides_nopromo}
  \caption{Weekday}
  \label{fig:weekday}
\end{subfigure}%
\begin{subfigure}{.33\textwidth}
  \centering \widgraph{\linewidth}{\figdir/fig_month_slides_nopromo}
  \caption{Month}
  \label{fig:month}
\end{subfigure}%
\begin{subfigure}{.33\textwidth}
  \centering \widgraph{1\linewidth}{\figdir/fig_heatmap_nopromo}
  \caption{Location}
  \label{fig:location}
\end{subfigure}
\caption{Loadings of the leading right singular vectors for the
  covariates.}
\label{fig:v1}
\end{figure}

\Cref{fig:v1} shows the loadings for different covariates (i.e.,
the leading right singular vector) and our algorithm is able to learn
interpretable patterns of the effects on the reward -- for weekdays,
the effects are drastically different during the weekend and during
the weekend; for months, the effects show different patterns during
the promotion month (June and November) from other months; for
location, the effects of the coastal provinces are different from the
rest, which exactly corresponds to the levels of economic development
of different regions in China.  In sum, our model can exploit the
underlying structure of the covariates and provide insights into
purchasing behavior and seasonality.

On the other hand, \Cref{tab:v1-flavor} explores the (scaled) loadings
for the arm on May 29th 2022, the last Sunday in our data (i.e., the
leading left singular vectors multiplied with
$\inprod{\bv_1}{\bar{\bx}}$ where $\bar{\bx}$ is the average of
$\bx_{\ell}$ for $\ell = 1, \ldots, L$ on May 29th
2022). Specifically, we investigate the effect of flavors on the
reward given the context.  We take the average of the loadings of the
linear and quadratic terms for each flavor in all 30 products and
compare with the total sales of each flavor across all Sundays in the
months of May.  For ease of comparison, we further scale the sales and
the loadings by their corresponding largest numbers.  The loadings and
sales are closely related to each other.\footnote{The correlation of
sales and the linear-term loadings is $0.95$ and that of the
quadratic-term loadings is $0.97$.}  As in~\Cref{tab:v1-flavor}, on
May 29th 2022, flavor $1$ (denoted $F1$) has the largest effect,
followed by flavors $13$, $10$, $11$, $7$, and $9$.  Therefore, our
model learns the values of the flavors (per unit).

\begin{table}[!t]
\centering
\begin{tabular}{rrrrrrrrrrrrrr}
  \hline
 & F1 & F2 & F3 & F4 & F5 & F6 & F7 & F8 & F9 & F10 & F11 & F12 & F13 \\ 
  \hline
Sales & 1.00 & 0.05 & 0.00 & 0.00 & 0.00 & 0.03 & 0.19 & 0.00 & 0.08 & 0.19 & 0.18 & 0.00 & 0.38 \\ 
  $\tilde{u}_1$ (linear) & 1.00 & 0.11 & 0.00 & 0.00 & 0.00 & 0.05 & 0.23 & 0.01 & 0.13 & 0.50 & 0.25 & 0.03 & 0.51 \\ 
  $\tilde{u}_1$ (quadratic) & 1.00 & 0.07 & 0.00 & 0.00 & 0.00 & 0.03 & 0.17 & 0.01 & 0.09 & 0.41 & 0.19 & 0.02 & 0.42 \\ 
   \hline
\end{tabular}
\caption{Total sales and loadings of the linear and quadratic terms  (scaled) of the 13 flavors.} 
\label{tab:v1-flavor}
\end{table}


\paragraph{Limitations.}
We would like to point out that the increase in cumulative sales in real life could be less than four times as claimed because there will be additional constraints on the action space due to constraints on production or supply chain.

\subsection{Case Study II: Manicure Start-up}
\label{Subsec-manime}

In this section, we apply our model to a joint assortment-pricing
problem faced by a manicure start-up.  Unlike the noodle company
treated in~\Cref{Subsec-mk}, this start-up company updates its product
line quite frequently, and is interested in determining the color and
style to guide its designs as well as optimal ways to offer discounts
and promotions. Accordingly, we formulate the problem differently by
using the aggregated product features and discount rate as the action
vector, thereby demonstrating the flexibility of our model.  As we
discuss here, our method not only boosts profit, but also provides
insightful interpretations.

\paragraph{Data Description.}

The start-up provided transaction-level data over the period February
1st, 2020 to April 26th, 2021, for a total of $T = 451$ days.  Over
this period, the product line was updated on a regular basis, with a
maximum number of products available online being 74 during October
2020, and a total of 84 products (SKU) avaiable at some point during
the entire time horizon.  Each product can be described by its texture
(glossy versus matted), transparency, and colors (solid or multiple
colors). The store also collaborates with designers and we measure
their popularity by the number of their Instagram followers.  The
price of the products is fixed for designer vs non-designer and the
cost of each product is known. Being a start-up company in a growth
phase, they provide discount promotions on a regular basis to attract
more customers.  For each transaction, the data contains the purchased
product, total price, discount, shipping address, and an indicator of
accepting marketing or not.

\paragraph{Experiment Setup and Results.}

Instead of a specific assortment of manicures, the start-up is most
interested in the trend of colors and styles. At the same time, they
need to decide on the number of available products and designer
products as well as promotions on a daily basis.  Therefore, we use
the daily aggregated product information as the action vector and the
daily aggregated customer information as well as the time as the
covariates.  Specifically, we specify the action vectors $\ba$ as the
count of different colors used in all the manicures (black, white,
gray, red, orange, yellow, green, blue, indigo, and violet), styles
(the proportions of glossy, transparent and designer manicures, and
the total number of Instagram followers of the designers), the
discount rate as well as the quadratic terms of all the above, 
leading to $d_a = 31$. The
covariate vector $\bx$ includes location (percentages of purchase from
Midwest, Northeast, South, and West), demographic proxies (average of
median income and racial distribution by ZIP code), and the
proportions of customers accepting marketing of last period, along
with dummy variables for the 12 months, weekdays and public holidays.
The dimension of the covariate vector is $d_x = 30$.
Given that costs are known, we use profit rather than total sales
revenue as our reward $\reward$.

Similar to our case study described in~\Cref{Subsec-mk}, we first
create a pseudo-truth model by estimating $\bThetaStar$ and $\sigma$
using all data and check our model assumptions in
Appendix~\ref{app:case-study-ii}.  We then run the simulation a total
of $100$ times with initialization $\tinit = 100$.

\begin{figure}
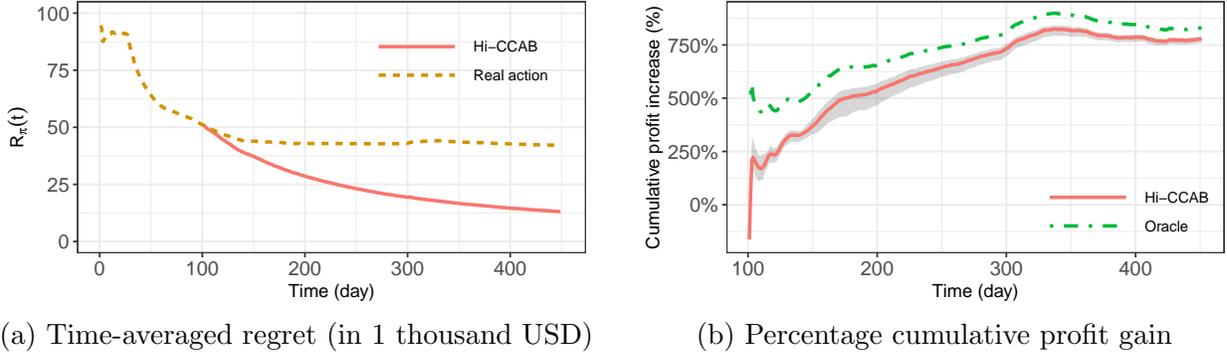

\begin{tabular}{ccc}
  \widgraph{0.48\textwidth}{\figdir/ManiMe/fig_regret} &&
  \widgraph{0.48\textwidth}{\figdir/ManiMe/fig_cum_sales_increase_main_band}
  \\
(a) Time-averaged regret (in 1 thousand USD) && (b) Percentage cumulative profit gain
\end{tabular} \\
\vspace*{0.05in}
\caption{Performance of \hicoab~compared with real actions in terms of
  the time-averaged regret and the percentage cumulative gain in profit. The
  boundaries of the shadow are the $5^{th}$ and $95^{th}$ quantiles. }
\label{fig:manime-hicoab-perf}
\end{figure}

\Cref{fig:manime-hicoab-perf} shows the performance of \hicoab~in
terms of the time-averaged regret and percentage cumulative gain
in profit compared to the real actions (averaged over 100
simulations).  The finding is similar to those from~\Cref{Subsec-mk}:
in particular, the time-averaged regret of \hicoab~converges
to zero, whereas the same quantity associated with the original choice
of actions stays bounded away from zero.  Meanwhile, the
\hicoab~approach boosts the cumulative profit by more than $7$ times.

\paragraph{Interpretation of the Representation Matrix.}

Our procedure learns a representation matrix $\bThetaHat$, and its
singular value decomposition has interesting structure.  It has rank
$5$, which is low compared to its dimensions $31$ by $30$.

\begin{figure}[h]
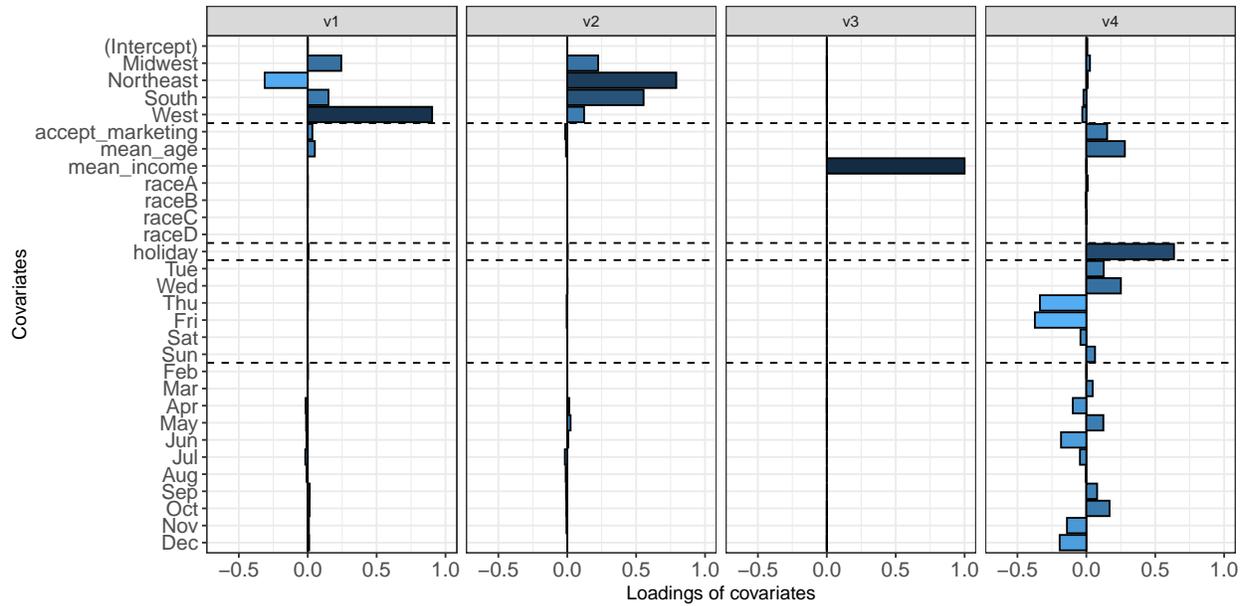

\begin{center}
  \widgraph{\textwidth}{\figdir/ManiMe/fig_right_singular_vector_vertical}
  \end{center}
\caption{Loadings associated with context vectors, as illustrated by
  stem plots of the singular vectors $\bv_1$, $\bv_2$, $\bv_3$ and
  $\bv_4$. Dashed lines separate the covarites into locations,
  customers demographic proxies, a holiday indicator, weekdays, and
  months.}
  \label{fig:manime-covariates}
\end{figure}

\begin{figure}[h]
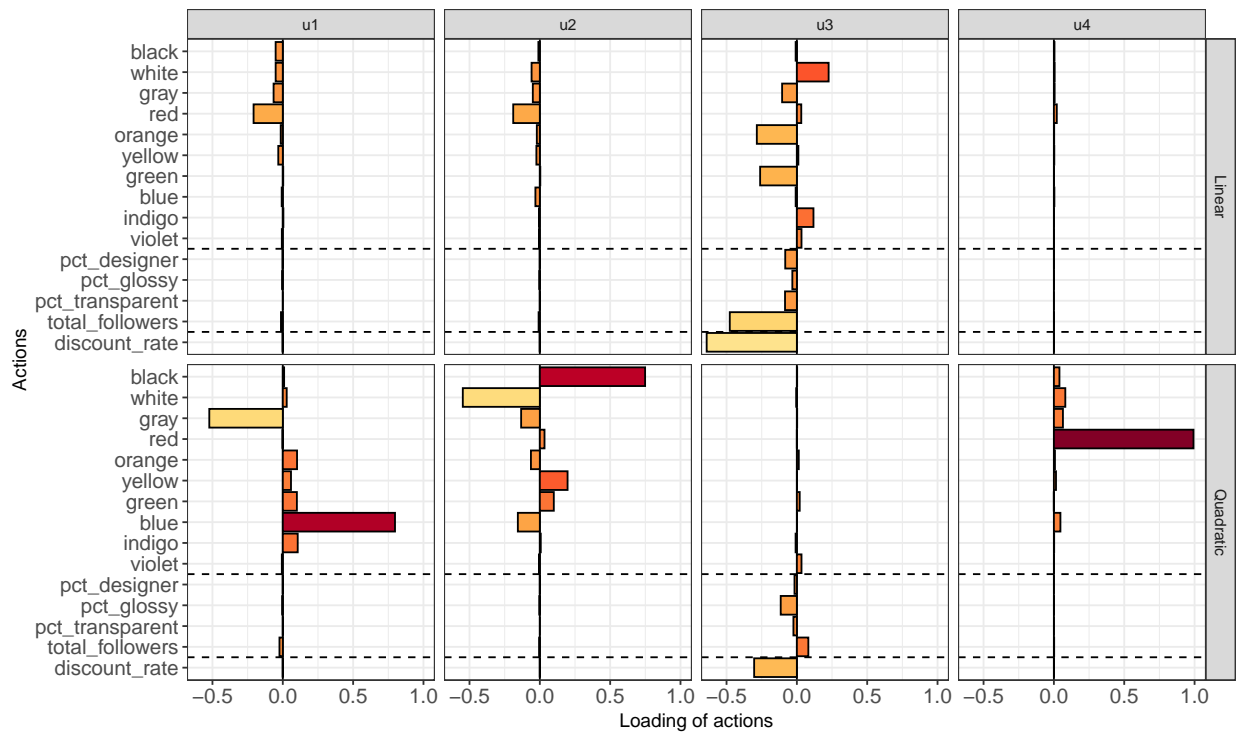

\begin{center}
\widgraph{\linewidth}{\figdir/ManiMe/fig_left_singular_vector_vertical}
\caption{Loadings associated with the action vectors, as illustrated
  by stem plots of the left singular vectors $\bu_1$, $\bu_2$,
  $\bu_3$, and $\bu_4$. Dashed lines separate the actions into colors,
  styles, and discount rate.}
  \label{fig:manime-actions}
    \end{center}
\end{figure}
Its ordered singular values are given by $(0.66, 0.54, 0.22, 0.16,
0.0008)$.  Noting that the last singular value is negligible compared
to the first four, we focus our discussion on the singular vectors
associated with the first four singular values.

Figures~\ref{fig:manime-covariates} and~\ref{fig:manime-actions}
illustrate (respectively) the loadings of the covariates and actions.
At a high level, the four factors capture the Western market $(\bu_1,
\bv_1)$, Northeastern/Southern market $(\bu_2, \bv_2)$, income effect
$(\bu_3, \bv_3)$, and time effect $(\bu_4, \bv_4)$ respectively.  The
covariate loadings associated with the factors are well-separated,
which facilitates interpretations for the actions.  Let us first look
at color and style.  For color, we examine the quadratic terms since
the color action vector represents the absolute count of each color's
appearance in the manicures so that the quadratic terms dominate.
Comparing the Western market ($\bu_1$ and $\bv_1$) and
Northeastern/Southern market ($\bu_2$ and $\bv_2$), we observe
distinct geographical preferences in color choices---blue is more
popular in the Western market while black is preferred in the
Northeastern/Southern market; gray does not sell well in the west
while white is lackluster in the Northeast and South.  As for the
income factor ($\bu_3, \bv_3$), the color preferences are reflected in
the linear terms (since the quadratic terms are negligible) where
white and indigo are more sought-after among higher-income customers
while gray, orange, and green are less favorable.  For the time factor
$(\bu_4, \bv_4)$, red stands out as a festive favorite, aligning with
holiday trends.  In terms of style, their loadings concentrate in the
income factor $(\bu_3, \bv_3)$.  Customers with higher incomes show
less interest in designer and glossy products and care less about the
popularity of the designers.

Next, we investigate the effects of discounts. Loadings of discount
rate are concentrated in the vector $\bu_3$, associated with the
income factor tracked by the pair ($\bu_3, \bv_3$).  Generally, for 
high-income customers, higher discount rates yield lower profits.
One plausible explanation goes as follows: the demand
function~\eqref{eq:ban-demand} is controlled by the market size (via
the linear term) and the price sensitivity (via the quadratic term),
both of which depend on income.  In our case, the market size in the
linear term dominates due to relatively small discount rates (median:
$0.067$; 5th and 95th quantiles: $0.03$ and $0.18$) and the fact that
the linear term in $\bu_3$ outweighs the quadratic term. Therefore,
the negative linear term suggests that profits will be lower with higher
discount rates for our customer base whose household income is of mid-to-high levels
(ranging from 95K to 110K).

As income increases, so does the market size, particularly for hedonic
purchases such as manicures among our customer base.
Consequently, discounts would have incurred more profit loss for higher-income customers.
On the other hand, it is
crucial to note that as a start-up, customer expansion and retention
are vital for long-term growth, in which context discounts can serve
as effective incentives. This unique additional dynamics of a startup, however, can only be revealed by a longer sequence of data. That being said, this longer-term aspect of decision-making,
while beyond the scope of our current case study, is worthy of further
investigation.


\section{Conclusions and Future Directions}
\label{Sec-Conclusion}

The growing need for online and data-driven decision-making has led to
increased interested in bandit models among both theorists and
practitioners.  Nonetheless, at least to date, 
we are not aware of any work on contextual bandits in which both the
covariate and action spaces are high-dimensional.  Our work in this
paper is motivated by applications of bandits---among them the joint
assortment-pricing problem---that have this ``doubly''
high-dimensional nature.  We proposed a structured bilinear matrix
model for capturing interactions between covariates and actions in
determing the reward function. This model is reasonably general,
including a number of structured bandit and pricing models as special
cases; at the same time, it is also highly interpretable in that the
spectral structure of representation matrix captures interactions
between actions and contexts.

We proposed an efficient algorithm \hicoab~that interleaves steps of
low-rank matrix estimation with exploration/exploitation, and we
proved a non-asymptotic upper bound on its time-averaged regret.
In addition, the generality and flexibility of our model enable its application to
the joint assortment-pricing problem, each of which has been studied
extensively in operations research and marketing but not jointly. In
real case studies with the largest instant noodle producers and a
manicure start-up, our method can boost sales/profits, while also
provide insights into the underlying structure of the effect on the
reward of arms and covariates such as purchasing behaviors.

We conclude by discussing some future research directions.  One key
contribution of this paper is our novel model for doubly
high-dimensional contextual bandits and its application to the joint
assortment-pricing problem.  We provided a non-asymptotic upper bound
on the regret while future work could aim to improve this regret
analysis by tightening the regret bound given the outstanding
performance of \hicoab~compared to other bandit algorithms, and
establishing a potentially matching lower bound. Another direction is on its generality in terms of mathematical expressiveness, i.e., when the choice of covariate vector and action vector are particular feature maps as explained in Appendix~\ref{sec:generalizability_relationship}, with a particular focus on theory.  

In terms of applications, the flexibility of model allows for
applications to other multiple decision-making problems in diverse
sectors, such as other business settings and healthcare.  In the realm
of business, our model can incorporate other quantifiable actions for
joint decision-making, particularly in marketing and
operations. Furthermore, it offers flexibility in the objective,
allowing it to be tailored to suit different outcomes, such as social
benefits for social enterprises and the Environmental, Social, and
Governance (ESG) performance for responsible investment.  In
healthcare, especially personalized healthcare, our model holds high
potential. For instance, health monitoring, which provides action
suggestions based on individual health conditions, aligns with our
model. Actions like sleeping patterns, exercise regimens, social media
usage, and dietary choices can be considered in high-dimensional and
continuous arms.  Meanwhile, health outcome also depends on contextual
variables such as age, gender, weight, height, basic health status,
and compliance level.  Traditional bandit models do not suffice for
such doubly high-dimensional contextual settings, but our bilinear
model with a low-rank $\bThetaTrue$ for mean reward fits well, as
health effects of actions and user
characteristics can typically be captured by a few latent factors.

Assessing the performance of our algorithms in other applications,
particularly within the field of personalized healthcare, would be
practically valuable.  Finally, our case studies partially relied on simulations
to evaluate the efficacy of our method; however, real operational
settings often impose additional constraints on the action space.  In
order to gauge and improve the real-world performance of our methods,
we anticipate further collaboration with companies in carrying out
live deployments.




\pagebreak
\bibliography{biblio.bib}
\bibliographystyle{apalike}



\ECSwitch



\ECHead{Supplementary material}

\section{Nonlinear Relationship Representations and Extensions to Reproducing Kernel Hilbert Spaces}
\label{sec:generalizability_relationship}

In this section, we describe how our bilinear form can capture
complicated non-linear effects in the actions and covariates.  Suppose
that the original (primitive) action vector is given by $\pa = (a_1,
a_2, \ldots, a_{d_1}) \in \Real^{d_1}$.  We then define an augmented
action vector by including all polynomial terms of $\pa$ up to order
$N_1$---that is
\begin{multline}
\label{eq:ba_poly}
\ba \defn (1, a_1, a_2, \ldots, a_{d_1}, a_1^2, a_2^2, \ldots,
a_{d_1}^2, a_1 a_2, a_1 a_3, \ldots, a_1 a_{d_1}, a_2 a_3, \ldots, a_2
a_{d_1},\ldots, a_{d_1-1} a_{d_1}, \ldots \\ \ldots, a_1^{N_1},\cdots,
a_{d_1}^{N_1}, a_1^{N_1-1}a_2, a_1^{N_1-1}a_3,\ldots, a_1^{N_1-1}a_{d_1},
\ldots, a_{d_1}^{N_1-1}a_{d_1-1}, \ldots, a_{d_1 - N_1 +1} a_{d_1 -N_1
  +2}\cdots a_{d_1} ).
\end{multline}
Similarly, let $\px \in \R^{d_2}$ denote the original (primitive)
context vector.  We then define an extended context vector $\bx$ by
including all the polynomial terms of $\px$ up to order $N_2$.

With these definitions, the expected reward takes the form
\begin{align}
\label{eq:reward_poly_approx}
\ba^\top\bTheta \bx = \sum_{\substack{k_i, \ell _j\ge 0\\ k_1 +
    k_2+\ldots +k_{d_1} \leq N_1,\\ \ell_1 +\ell _2 + \ldots +
    \ell_{d_2} \leq N_2}} h(k_1,k_2,\ldots,k_{N_1}, \ldots, \ell_1,
\ell_2,\ldots, \ell_{N_2}) \cdot a_1^{k_1}a_2^{k_2} \cdots
a_{d_1}^{k_{d_1}} \cdot x_1^{\ell_1} x_2^{\ell_2} \cdots
x_{d_2}^{\ell_{d_2}},
\end{align}
where $h(k_1,k_2,\ldots,k_{N_1},\ldots, \ell_1, \ell_2, \ldots,
\ell_{N_2})$ is an element in $\bTheta$.

It is evident that the function $g(\pa, \px) = \ba^\top\bTheta\bx$ as
in~\Cref{eq:reward_poly_approx} can capture nonlinear relationships in
terms of $(\pa,\px)$. Furthermore, by suitably choosing the orders
$(N_1, N_2)$ of lifting, such a function $g$ can be used to
approximate any continuous functions $(\pa, \px) \mapsto f(\pa,\px)$
to arbitrary accuracy on any compact set $\mathcal{C}$.  (E.g., see
the paper~\cite{foupouagnigni2020multivariate} for results on the
approximation (multivariate) functions with Bernstein polynomials).
Note that besides polynomial type basis~\eqref{eq:ba_poly} that yields
the form~\eqref{eq:reward_poly_approx}, we can also use other bases
such as Fourier or Haar.

Sometimes we may want to use the basis in an RKHS defined with kernel
$\kernel_{\px}(\cdot, \cdot)$: $\phi_1(\cdot),\phi_2(\cdot),
\phi_3(\cdot), \ldots$, for covariates, and the basis associated with
kernel $\kernel_{\pa}(\cdot,\cdot)$: $\psi_1(\cdot), \psi_2(\cdot),
\psi_3(\cdot), \ldots,$ for actions. The reasons for this choice are
numerous, including the vanishing contribution to the reward of large
primitive covariates (or primitive action), the need for transforming
the domain of the primitive covariates (or primitive action), etc. One
difficulty in using the basis in RKHS is that we usually are not able
to write out the eigenfunctions by the order of eigenvalues (see
Chapter 12 in~\cite{wainwright2019high} for basic properties
of RKHS). But this difficulty can be dealt by mapping $\pa$ to a
lifted vector of the form
\begin{align}
\px \mapsto \bx \defn (\kernel_{\px}( \cx_1 , \px), \kernel_{\px}(
\cx_2 , \px), \kernel_{\px}( \cx_3 , \px), \cdots, \kernel_{\px}(
\cx_{N_x} , \px)),
\end{align}
where $\cx_1, \cx_2, \ldots, \cx_{N_x}$ are ``characteristic indices''
satisfying 1. $\kernel(\cx_i,\cx_j) = 0$ for all $i \neq j$, and 2. $
P_{\px}(\kernel(\cx_i, \px) \neq 0) >0 $ for all $i$. In this way, we
include the basis that form the relevant space and the matrix
$\bTheta$ can learn the right structure.

Lifting the space of $\ba$ and $\bx$, either by explicitly expanding
the vector through basis or by using RKHS, significantly increases the
expressiveness and flexibility of our model. Despite the increasing
complexity in space, the learning task does not necessarily become
more complex thanks to our low-rank assumptions. Efficient algorithms
designed for low-rank structures can easily handle the increasing
space complexity.


\newcommand{\DtermA}{\ensuremath{D^A_T}}
\newcommand{\DtermB}{\ensuremath{D^B_T}}
\newcommand{\Term}{\ensuremath{Q}}
\newcommand{\wtail}{\ensuremath{s}}

\section{Technical Details for Proofs}
\label{app:proof}

In this section, we first provide the full technical details of the
proofs that were deferred from the main body.  We begin with details
for the proof of~\Cref{prop:theta_bound}, which bounds the error in
estimating the representation matrix $\bThetaTrue$. We then turn to
the proof of~\Cref{thm:main}. The two lemmas involved in the proof
of~\Cref{prop:theta_bound} stated in the main text are proved
in~\Cref{proof:lem:nabla} and~\Cref{proof:lem:main}.


\subsection{Proof of \Cref{prop:theta_bound}}
\label{app:proof-prop1}

Consider the singular value decomposition $\bThetaStar = \bU \bS
\bV^\top$, where $\bS$ is an $r \times r$ diagonal matrix.  Let
$\bU_{\perp} $ be an $d_a \times ({d_a -r) }$ matrix satisfying
$(\bU,\bU_{\perp})(\bU,\bU_{\perp})^\top = \bI_{d_a}$, and define the
matrix $\bV_{\perp}$ in an analogous manner.

Introducing the shorthand notation $\dTheta_{t,\perp} \defn
\bU_{\perp} \bU_{\perp}^{\top} \dTheta_t \bV_{\perp}\bV_{\perp}^\top$,
this construction ensures that $\mynucnorm{ \bThetaStar +
  \dTheta_{t,\perp}} = \mynucnorm{\bThetaStar} +
\mynucnorm{\dTheta_{t, \perp}}$.  This fact, along with some
applications of the triangle inequality, yields the lower bound
\begin{align*}
\mynucnorm{\bThetaStar + \dTheta_t} & \geq  \mynucnorm{ \bThetaStar +
  \dTheta_{t,\perp}} - \mynucnorm{ \dTheta_t - \dTheta_{t,\perp}}
\notag \\
& = \mynucnorm{\bThetaStar} + \mynucnorm{\dTheta_{t,\perp}} -
\mynucnorm{ \dTheta_t -\dTheta_{t,\perp} } \\
& \geq  \mynucnorm{\bThetaStar} + \mynucnorm{\dTheta_{t,\perp}} -
\sqrt{2r} \fronorm{\dTheta_t -\dTheta_{t,\perp}},
\end{align*}
where the final inequality follows from the fact that the matrix
$\dTheta_t -\dTheta_{t,\perp}$ has rank at most $2 r$.

Combining this inequality with~\Cref{eq:first} yields the upper bound
\begin{align}
  \label{eq:second}
\ErrTerm_t(\dTheta_t) & \leq \big| \tracer{\nabla
  \LL[t]{\bThetaStar}}{\dTheta_t} \big| + \lambda_t \left( \sqrt{2r}
\fronorm{\dTheta_t -\dTheta_{t,\perp}} - \mynucnorm{\dTheta_{t,\perp}}
\right).
\end{align}
Next, we apply the results of~\Cref{lem:main} and~\Cref{lem:nabla}
to~\Cref{eq:second}.  Doing so guarantees that, with probability at
least $1 -{4\over t} -{1\over t^2} -{3\over Lt} - {2\over L^3t^3} -
{1\over t^2L}$, we have
\begin{equation}
\label[ineq]{eq:long_ineq}
\begin{split}
  &\left( { \floor{t^{2\over 3}} \over 2t} h^2 - \simplifykappa{t} \right) \fronorm{\dTheta_t}^2 \\
\leq & \fronorm{\dTheta_t} { \sigma (6+ 3\sqrt{t}) ( \sqrt{d_x\log{(tL)}}+2 \log(tL) )
  \over t} + \\
& \Bigg( { 8h\sigma \over t }\sqrt{\log(tL)} \sqrt{(d_x+3\log(tL))( d_a
  + 3 \log(t) )}( \log(d_x+d_a) +2 \log(t) ) \\
& \quad + 2 h \sigma t^{-2/3} \log(t) \sqrt{ \max\{d_a,d_x\} \log(d_a+d_x)
  \over L} \Bigg) \left( \mynucnorm{\dTheta_t - \dTheta_{t,\perp}} +
\mynucnorm{\dTheta_{t,\perp}} \right) \\
& + \lambda_0 {\sqrt{t}\over t} \sqrt{2r} \fronorm{\dTheta_t} -
\lambda_0 \frac{\sqrt{t}}{t} \mynucnorm{\dTheta_{t,\perp}} .
\end{split}
\end{equation}

Next, we make the following updates to the
\Cref{eq:long_ineq}: we apply the bound
$\nucnorm{\dTheta_t -\dTheta_{t,\perp}} \leq \sqrt{2r}
\fronorm{\dTheta_t - \dTheta_{t,\perp}}$, divide both sides by
$\fronorm{\dTheta_t}$, and multiply both sides with $ 3t^{1\over
  3}/h^2$.  Suppose $\burnin$ is the smallest integer such that for
all $t\geq  \burnin$, the following inequalities hold
\begin{subequations}
\label[ineq]{neq:T1}  
\begin{align}
\label{EqnA}
t \geq  & 8 , \\
\label{EqnB}
t^{1\over 3} \geq  & 120 \, \big( 1 +\frac{1}{h} \big) \Big(
(d_a+d_x)(1+\frac{1}{2}\log{\log{t}}) + 6\log{t} + 3\log{L} \Big)^2
\log{t} \\
\label{EqnC}
\begin{split}
   \lambda_0 \geq  & \frac{8 h \sigma}{t^{1\over 2}} \sqrt{\log(t
     L)}\sqrt{(d_x+ 3 \log(tL))( d_a + 3 \log(t) )}( \log(d_x + d_a)
   +2 \log(t) ) + \\
  &\qquad 2 h \sigma t^{-1/6} \log(t) \sqrt{ \frac{\max\{d_a, d_x\}
       \log(d_a + d_x)}{ L}}.
\end{split}
\end{align}
\end{subequations}
Clearly, the burn-in period $\burnin$ is well-defined because there is
a constant $C_{h,L,\lambda_0}$ depending on $L, h$ and $\lambda_0$
such that for any
\begin{align*}
  t \geq C_{h,L,\lambda_0} (d_x+d_a)^6 \left(\log (d_x + d_a)\right)^4,
\end{align*}
the \Cref{neq:T1} hold. Therefore, we have the upper bound
\begin{align*}
\burnin\leq C_{h,L,\lambda_0} (d_x+d_a)^6 \left(\log(d_x+d_a)\right)^4.
\end{align*}
Then we have for $t\geq \burnin$,
\begin{align}
\label{eq:Theta_est_precision}
\fronorm{\hTheta_t - \bThetaStar} & = \fronorm{\dTheta_t} \leq {
  9t^{1\over 3}(2+ \sqrt{t}) ( \sqrt{d_x \log{(tL)} }+2\log(tL) )\sigma \over th^2} +
6\lambda_0 {\sqrt{2r} \over h^2 t^{1\over 6} }.
\end{align}


\subsection{Proof of~\Cref{thm:main}}
 \label{proof:thm}

Let the oracle optimal action at time $t$ be $\ba^*_t$ and $\bb_t =
\sum_{\ell=1}^L \bx_{t,\ell}$. We can decompose the cumulative regret
$T\Regret{T}{\pi}$ as the sum $B_T + D_T$, where we define: (i)  the
cumulative regret until time $\burnin$
  \begin{align*}
 B_{T} & \defn\E\left( \sum_{t=0}^{\min(\burnin-1,
   T-1)}\sum_{\ell=1}^L \left( \ba_{t+1}^{*\top} \bThetaStar
 \bx_{t+1,l} - \ba_{t+1}^\top\bThetaStar \bx_{t+1,l} \right) \right),
  \end{align*}
and (ii) the cumulative regret after the time $\burnin$:
\begin{align*}
  D_T & \defn \E\left( \sum_{t=\min(\burnin,T)}^{T-1}\sum_{\ell=1}^L
  \left( \ba_{t+1}^{*\top} \bThetaStar \bx_{t+1,l} -
  \ba_{t+1}^\top\bThetaStar \bx_{t+1,l} \right) \right) .
\end{align*}
When $\burnin \geq T$, $D_T = 0$ as it does not include any terms in
the summation. We use the same convention for all the summations
introduced later: when the index of the beginning of the summation is
larger than that of the end, then the summation is set to 0.

We focus on each of these two terms in turn.

\subsubsection{Bounding $D_T$:}
We begin by bounding $D_T$.  In order to do so, we let $\Event_t$
denote the event that~\Cref{eq:Theta_est_precision} holds, and let
$\Event_t^c$ be its complement.  By~\Cref{prop:theta_bound}, we have
the bound
\begin{align*}
\Prob(\Event_t^c)\leq {4\over t} +{1\over t^2} + {3\over Lt} + {2\over
  L^3t^3} + \frac{1}{Lt^2} \qquad \mbox{for all $t \geq  \burnin$,}
\end{align*}
and moreover, the event $\Event_t^c$ and estimate $\hTheta_{t}$
are jointly independent of the random vector $\bb_{t + 1}$.

Now by definition, we have
\begin{align}
\label{eq:a_star}
\ba_{t+1}^{*\top} =\frac{ \bThetaStar\bb_{t+1} }{
  \opnorm{\bThetaStar\bb_{t+1}} }, \quad \mbox{and} \quad
\E\left(\ba_{t+1}^{\top} \mid \{\bx_{i,\ell}\}_{j\leq t+1, \ell\leq L},
\{\ba_{i}\}_{i\leq t}, \{ y_{i,\ell} \}_{i\leq t,\ell\leq L} \right) =
\frac{ \hTheta_t \bb_{t+1} }{ \opnorm{\hTheta_t \bb_{t+1}}
}.
\end{align}
Taking the expectation of $D_T$ conditional on
$\{\bx_{t+1,\ell}\}_{\ell=1}^L$, and then substituting the two
relations from \Cref{eq:a_star} yields
\begin{align*}
  D_T = \E \left( \sum_{t=\min(\burnin,T)}^{T-1} \inprod{
    {\bThetaStar\bb_{t+1} \over \opnorm{\bThetaStar\bb_{t+1}} } -{
      \hTheta_{t}\bb_{t+1} \over \opnorm{
        \hTheta_{t}\bb_{t+1}}}} {\bThetaStar \bb_{t+1}}\right),
\end{align*}
where we recall that $D_T = 0$ when $\burnin \geq  T$.

Next, we use the pair of events $\Event_t$ and $\Event_t^c$ to write
the decomposition $D_T = \DtermA + \DtermB$, where
\begin{align*}  
  \DtermA & \defn \sum_{t=\burnin \wedge T}^{T-1} \E\left( \inprod{
    \frac{\bThetaStar \bb_{t+1}}{\opnorm{\bThetaStar\bb_{t+1}}} -
    \frac{\hTheta_{t}\bb_{t+1}}{\|\hTheta_{t}\bb_{t+1}
      \|_2}} {\bThetaStar \bb_{t+1}}\mathbbm{1} \{
  \Event_{t}\}\right), \quad \mbox{and} \\
\DtermB & \defn \E\left( \inprod{
  \frac{\bThetaStar\bb_{t+1}}{\|\bThetaStar\bb_{t+1}\|_2} -
  \frac{\hTheta_{t}\bb_{t+1}}{\|\hTheta_{t}\bb_{t+1}
    \|_2}}{\bThetaStar \bb_{t+1}} \mathbbm{1} \{ \Event_{t}^c \}
\right).
\end{align*}
We analyze each of these two terms in turn.

\paragraph{Analysis of $\DtermA$:}
By adding and subtracting terms, we can write
\begin{align*}
\DtermA & = \sum_{t=(\burnin\wedge T)}^{T-1} \Bigg( \E\left( \left
\langle \frac{ (\bThetaStar - \hTheta_{t})\bb_{t+1} }{
  \|\bThetaStar\bb_{t+1} \|_2 } +
\frac{\|\hTheta_{t}\bb_{t+1}\|_2 - \|\bThetaStar\bb_{t+1}\|_2 }{
  \|\hTheta_{t}\bb_{t+1} \|_2 \|\bThetaStar\bb_{t+1}\|_2}
\hTheta_{t}\bb_{t+1} , \bThetaStar \bb_{t+1} \right \rangle
\mathbbm{1}\{\Event_{t}\} \right) \\ & \leq \sum_{t=(\burnin\wedge
  T)}^{T-1} 2\E\left( \opnorm{\dTheta_t} \|\bb_{t+1}\|
\mathbbm{1}\{\Event_t\} \right)\\
& \leq \sum_{t=(\burnin\wedge T)}^{T-1} 2\E\left( \fronorm{\dTheta_t}
\|\bb_{t+1}\| \mathbbm{1}\{\Event_t\} \right) \\
& \leq 2 \sum_{t=(\burnin \wedge T)}^{T-1} \left( {
  9t^{1\over 3}(2+ \sqrt{t}) ( \sqrt{d_x\log{(tL)}}+2\log(tL) )\sigma \over th^2} +
6\lambda_0 {\sqrt{2r} \over h^2 t^{1\over 6}} \right) \sqrt{Ld_x} \\
& \leq T^{5\over 6} \log(T) \left(1 + ( \sqrt{d_x \log{TL}}+2\log(L))(\frac{1+ 5
  T^{-1/2}}{2\log(T)}) + 5 \frac{1}{\sqrt{T}}\right)\sqrt{Ld_x}
\frac{72\sigma}{5h^2} + \frac{108}{5}\lambda_0
\frac{\sqrt{2rd_xL}}{h^2}T^{\frac{5}{6}} .
\end{align*}


\paragraph{Analysis of $\DtermB$:}
We have
\begin{align*}
\DtermB & \stackrel{(i)}{\le} \sum_{t=\burnin \wedge T}^{T-1} 2
\E\left( \| \bThetaStar\bb_{t+1} \|_2 \mathbbm{1} \{ \Event_{t}^c \}
\right) \stackrel{(ii)}{ = } \sum_{t=\burnin \wedge T}^{T-1} 2
\E\left( \| \bThetaStar\bb_{t+1} \|_2 \right) \E\left( \mathbbm{1} \{
\Event_{t}^c \} \right)\\ & \stackrel{(iii)}{ \leq } \sum_{t=\burnin
  \wedge T}^{T-1} 2 \opnorm{\bThetaStar} \sqrt{\E\left( \|
  \bb_{t+1}\|^2 \right)} \left( {4\over t} +{4\over t^2} + {3\over Lt}
+ {2\over L^3t^3} + \frac{1}{Lt^2}\right) \\
& \stackrel{(iv)}{ < } 8 \opnorm{\bThetaStar} \sqrt{L d_x} \log(T),
\end{align*}
where step (i) follows from $\big\| \frac{ \bThetaStar \bb_{t+1} }{ \|
  \bThetaStar \bb_{t+1} \|_2 } -\frac{ \hTheta_{t}\bb_{t+1} }{
  \|\hTheta_{t}\bb_{t+1} \|_2 } \big\|_2 \leq 2$; step (ii)
follows from $\left(\Event_t^c , \hTheta_{t}\right) \perp \!\!\!
\perp \bb_{t+1}$; step (iii) uses the Cauchy--Schwarz inequality and
$\Prob(\Event_t^c)\leq {4\over t} +{1\over t^2} + {3\over Lt} + {2\over
  L^3t^3} + \frac{1}{Lt^2}$ for $t\geq \burnin$; and step (iv) follows
from elementary calculation.

      \vspace*{0.05in}

\paragraph{Putting together the pieces:}
      
Combining our bounds on $\DtermA$ and $\DtermB$, we find that
{\small
\begin{multline}
    D_T < T^{5\over 6} \log(T) \left( {3\over 2} \left(1 +
    (\sqrt{d_x\log{(TL)}}+2\log(L))(\frac{1+ 5 T^{-1/2}}{2\log(T)}) + 5
    \frac{1}{\sqrt{T}}\right)\sigma\sqrt{d_xL} + \frac{\lambda_0
      \sqrt{2rd_xL}}{\log{T}} \right)\frac{72}{5h^2} \\ + 8
    \opnorm{\bThetaStar} \sqrt{L d_x} \log(T)
\end{multline}
}


\vspace*{0.2in}

\subsubsection{Bounding $B_T$:}
By similar arguments, we can establish the bound
\begin{align*}
B_T = \E \left(\sum_{t=0}^{(\burnin-1)\wedge (T-1)}\sum_{\ell=1}^L \E
\left( \ba_t^{*\top} \bThetaStar \bx_{t,\ell} - \ba_t^\top\bThetaStar
\bx_{t,\ell} \right) \right) \leq \burnin \, \big \{ 2 \sqrt{Ld_x}
\opnorm{\bThetaStar} \big \}.
\end{align*}

Therefore,
\begin{multline}
\Regret{T}{\pi} = \frac{B_T+D_T}{T} \leq {2\sqrt{Ld_x}
  \opnorm{\bThetaStar} \burnin } T^{-1} + 8
\sqrt{Ld_x} \opnorm{\bThetaStar} \frac{\log(T)}{T} \\
+ \frac{\log(T)}{T^{1\over 6}} \left( \left(1 + (\sqrt{d_x\log{TL}} + 2
\log(L))(\frac{1+ 5 T^{-1/2}}{2\log(T)}) + 5
\frac{1}{\sqrt{T}}\right)\sigma\sqrt{Ld_x} + \frac{\lambda_0
  \sqrt{2rd_xL}}{\log(T)} \right)\frac{72}{5h^2}.
\end{multline}
Setting $c_1 = 2$, $c_2 = 8$, $c_3 = \frac{72}{5}$
and $c_4 =\frac{324}{5}$
gives the statement of the theorem.

The choice of $c_4$ is based on
\begin{align*}
    \left(1 + (\sqrt{d_x\log{TL}} + 2
\log(L))(\frac{1+ 5 T^{-1/2}}{2\log(T)}) + 5
\frac{1}{\sqrt{T}}\right) < 5 + 3\sqrt{d_x}(1+\sqrt{\log{L}}) +  7\log{L} . 
\end{align*}


\subsection{Proof of~\Cref{lem:nabla}}
\label{proof:lem:nabla}

Let us prove the claim with $t$ replaced by $T$, so as to allow
ourselves to use $t$ as an index of summation.  Recall that the noisy
reward takes the form $\reward_{t,\ell}=
\ba^T_{t}\bThetaStar\bx_{t,\ell} + \sigma \varepsilon_{t,\ell}$, so
that the loss gradient can be written as
\begin{align*}
\nabla \LL{\bThetaStar} &= {\sigma\over LT}
\sum_{t=1}^T\sum_{\ell=1}^L - \varepsilon_{t,\ell} \bx_{t,\ell}
\ba_{t}^\top  \\
& = \underbrace{ {\sigma\over LT} \sum_{\ell=1}^L -\varepsilon_{1, \ell}
  \bx_{1, \ell}\ba_{1}^\top + {\sigma\over LT}
  \sum_{t=2}^T\sum_{\ell=1}^L -\varepsilon_{t, \ell} \bx_{t, \ell}
  \hat{\ba}_{t}^\top}_{\Smat_2} + \underbrace{ {\sigma\over LT}
  \sum_{t=2}^T\sum_{\ell=1}^L -\varepsilon_{t, \ell} \bx_{t, \ell}
  \bdelta_{t}^\top}_{\Smat_3}.
\end{align*}
From this decomposition, we have the upper bound
\begin{align}
\label{EqnKeyUpper}
\big| \tracer{\nabla \LL{\Theta^*}}{\dTheta} \big | = |
\tracer{\Smat_2}{\dTheta_T} + \tracer{ \Smat_3}{\dTheta} \big| & \leq
\fronorm{\Smat_2} \fronorm{\dTheta}+ \opnorm{\Smat_3}
\mynucnorm{\dTheta}.
\end{align}
Consequently, in order to establish the claim of the lemma, it
suffices to show that the inequalities
\begin{subequations}
\begin{align}
\label{EqnBoundSmatTwo}  
  \fronorm{\Smat_2} & \leq \underbrace{ { \sigma (6+ 3\sqrt{T}) ( \sqrt{d_x\log{(TL)}}
      + 2 \log{T L} ) \over T} }_{\preone{T}} \\
  \begin{split}
    \label{EqnBoundSmatThree}    
  \opnorm{\Smat_3} & \leq 
  \underbracea{ \frac{2h\sigma
      \log(T)}{T^{2/3}} \sqrt{ \max\{d_a,d_x\} \log{(d_a+d_x)} \over
      L}}  \\
     & \quad \underbracebd{ + \frac{8h\sigma}{T} \sqrt{\log{(T
        L)}}\sqrt{(d_x + 3 \log(LT))( d_a + 3 \log(T) )}(
    \log(d_x+d_a) +2 \log(T) )}_{\pretwo{T}}
    \end{split}
\end{align}
\end{subequations}
both hold with the probability claimed in the lemma statement.

In order to prove these bounds, let us recall a basic concentration
inequalities (cf. Lemma 1 in~\citet{laurent2000adaptive}): for a
$\chi^2$-variable $U$ with $k$ degrees of freedom, we have
\begin{align}
\label[ineq]{ineq:chisquare}
\Prob \Big[ U - k \geq 2\sqrt{k\nu} + 2 \nu \Big] & \leq \exp{(-\nu)}
\qquad \mbox{for any $\nu > 0$.}
\end{align}

Define the event
\begin{align*}
\Jevent_{T} & \defn \Big \{ \max_{ \substack{t \in [T] \\ \ell \in
    [L]}} \|\bx_{t, \ell}\|^2_2 \leq d_x + 2\sqrt{ 2 d_x \log(TL)} +
4\log(T L) \} \Big \}.
\end{align*}
Since the random variable $\|\bx_{t, \ell}\|^2_2$ follows a
$\chi^2$-distribution with degree of freedom $d_x$, applying
\Cref{ineq:chisquare} yields
\begin{align*}
\Prob(\Jevent_T) & \geq  1- T \, L \, \frac{1}{T^2L^2} \; = \; 1 -
\frac{1}{TL}.
\end{align*}

\subsubsection{Proof of the bound~\eqref{EqnBoundSmatTwo}:}

We introduce the convenient shorthand
\begin{align*}
  \XB \defn d_x + 2\sqrt{ 2 d_x \log(TL)} + 4 \log(TL),
\end{align*}
and define the sum \mbox{$\W{T} \defn \sum_{t=2}^T -\varepsilon_{t,
    \ell} \bx_{t, \ell} \hat{\ba}_{t}^\top$.}  With this notation, we
have the upper bound
\begin{align}
\label{eq:S2_analysis}
    \fronorm{\Smat_2} & \leq \underbrace{\frac{\sigma}{LT} \sum_{\ell=1}^L
    |\varepsilon_{1,\ell}|\cdot \fronorm{\bx_{1,\ell}}}_{V_1} +
    \underbrace{\frac{\sigma}{L T} \sum_{\ell=1}^L \fronorm{\W{T}}}_{V_2}.
\end{align}
We analyze each of the quantities $V_1$ and $V_2$ in turn.

\paragraph{Analysis of $V_1$:}
Define the event
\begin{align*}
  \JeventTil_T & \defn \big \{ \max_{\ell \in [L]}
  |\varepsilon_{1,\ell} \mid \leq 2\sqrt{\log(TL)} \big \}.
\end{align*}
Conditioned on the event $\Jevent_T \cap \JeventTil_T$, we have the
bound $V_1 \leq \frac{\sigma}{T} 6 (d_x + 2\log (T L))$.  Elementary
calculation shows that $\Prob(\JeventTil_T) \geq  1 - \frac{1}{T^2L}$.

\paragraph{Analysis of $V_2$:}
Denote the history up to and including time $T$ by
\begin{align}
\label{eq:his}
H_T & \defn \big \{ \bx_{t, \ell},\ba_{t, \ell},\reward_{t, \ell} \,
\mid t = 1, \ldots, T, \; \ell = 1, \ldots, L \big \}.
\end{align}
Note tha the noise variables $\big \{ \varepsilon_{T, \ell} \, \mid
\ell = 1, \ldots, L \big \}$ at time $T$ are independent of $H_{T-1}$.
For $\lambda > 0$, elementary calculation shows that the quantity
$\Term \defn \E\left[ \exp{\left(\lambda \fronorm{\W{T}}^2
    \right)}\mathbbm{1}\{\Jevent_T\} \right]$ can be upper bounded as
\begin{align*}
\Term & \leq \E\Bigg[ \E\Big[ \exp{(\lambda \fronorm{\W{T-1}}^2
      -2\lambda\hat{\ba}_{T}^\top \W{T-1}^\top \bx_{T,
        \ell}\varepsilon_{T, \ell} + \lambda\|\bx_{T, \ell}\|_2^2
      \varepsilon_{T, \ell}^2 )} \mathbbm{1}\{\Jevent_T\} \mid
    H_{T-1}, \bx_{T, \ell} \Big] \Bigg] \\
& = \E\left[ \frac{1}{1-2\lambda\|\bx_{T, \ell}\|_2^2}
  \exp{\left(\lambda \fronorm{\W{T-1}}^2 + \frac{(-2\lambda
      \hat{\ba}_{T}^\top \W{T-1}^\top \bx_{T, \ell})^2}{
      2(1-2\lambda\|\bx_{T, \ell}\|_2^2 )} \right)}
  \mathbbm{1}\{\Jevent_T\}\right] \\
& \leq \frac{1}{1-2\lambda \XB} \E\left[ \exp{\left( \lambda
    \fronorm{\W{T-1}}^2 + 2 \lambda^2 \frac{\opnorm{\W{T-1}}^2 \XB}{
      1-2\lambda \XB}\right)}\mathbbm{1}\{\Jevent_T\} \right] \\
& \leq \frac{1}{1-2\lambda\XB} \E\left[ \exp{\left(
    \frac{\lambda}{1-2\lambda \XB} \|\W{T-1}\|_F^2 \right)}
  \mathbbm{1}\{\Jevent_T\} \right].
\end{align*}
Setting $\lambda = \frac{1}{2T\XB}$ and recursively applying the above
arguments, we find that
\begin{align}
  \label{EqnKorea}
 \E \Big[ \exp{\big( \frac{1}{2T\XB} \fronorm{\W{T}}^2\big)}
   \mathbbm{1}\{\Jevent_T\} \Big] & \; \leq \; \prod_{t=2}^{T}
 \Big(\frac{1}{1-1/t} \Big) = T.
\end{align}
Therefore, for any $\wtail > 0$, we have
\begin{align*}
    \Prob \Big[ \fronorm{\W{T}}^2 \mathbbm{1} \{\Jevent_T\} \geq 
      \wtail^2 \Big] & = \Prob \Big[ \exp(\tfrac{1}{2 T \XB}
      \fronorm{\W{T}}^2 \mathbbm{1} \{\Jevent_T\}) \geq  \exp(
      \tfrac{1}{2 T \XB} \wtail^2) \Big] \\
& \leq T \exp( - \frac{1}{2 T \XB} \wtail^2),
\end{align*}
where the last step uses Markov's inequality, along
with the bound~\eqref{EqnKorea}.

We now set $\wtail \defn \sqrt{4T\XB \log(TL)} = \sqrt{4T(d_x +
  2\sqrt{ 2 d_x \log(TL)} + 4\log(TL)) \log(TL)}$, and find that
\begin{align*}
\Prob \Big[ \big \{\fronorm{\W{T}} \leq \wtail \text{ for all $\ell =
    1, \ldots, L$} \big \}^c \cap \Jevent_T \Big] & \leq L \frac{1}{T
  L^2}
\end{align*}
Noting that $\wtail \leq \sqrt{T}(3\sqrt{d_x \log{(TL)}} + 6\log(TL))$, we have that
\begin{align}
\Prob \Big[ \big \{ \fronorm{\Smat_2} > \frac{\sigma}{T}\big( 6 + 3
  \sqrt{T} \big)(\sqrt{d_x\log{(TL)}}+2\log(TL)) \big\} \cap \Jevent_T \Big] & \leq
\Prob(\JeventTil_T^c) + \frac{1}{TL} \notag \\
\label[ineq]{eq:Smat2}
& \leq \frac{1}{TL} + \frac{1}{T^2 L}.
\end{align}

\subsubsection{Proof of the bound~\eqref{EqnBoundSmatThree}:}

Turning to the analysis of $\Smat_3$, consider the inequalities
\begin{align*}
\max_{ \substack{t \in [T] \\ \ell \in [L]}} |\varepsilon_{t, \ell}|
\leq 3\sqrt{\log(TL)}, \quad
 \max_{ \substack{t \in [T] \\ \ell \in [L]}} \|\bx_{t, \ell}\|_2^2
 \leq 2d_x + 6\log(LT) , \quad \mbox{and}  \quad
 \max_{t \in [T]} \|\bdelta_t/h\|_2^2 & \leq 2d_a + 6\log(T).
\end{align*}
and let $\Gevent$ be the event that all three hold simultaneously.  An
elementary calculation shows that
\begin{align}
\label[ineq]{eq:Gevent_prob}
\Prob(\Gevent^c) & \leq \frac{2}{T^3 L^3} + \frac{1}{L T} +
\frac{1}{T},
\end{align}
and moreover, we have the inclusion $\Gevent \subset \Jevent_T$.

Define the following truncated variables:
\begin{align*}
     \tvarepsilon_{t, \ell} = \varepsilon_{t, \ell} \mathbbm{1}\{|\varepsilon_{t, \ell}|\leq  3\sqrt{\log(TL)} \}, & \quad \tbx_{t, \ell} = \bx_{t, \ell} \mathbbm{1}\{\|\bx_{t, \ell}\|_2^2\leq 2d_x + 6\log(LT)\}, \\
    \tbdelta_t = \bdelta_t  \mathbbm{1}\{\|\bdelta_t/h\|_2^2 \leq 2d_a + 6\log(T)\}, &\quad 
    \mbox{and}\quad  \tSmat_3 = {\sigma\over LT}
  \sum_{t=2}^T\sum_{\ell=1}^L -\tvarepsilon_{t, \ell} \tbx_{t, \ell}
  \tbdelta_{t}^\top.
\end{align*}

Clearly, on the event $\Gevent$, we have the equivalence $\Smat_3 =
\tSmat_3$. Therefore, for any $\alpha > 0$, we have that
\begin{align}
\label{eq:tSmat_Smat}
\Prob(\{\opnorm{\Smat_3} \geq  \alpha \}\cap \Gevent) =
\Prob(\opnorm{\Smat_3} \geq  \alpha).
\end{align}

Further, by construction, $\{\tvarepsilon_{t, \ell}, \tbx_{t, \ell} ,
\tbdelta_t \}_{t,\ell}$ are independent random variables. Applying a
matrix Bernstein Inequality~(Theorem 1.5 of \cite{tropp2012user}) to
$\tSmat_3$, we have that for any $\alpha \geq  0$,
\begin{align*}
\Prob \Big[ \big \{ \frac{LT}{\sigma} \opnorm{\tSmat_3} \geq  \alpha \big \}
  \cap \Gevent \Big] &  \leq (d_x+d_a)\exp{ \left({-\alpha^2 \over
      2\sigma_{\tSmat_3}^2 + 2 D \alpha/3} \right) },
\end{align*}
for any $\sigma_{\tSmat_3}^2$ lower bounded by
\begin{align}
\label{eq:s3}
\max \Bigg \{ \bigopnorm{ \sum_{t=1}^T\E
   \big( \big(\sum_{\ell=1}^L \tvarepsilon_{t, \ell} \tbx_{t, \ell}
   \tbdelta_t^\top \big)\big( \sum_{\ell=1}^L \tvarepsilon_{t, \ell}
   \tbx_{t, \ell} \tbdelta_t^\top\big)^\top \big)} , 
 \bigopnorm{\sum_{t=1}^T\E \big( \big(\sum_{\ell=1}^L \tvarepsilon_{t,
   \ell} \tbx_{t, \ell} \tbdelta_t^\top \big)^\top \big(
 \sum_{\ell=1}^L \tvarepsilon_{t, \ell} \tbx_{t, \ell}
 \tbdelta_t^\top\big) \big) } \Bigg\},
\end{align}
and
\begin{align*}
 D = \max_{t}  \opnorm{\sum_{\ell=1}^L
 -\tvarepsilon_{t, \ell} \tbx_{t, \ell} \tbdelta_t^\top} \leq 6
 Lh\sqrt{\log(TL)}\sqrt{(d_x + 3\log(LT))(d_a + 3\log(T))} .
\end{align*}

Elementary calculation shows that the choice
\mbox{$\sigma_{\tSmat_3}^2 \defn h^2 \floor{T^{2\over 3}}L \max\{
  d_a,d_x \}$} satisfies~\Cref{eq:s3}.  Moreover, we set
\begin{multline*}
\alpha = 2h T^{1\over 3}\log(T) \sqrt{L\max\{d_a,d_x\}\log(d_a+d_x)}
\\ + 8hL\sqrt{\log(TL)} \sqrt{(d_x+ 3 \log(LT))(d_a+
  3\log(T))}(\log(d_x + d_a)+ 2 \log(T))
\end{multline*}
With this choice, we have
\begin{align}
\label[ineq]{eq:tSmat_3}
\Prob\Bigg(\opnorm{\frac{LT}{\sigma} \tSmat_3} \geq \alpha\Bigg) \le
\frac{1}{T^2}.
\end{align}
The \Cref{eq:tSmat_3} combined the
relation~\eqref{eq:tSmat_Smat} gives
\begin{align}
\label[ineq]{eq:Smat3}
\Prob \Big[ \Big \{ \opnorm{\Smat_3} > \pretwo{T} \big \} \cap \Gevent
  \Big] & \leq \frac{1}{T^2},
\end{align}
as claimed.


\subsection{Proof of \Cref{lem:main}}
\label{proof:lem:main}

For notational simplicity, we replace $t$ in the statement of the
lemma in the main paper with $T$ in the proof. Basically, we prove
that for any $T \geq 2$, we have the lower bound
\begin{align*}
    \ErrTerm_{T}(\dTheta) \geq { \floor{T^{2\over 3}} \over 2T} h^2
    \fronorm{\dTheta}^2 - \simplifykappa{T} \fronorm{\dTheta}^2
\end{align*}
with probability at least $1-{1\over LT} -{3\over T} $.

Let $\bb_t = \sum_{\ell=1}^L \bx_{t, \ell} $, and set $\bdelta_t =
\mathbf{0}$ for exploitation rounds.  Then we have the decomposition
\begin{align*}
     \ErrTerm_T(\dTheta) &= {1\over 2LT}\sum_{t=1}^T \sum_{\ell=1}^L
     (\ba_{t, \ell}^\top \dTheta \bx_{t, \ell} )^2 \; = \; {1\over
       2LT}\sum_{t=1}^T \sum_{\ell=1}^L \Big \{ \big( \tfrac{
       \bb_t^\top \hTheta_{t-1}^\top}{\|\bb_t^\top
       \hTheta_{t-1}^\top \|_2 }+ \bdelta_{t}^{\top} \big )
     \dTheta \bx_{t, \ell} \Big \}^2.
\end{align*}
This decomposition is well defined and valid with probability one,
since we interpret $\hTheta_{t-1}$ as $\hTheta_{t-1}$ when
$\hTheta_{t-1} \neq 0$ and as $\ba_{t,\ell}\bb_t^{\top}$ when
$\hTheta_{t-1}=0$.  Introduce the shorthand notation
\begin{align}
\label{eq:dterm}  
 \dterm_T(\dTheta) & \defn {1\over 2LT} \sum_{t=1}^T \sum_{\ell=1}^L
 \left(( \tfrac{\bb_t^\top \hTheta_{t-1}^{\top}}{\|\bb_t^\top
   \hTheta_{t-1}^\top \|_2 } \dTheta \bx_{t, \ell} )^2 + (
 \bdelta_{t}^{\top} \dTheta \bx_{t, \ell} )^2\right) \\
 \dterm_{1,T}(\dTheta) & \defn {1\over 2LT}\sum_{t=1}^T
 \sum_{\ell=1}^L ( \tfrac{ \bb_t^\top \hTheta_{t-1}^\top}{
   \|\bb_t^\top \hTheta_{t-1}^\top \|_2 } \dTheta \bx_{t, \ell}
 )^2, \quad \mbox{and} \\
\label{eq:dtermtwo} 
 \dterm_{2,T}(\dTheta) & \defn {1\over 2LT}\sum_{t=1}^T
 \sum_{\ell=1}^L ( \bdelta_{t}^{\top} \dTheta \bx_{t, \ell} )^2.
  \end{align}
With these definitions, we have the relations
\begin{align*}
 \ErrTerm_T(\dTheta) - \dterm_T(\dTheta) & = {1\over LT}\sum_{t=1}^T
 \sum_{\ell=1}^L ( \tfrac{ \bb_t^\top
   \hTheta_{t-1}^\top}{\|\bb_t^\top \hTheta_{t-1}^\top
   \|_2 } \dTheta \bx_{t, \ell}) ( \bdelta_{t}^{\top} \dTheta \bx_{t,
   \ell} ), \quad \\
    \E(\ErrTerm_T(\dTheta) - \dterm_T(\dTheta) ) & = 0, \quad \mbox{and} \\
\E( \dterm_{2,T}(\dTheta) ) & \geq  {\floor{T^{2\over 3}} \over 2T} h^2
\fronorm{\dTheta}^2 .
\end{align*}

Clearly, we have
\begin{align}
\label[ineq]{eq:Errterm_main_decomposition}
    \ErrTerm(\dTheta)\geq {\floor{T^{2\over 3}} \over 2T} h^2
    \fronorm{\dTheta}^2 +{ \ErrTerm_T(\dTheta) - \dterm_T(\dTheta)
      \over \fronorm{\dTheta}^2 } \fronorm{\dTheta}^2 + {
      \dterm_{2,T}(\dTheta) - \E( \dterm_{2,T}(\dTheta) ) \over
      \fronorm{\dTheta}^2} \fronorm{\dTheta}^2.
\end{align}

\newcommand{\kappatwo}{ -\frac{4h^2}{T^{2/3}}(2d_a + 6\log{T} + d_a
  \log\log{T})( 2d_x + 6 \log{TL} + d_x \log\log{TL} )\log{T} }

\newcommand{\kappaone}{- \frac{2\sqrt{2}h}{T^{2/3}}(d_x + 3\log{T} +
  \log{L}) (d_a+ 3\sqrt{d_a}+ 4+ d_x) \sqrt{\log{T}} }

Next we prove that the following two bounds hold with high
probability:
{\small
\begin{subequations}
  \begin{align}
\label{EqnBoundA}    
  \inf_{ \fronorm{\dTheta} > 0 } {\ErrTerm_T(\dTheta) -
    \dterm_T(\dTheta) \over \fronorm{\dTheta}^2 } & \geq \kappaone \\
\label{EqnBoundB}
\inf_{ \fronorm{\dTheta} > 0 } { \dterm_{2,T}(\dTheta) - \E(
  \dterm_{2,T}(\dTheta) ) \over \fronorm{\dTheta}^2 } & \geq \kappatwo
.
\end{align}
\end{subequations}
}

Before proceeding to proving~\Cref{EqnBoundA} and ~\Cref{EqnBoundB},
we define an event and introduce some truncated random variables.

Note that the random variable $\|\bx_{t, \ell}\|_2^2$ follows a
\mbox{$\chi_{d_x}^2$-distribution,} whereas $\|\bdelta_t/h\|_2^2$
follows a \mbox{$\chi_{d_a}^2$-distribution.} Therefore, by combining
standard $\chi^2$-tail bounds (Lemma 1 in~\cite{laurent2000adaptive})
with the union bound, for any choice of $\epsilon_1, \epsilon_2 > 0$,
we have
\begin{align}
\label[ineq]{eq:chisquare}
 \max_{\substack{t \in [T] \\ \ell \in [L]}} \|\bx_{t, \ell}\|_2^2
 \leq d_x + 2 \epsilon_1 +2 \sqrt{\epsilon_1 d_x}, \quad \mbox{and}
 \quad \max_{t \in [T]} \|\bdelta_t/h\|_2^2 \leq d_a + 2\epsilon_2 +
 2\sqrt{\epsilon_2 d_a}
\end{align}
with probability at least $1-\left( LT \exp{(-\epsilon_1)} +
T\exp{(-\epsilon_2)} \right)$.

Now we set $\epsilon_1 = 2\log(LT)$ $\epsilon_2 = 2\log(T)$, and we
introduce the shorthand
\begin{align*}
U_1 & \defn d_x + 2 \epsilon_1 +2\sqrt{\epsilon_1 d_x} \quad
\mbox{and} \quad U_2 \defn d_a + 2\epsilon_2 + 2\sqrt{\epsilon_2 d_a},
\quad \mbox{and the event} \\
\Oevent & \defn \Big \{ \max_{\substack{ t \in
    [T] \\ \ell \in [L]}} \|\bx_{t, \ell}\|_2^2 \leq U_1 \quad
\mbox{and} \quad \max_{t \in [T]} \|\bdelta_t/h\|_2^2 \leq U_2 \Big \}.
\end{align*}
Clearly, $\Prob(\Oevent) \geq  1 - \frac{1}{LT} - \frac{1}{T}$.

Now we introduce the truncated variables
\begin{align*}
    \tbx_{t,\ell} =  \bx_{t,\ell} \mathbbm{1}\{ \bx_{t,\ell} \leq U_1 \}, 
    \quad \tbdelta_t = \bdelta_t \mathbbm{1}\{ \|\bdelta_t/h\|_2^2 \leq U_2 \}.
\end{align*}
Using the truncated variables, we define the truncated version of the quantities defined above.
\begin{align*}
    &\tErrTerm_T (\dTheta)  =\tErrTerm_T (\dTheta) , \defn {1\over 2LT}\sum_{t=1}^T \sum_{\ell=1}^L \Big \{ \big( \tfrac{
       \bb_t^\top \hTheta_{t-1}^\top}{\|\bb_t^\top
       \hTheta_{t-1}^\top \|_2 }+ \tbdelta_{t}^{\top} \big )
     \dTheta \tbx_{t, \ell} \Big \}^2, \quad \mbox{ and }\\
     & \tdterm_T(\dTheta)  \defn {1\over 2LT} \sum_{t=1}^T \sum_{\ell=1}^L
 \left(( \tfrac{\bb_t^\top \hTheta_{t-1}^{\top}}{\|\bb_t^\top
   \hTheta_{t-1}^\top \|_2 } \dTheta \tbx_{t, \ell} )^2 + (
 \tbdelta_{t}^{\top} \dTheta \tbx_{t, \ell} )^2\right). 
 %
\end{align*}
Clearly, on event $\Oevent$, we have that 
\begin{align*}
  \tErrTerm_T (\dTheta)  =\ErrTerm_T (\dTheta) ,  & \qquad  \tdterm_T(\dTheta) = \dterm_T(\dTheta).
\end{align*}

\subsubsection{Proof of the bound~\eqref{EqnBoundA}:}
\label{sec:proof_EqnBoundA}

We introduce the shorthand notation
\begin{align*}
\interterm (T; \dTheta) \defn \frac{ \ErrTerm_T(\dTheta) -
  \dterm_T(\dTheta) }{\fronorm{\dTheta}}, \quad \mbox{and} \quad
\tinterterm (T; \dTheta) \defn \frac{ \tErrTerm_T(\dTheta) -
  \tdterm_T(\dTheta) }{\fronorm{\dTheta}}
\end{align*}
Then conditioned on the event $\Oevent$, we have
\begin{align*}
    \inf_{ \fronorm{\dTheta} > 0 } \interterm (T;\dTheta) = \inf_{
      \fronorm{\dTheta} > 0 } \tinterterm (T;\dTheta).
\end{align*}
Next we focus on bounding $ \inf_{ \fronorm{\dTheta} > 0 } \tinterterm
(T;\dTheta)$.  Elementary calculation gives
\begin{align*}
    \tinterterm(T;\dTheta) = {1\over T}\sum_{t=1}^T \Bigg\{
    \frac{1}{L} \sum_{\ell=1}^L ( \tfrac{ \bb_t^\top
      \hTheta_{t-1}^\top}{\|\bb_t^\top \hTheta_{t-1}^\top
      \|_2 } \frac{\dTheta}{\fronorm{\dTheta}} \tbx_{t, \ell}) (
    \tbdelta_{t}^{\top} \frac{\dTheta}{\fronorm{\dTheta}} \tbx_{t,
      \ell} ) \Bigg\},
\end{align*}
which gives \mbox{$\inf_{ \fronorm{\dTheta} > 0 } \tinterterm
  (T;\dTheta) = \inf_{ \fronorm{\dTheta} = 1 } \tinterterm
  (T;\dTheta)$.}

Next we take an $\eta$-covering of the set $\mathbbm{B} \defn
\{\dTheta \mid \fronorm{\dTheta}= 1\}$: $\CoveringSet = \{
\tdTheta_1,\tdTheta_2,\ldots, \tdTheta_{\Nc}\}$. By a standard
covering number calculation (e.g. see Example 5.8
in~\cite{wainwright2019high}), $\log{\Nc} \leq d_x d_a
\log(1+\frac{2}{\eta})$).  We have
\begin{align}
\inf_{ \fronorm{\dTheta} = 1 } \tinterterm (T;\dTheta) \geq
\underbrace{\inf_{ \dTheta \in \CoveringSet } \tinterterm (T;\dTheta)
}_{W_1}- \underbrace{\sup_{\stackrel{\fronorm{\dTheta_1-\dTheta_2}
      \leq \eta,} { \fronorm{\dTheta_1} =1, \; \fronorm{\dTheta_2}=1}
  } |\tinterterm (T;\dTheta_1) - \tinterterm (T;\dTheta_2)|}_{W_2}.
\end{align}

Next we analyze the quantities $W_2$ and $W_1$ separately.

\paragraph{Analysis of $W_2$: }
\begin{align*}
 W_2 & \leq \sup_{\stackrel{ \fronorm{\dTheta_1} =1,
     \fronorm{\dTheta}\leq \eta }{\fronorm{\dTheta_1 + \dTheta} =1 }}
 |\tinterterm (T;\dTheta_1 + \dTheta) - \tinterterm (T;\dTheta_1)|\\
  & \leq \sup_{\stackrel{ \fronorm{\dTheta_1} =1,
     \fronorm{\dTheta}\leq \eta }{\fronorm{\dTheta_1 + \dTheta} =1 }}
 \Bigg| {1\over LT}\sum_{t=1}^T \sum_{\ell=1}^L \bigg\{ \big( \tfrac{
   \bb_t^\top \hTheta_{t-1}^\top}{\|\bb_t^\top
   \hTheta_{t-1}^\top \|_2 } \dTheta \tbx_{t, \ell} \big) \big(
 \tbdelta_{t}^{\top} \dTheta_1 \tbx_{t, \ell} \big) + \big( \tfrac{
   \bb_t^\top \hTheta_{t-1}^\top}{\|\bb_t^\top
   \hTheta_{t-1}^\top \|_2 } (\dTheta+\dTheta_1) \tbx_{t, \ell}
 \big) \big( \tbdelta_{t}^{\top} \dTheta \tbx_{t, \ell} \big)
     \bigg\} \Bigg| \\
     & \leq \frac{\eta }{T^{1/3} } \cdot 2 U_1 \sqrt{U_2} h.
\end{align*}

\paragraph{Analysis of $W_1$: }
Our strategy is to bound $\E( \exp{(\lambda ( -W_1) )})$ for $\lambda
> 0$.  Recall from~\Cref{eq:his} our notation for the history up to
and including time $T$--- that is,
\begin{align*}
H_T & \defn \big \{ \bx_{t, \ell},\ba_{t, \ell},\reward_{t, \ell} \,
\mid \, t = 1, \ldots, T, \ell = 1, \ldots, L \big \}.
\end{align*}
We take conditional expectation on $(H_t, \{ \tbx_{t,\ell}\}_{\ell\in
  [L]}) $ iteratively for $t =T, T-1,\ldots,$ in our calculation of
$\E( \exp{(\lambda ( -W_1) )})$.  It should be noted that the sum
$\tinterterm(T;\dTheta)$ contains many zero terms corresponding to the
exploitation round. We introduce the shorthand for the set of
exploration rounds: $\rat \defn \{\floor{w^{3/2}}| w=1,2,\ldots\}$.
With this set-up, we have
\begin{align*}
 \E\Big( \exp{\big(\lambda ( -W_1) \big)} \Big) & = \E\Big(
 \exp{\big(\sup_{ \dTheta \in \CoveringSet } -\lambda \tinterterm
   (T;\dTheta) \big)} \Big)
 \\ & \leq \sum_{\dTheta\in \CoveringSet } \E\Big( \exp{\big( -
   \lambda \tinterterm (T;\dTheta) \big)} \Big) = \sum_{\dTheta\in
   \CoveringSet } \E\bigg\{ \E\Big[ \exp{\Big( - \lambda \tinterterm
     (T;\dTheta) \Big)} \bigg| H_{T-1}, \{ \tbx_{T,\ell}\}_{\ell\in
     [L]} \Big] \bigg\} \\
 &= \sum_{\dTheta\in \CoveringSet } \E\bigg\{ \underbrace{ \exp{\big(
     - \lambda \tinterterm (T-1;\dTheta) \big) } }_{\text{measurable
     w.r.t. } \big(H_T, \; \{ \tbx_{T,\ell}\}_{\ell\in [L]} \big) }
 \times\\
 & \qquad \underbrace{ \E\Big[ \exp{( - \frac{\lambda}{T} \Bigg\{
       \underbrace{ \tfrac{ \bb_T^\top
           \hTheta_{T-1}^\top}{\|\bb_T^\top
           \hTheta_{T-1}^\top \|_2 } \dTheta \big( \frac{1}{L}
         \sum_{\ell=1}^L \tbx_{T, \ell} \tbx_{T, \ell}^\top \big)
         \dTheta^\top }_{ \text{ vector } \bzeta: \text{ measurable
           w.r.t. } \big(H_T, \; \{ \tbx_{T,\ell}\}_{\ell\in [L]}
         \big) } \;\;\; \tbdelta_{T} \Bigg\} )} \; \Big| \; H_{T-1},
     \{ \tbx_{T,\ell}\}_{\ell\in [L]} \Big] }_{\text{conditional
     expectation to be calculated}} \bigg\} \\
 & \leq \sum_{\dTheta\in \CoveringSet } \E\bigg\{ \exp{\big( - \lambda
   \tinterterm (T-1;\dTheta) \big) } \times E\Big(\exp{ (
   -\frac{\lambda}{T} \bzeta^\top \bdelta_T ) } \Big| H_T, \; \{
 \tbx_{T,\ell}\}_{\ell\in [L]} \Big) \bigg] \Bigg\} \\
& = \sum_{\dTheta\in \CoveringSet } \E\bigg\{ \exp{\big( - \lambda
     \tinterterm (T-1;\dTheta) \big) } \times \bigg[
     \mathbbm{1}\Big\{T\notin \rat \Big\} + \mathbbm{1}\Big\{ T\in
     \rat \Big\} \exp{\big( \frac{h^2}{2} \frac{\lambda^2}{T^2}
       \|\bzeta\|_2^2 \big) } \bigg] \Bigg\} \\
& \leq \sum_{\dTheta\in \CoveringSet } \E\bigg\{ \exp{\big( - \lambda
     \tinterterm (T-1;\dTheta) \big) } \times \bigg[
     \mathbbm{1}\Big\{T\notin \rat \Big\} + \mathbbm{1}\Big\{ T\in
     \rat \Big\} \exp{\big( \frac{h^2}{2} \frac{\lambda^2}{T^2} U_1^2
       \big) } \bigg] \Bigg\} \\
& \leq \Nc \exp(\frac{h^2}{2} \frac{\lambda^2 T^{2/3}}{T^2} U_1^2 )
   \leq \exp{( \lambda^2 \frac{1}{T^{4/3}} \frac{h^2U_1^2}{2} + d_a
     d_x \log{(1 + \frac{2}{\eta})} )}
\end{align*}

Therefore, for any $\gamma_1 > 0$,
\begin{align*}
 \Prob(W_1< -\gamma_1) & \leq E( \exp(-\lambda W_1 - \lambda
 \gamma_1)) \\
 & \leq \exp{( \lambda^2 \frac{1}{T^{4/3}} \frac{h^2U_1^2}{2} + d_a d_x
   \log{(1 + \frac{2}{\eta})} -\lambda \gamma_1 ) } \\
 & = \exp{ \Big( \frac{1}{2}( \underbrace{\frac{\lambda}{T^{2/3}}hU_1
     - \gamma_1 \frac{T^{2/3}}{hU_1} }_{\tau } )^2 \Big) } \exp{\Big(
   d_a d_x \log(1+\frac{2}{\eta}) - \frac{1}{2}(\frac{\gamma_1
     T^{2/3}}{hU_1})^2 \Big)}.
\end{align*}
The choice $\lambda =\gamma_1 \frac{ T^{4/3} }{h^2U_1^2}$ ensures that
$\tau =0$ in the above display.  Moreover, if we set
\begin{align*}
  \gamma_1 \defn \frac{hU_1}{T^{2/3}} \sqrt{2 d_a d_x
    \log{(1+\frac{2}{\eta})} + 2 \log{T}},
\end{align*}
then we obtain the bound $\Prob(W_1< -\gamma_1) \leq \frac{1}{T}$.

Combining the analysis of $W_1$ and $W_2$, we see that, with
probability at most $\frac{1}{T}$,
\begin{align*}
    \inf_{ \fronorm{\dTheta} = 1 } \tinterterm (T;\dTheta) <
    \underbrace{- \frac{hU_1}{T^{2/3}} \sqrt{2 d_a d_x
        \log{(1+\frac{2}{\eta})} + 2 \log{T}} -
      \frac{\eta}{T^{1/3}}\cdot 2U_1\sqrt{U_2}h }_{\xi} .
\end{align*}

Setting $\eta \defn \frac{1}{T^{1/3}}$, we can lower bound $\xi$ as
\begin{align*}
   \xi & \geq - \frac{hU_1}{T^{2/3}} \sqrt{2d_a d_x ( \log{3} +
     \frac{1}{3}\log{T} ) + 2\log{T}} - \frac{2 U_1\sqrt{U_2}
     h}{T^{2/3} } \\
   & \geq - \frac{hU_1}{T^{2/3}} (d_a+d_x) \sqrt{ 2 \log{T} } -
   \frac{2 U_1\sqrt{U_2} h}{T^{2/3} } .
\end{align*}
Note that $$ U_1 \leq 2d_x + 6\log{T} + 6\log{L} ,\qquad U_2 \leq 2d_a
+ 6\log{T} ,$$ Further calculation gives that
\begin{align}
\label{def:kappaone}
    \xi > \underbrace{- \frac{2\sqrt{2}h}{T^{2/3}}(d_x + 3\log{T} +
      \log{L}) (d_a+ 3\sqrt{d_a}+ 4+ d_x) \sqrt{\log{T}} }_{\kappa_1}.
\end{align}
Therefore,
\begin{align}
\label[ineq]{eq:bound_interterm}
    \Prob\Bigg( \Big\{ \inf_{\fronorm{\dTheta}> 0}   \interterm(T;\dTheta) < \kappa_1 \Big\} \cap \Oevent \Bigg) < \frac{1}{T}.
\end{align}


\subsubsection{Proof of the bound~\eqref{EqnBoundB}:}
We first introduce a slightly different form of truncation.  Define
the truncation thresholds
\begin{align}
\label{def:checkvariables}
\cU_1 \defn 2 d_x + 6\log{(TL)} + d_x\log{\log{(TL)}} \quad \mbox{and}
\quad \cU_2 \defn 2d_a + 6\log{(T)} + d_a \log{\log{(T)}} .
\end{align}
Using these truncation levels, we define the ``truncation event'' as
\begin{align*}
  \cOevent \defn & \Big \{ \max_{\substack{ t \in [T] \\ \ell \in
      [L]}} \|\bx_{t, \ell}\|_2^2 \leq \cU_1 \quad \mbox{and} \quad
  \max_{t \in [T]} \|\bdelta_t/h\|_2^2 \leq \cU_2 \Big \}.
\end{align*}
Clearly, this newly defined ``truncation'' event $\cOevent \subset
\Oevent$.

We introduce the shorthand for truncated variables associated with
$\cOevent$:
\begin{align*}
  \cbx_{t,\ell} = \bx_{t,\ell} \mathbbm{1}\{ \| \cbx_{t,\ell} \|_2^2
  \leq \cU_1 \} , \qquad \cbdelta_t = \bdelta_t \mathbbm{1}\{
  \|\bdelta_t/h\|_2^2\leq \cU_2 \}.
\end{align*}
With the truncated variables, we introduce the shorthand for the
quantity to bound and its truncated version.
\begin{subequations}
\begin{align}
  \cdterm_{2,T} (\dTheta) = \frac{1}{2LT} & \sum_{t=1}^{T} \sum_{\ell
    = 1}^L \big( \cbdelta_t^\top \dTheta \cbx_{t,\ell}
  \big)^2, \label{eq:cdtermtwo}\\
    \fluctterm(T;\dTheta) = { \dterm_{2,T}(\dTheta) - \E(
      \dterm_{2,T}(\dTheta)) \over \fronorm{\dTheta}^2 }, & \qquad
    \cfluctterm(T;\dTheta) = { \cdterm_{2,T}(\dTheta) - \E(
      \cdterm_{2,T}(\dTheta)) \over \fronorm{\dTheta}^2 } .
\end{align}
\end{subequations}
Elementary calculation shows that
\begin{align}
 \inf_{\fronorm{\dTheta}>0} \fluctterm(T;\dTheta) &=
 \inf_{\fronorm{\dTheta}=1} \fluctterm(T;\dTheta) \notag \\
\label{eq:truncation_decomposition} 
 & \geq \underbrace{\inf_{\fronorm{\dTheta}=1}
  \cfluctterm(T;\dTheta)}_{Z_1(T)} -
\underbrace{\sup_{\fronorm{\dTheta}=1} |\fluctterm(T;\dTheta) -
  \cfluctterm(T;\dTheta)| }_{Z_2(T)}.
\end{align}
Therefore, we only need to bound $Z_2(T)$ and $Z_1(T)$.

\paragraph{Analysis of $Z_1(T)$: }
Recall that we use $\rat$ to denote the set of exploration
rounds. Rewrite $Z_1(T)$ as
\begin{align*}
    Z_1(T) = \inf_{\fronorm{\dTheta}=1}\frac{1}{2LT} \sum_{t\in \rat
      \cap [T]} \sum_{\ell = 1}^L \bigg\{ \big( \cbdelta_t^\top
    \dTheta \cbx_{t,\ell} \big)^2 - \E\Big( \big( \cbdelta_t^\top
    \dTheta \cbx_{t,\ell} \big)^2 \Big) \bigg\}.
\end{align*}
Clearly, for truncated variables $\cbdelta_t$ and $\cbx_{t,\ell}$,
\begin{align*}
\sup_{\fronorm{\dTheta}=1}\bigg| \big( \cbdelta_t^\top \dTheta
\cbx_{t,\ell} \big)^2 - \E\Big( \big( \cbdelta_t^\top \dTheta
\cbx_{t,\ell} \big)^2 \Big) \bigg| \leq \cU_1\cU_2 h^2 .
\end{align*}
As there are at most $\floor{T^{2/3}}$ terms in $\rat \cap [T]$,
applying Functional Hoeffding Theorem (see Theorem 3.26 in
\cite{wainwright2019high}) gives
\begin{align*}
\Prob \Big( Z_1(T) \leq \E(Z_1(T)) - \alpha \Big) \leq \exp{\Big( -
  \frac{ \alpha^2 T^2 }{ 4\floor{T^{2/3}} \cU_1^2\cU_2^2 h^4} \Big)},
\end{align*}
for $\alpha \geq 0$.

Setting $\alpha = \frac{2}{T^{2/3}}\cU_1\cU_2h^2 \log{T}$ yields
\begin{align}
\label[ineq]{eq:tail_Z1}
\Prob \Big( Z_1(T) \leq \E(Z_1(T)) - \frac{2}{T^{2/3}} \cU_1 \cU_2 h^2
\log{T} \Big) \leq \frac{1}{T}.
\end{align}

Next we will bound $\E(Z_1(T))$.  Note that $\{\dTheta \mid
\fronorm{\dTheta} =1\}$ is separable compact space. By symmetrization
argument (see Lemma 11.4 in \cite{boucheron2013concentration}), we
have
\begin{align*}
\Big| \E(Z_1(T)) \Big| & \leq \frac{1}{2L} \sum_{\ell = 1}^L \E\Bigg\{
\sup_{\fronorm{\dTheta}=1} \bigg| \frac{1}{T} \sum_{t\in \rat \cap
  [T]} \big( \cbdelta_t^\top \dTheta \cbx_{t,\ell} \big)^2 - \E\Big(
\big( \cbdelta_t^\top \dTheta \cbx_{t,\ell} \big)^2 \Big) \bigg|
\Bigg\}\\
   & \stackrel{\substack{\text{symmetrization}\\\text{argument}}}{\le}
\frac{1}{L} \sum_{\ell = 1}^L \E\Bigg\{ \sup_{\fronorm{\dTheta}=1}
\Big| \frac{1}{T} \sum_{t\in \rat \cap [T]} \omega_t \big(
\cbdelta_t^\top \dTheta \cbx_{t,\ell} \big)^2 \Big| \Bigg\},
\end{align*}
where $\{w_i\}$ are independent Rademacher variables.

By the contraction principle (e.g., Theorem 11.6
of~\cite{boucheron2013concentration}), we have
\begin{align*}
  \frac{1}{L} \sum_{\ell = 1}^L \E\Bigg\{ \sup_{\fronorm{\dTheta}=1}
  \Big| \frac{1}{T} \sum_{t\in \rat \cap [T]} \omega_t \big(
  \cbdelta_t^\top \dTheta \cbx_{t,\ell} \big)^2 \Big| \Bigg\} &
  \stackrel{\substack{\text{contraction}\\\text{principle}}}{\le}
  \frac{1}{L} \sum_{\ell = 1}^L \E\Bigg\{ \sup_{\fronorm{\dTheta}=1} 2
  h\sqrt{\cU_1\cU_2}\bigg| \frac{1}{T} \sum_{t\in \rat \cap [T]}
  \omega_t \cbdelta_t^\top \dTheta \cbx_{t,\ell} \bigg| \Bigg\} \\
  & \stackrel{\text{by }\fronorm{\dTheta}=1 }{\le}
  \frac{1}{L} \sum_{\ell = 1}^L  2 h\sqrt{\cU_1\cU_2} \frac{1}{T} \E\bigg( \fronorm{  \sum_{t\in \rat \cap [T]} \omega_t    \cbx_{t,\ell}\cbdelta_t^\top   } \bigg)\\
  & \stackrel{\substack{\text{Cauchy}\\\text{Schwartz}}}{\le}
  \frac{1}{L} \sum_{\ell = 1}^L 2h \sqrt{\cU_1\cU_2} \frac{1}{T}
  \sqrt{\E\bigg( \fronorm{ \sum_{t\in \rat \cap [T]} \omega_t
      \cbx_{t,\ell}\cbdelta_t^\top }^2 \bigg)} \\
& \leq 2h^2\sqrt{\cU_1\cU_2} \frac{1}{T^{2/3}} \sqrt{d_ad_x}.
\end{align*}
Consequently, we have established the lower bound
\begin{align}
\label[ineq]{eq:EZ1}
    \E(Z_1(T)) \geq - 2h^2 \sqrt{\cU_1\cU_2} \frac{1}{T^{2/3}}
    \sqrt{d_a d_x}.
\end{align}
Combining~\Cref{eq:EZ1} with~\Cref{eq:tail_Z1} yields
\begin{align*}
    \Prob \Big( Z_1(T) \leq - 2h^2 \sqrt{\cU_1\cU_2} \frac{1}{T^{2/3}}
    \sqrt{d_a d_x} - \frac{2}{T^{2/3}} \cU_1 \cU_2 h^2 \log{T} \Big) &
    \leq \frac{1}{T}.
\end{align*}

\paragraph{Analysis of $Z_2(T)$: }
By the definition of $Z_2(T)$, we have the following decomposition
\begin{align*}
    Z_2(T)\leq \sup_{\fronorm{\dTheta}=1} \bigg|
    \E\Big(\dterm_{2,T}(\dTheta)\Big) - \E\Big( \cdterm_{2,T}(\dTheta)
    \Big) \bigg| + \mathbbm{1}\{ \cOevent^c\} \cdot
    \sup_{\fronorm{\dTheta}=1} \big| \dterm_{2,T}(\dTheta) -
    \cdterm_{2,T}(\dTheta) \big| ,
\end{align*}
where $\dterm(T;\dTheta)$ and $\cdterm(T;\dTheta)$ are defined
in~\Cref{eq:dtermtwo} and~\Cref{eq:cdtermtwo}.  The second term is
equal to zero with high probability, since the event $\cOevent$
happens with high probability.

Turning to the first term, we have
{\small
\begin{align*}
\sup_{\fronorm{\dTheta}=1} \bigg| \E\Big(\dterm_{2,T}(\dTheta)\Big) -
\E\Big( \cdterm_{2,T}(\dTheta) \Big) \bigg| & =
\sup_{\fronorm{\dTheta}=1} \frac{1}{2LT} \sum_{t=1}^T \sum_{\ell =
  1}^L \E\bigg( \Big(\bdelta_t^\top \dTheta \bx_{t,\ell} \Big)^2
\mathbbm{1}\Big\{ \| \bdelta_t/h \|_2^2 > \cU_2 \text{ or } \|
\bx_{t,\ell}\|_2^2 > \cU_1 \Big\} \bigg) \\
& \leq \underbrace{\sup_{\fronorm{\dTheta}=1} \frac{1}{2LT}
  \sum_{t=1}^T \sum_{\ell = 1}^L \E\Big( \big( \bdelta_t^\top \dTheta
  \bx_{t,\ell} \big)^2 \mathbbm{1}\big\{ \|\bdelta_t/h\|_2^2 >
  \cU_2\big\} \Big) }_{\varsigma_2} \\
& + \underbrace{ \sup_{\fronorm{\dTheta}=1} \frac{1}{2LT} \sum_{t=1}^T
  \sum_{\ell = 1}^L \E\bigg( \Big(\bdelta_t^\top \dTheta \bx_{t,\ell}
  \Big)^2 \mathbbm{1}\{ \|\bdelta_t/h\|_2^2\leq \cU_2 \text{ and } \|
  \bx_{t,\ell}\|_2^2 > \cU_1 \}\bigg) }_{\varsigma_1}.
\end{align*}
}
Next we bound $\varsigma_2$ and $\varsigma_1$ separately.

For $\varsigma_2$, note that only $\floor{T^{2/3}}$ many terms are
non-zero, so that
\begin{align*}
    \varsigma_1 &=\sup_{\fronorm{\dTheta}=1} \frac{1}{2T} \sum_{t=1}^T
    \E\Big( \| \bdelta_t^\top \dTheta \|_2^2 \mathbbm{1}\big\{
    \|\bdelta_t/h\|_2^2 > \cU_2\big\} \Big) \\
& = \sup_{\fronorm{\dTheta}=1} \frac{1}{2T} \sum_{t=1}^T \E\Big(
    \tracer{ \dTheta\dTheta^\top }{\bdelta_t \bdelta_t^\top}
    \mathbbm{1}\big\{ \|\bdelta_t/h\|_2^2 > \cU_2\big\} \Big)\\
& = \frac{\floor{T^{2/3}}h^2}{2T} \E( \frac{\|\bdelta\|_2^2}{d_a}
    \mathbbm{1}\{ \|\bdelta\|_2 >\cU_2 \} ),
\end{align*}
where $\delta \sim N(\mathbf{0}, \mathbf{I}_{d_a})$.  Next, we will
calculate $\E( \|\bdelta\|_2^2 \mathbbm{1}\{ \|\bdelta\|_2 >\cU_2 \}
)$. We use the spherical coordinates. and let $V(d)$ denote the volume
of the unit ball in $\R^d$.  Then by dividing by the integral of the
normal distribution density and canceling the same terms, we have
\begin{align*}
    \E( |\bdelta\|_2^2 \mathbbm{1}\{ \|\bdelta\|_2 >\cU_2 \} ) =
    \frac{ \int_{\sqrt{\cU_2}}^{\infty} \exp{(-r^2/2)} r^2 \cdot
      r^{d_a-1} \dd r }{ \int_{0}^{\infty} \exp{(-r^2/2)} \cdot
      r^{d_a-1} \dd r}.
\end{align*}
Elementary calculation shows that
\begin{multline*}
    \int_{\sqrt{\tau}}^{\infty} \exp{(-r^2/2)} \cdot r^{d} \dd r \\=
    \begin{cases}
     \exp{(-\frac{\tau}{2})} 2^{k-\frac{1}{2}} \Big( \frac{(k-1/2)!}{
       (1/2)!} \int_{\frac{\tau}{2}}^{\infty}
     \sqrt{t}\exp{(\frac{\tau}{2}-t)} \dd t + \sum_{i = 0}^{k-2}
     (\frac{\tau}{2})^{k-\frac{1}{2}-i} \frac{(k-1/2)!}{ (k-(1/2)-i)!}
     \Big), & \text{ even } d=2k \\
      \exp{(-\frac{\tau}{2})} 2^k \sum_{i = 0}^{k} (\frac{\tau}{2})^{k-i} \frac{k!}{ (k-i)!}   , & \text{ odd } d=2k+1
    \end{cases}.
\end{multline*}
Note that $\cU_2>d_a$.  Therefore, when $d_a$ is even, we have
\begin{align*}
  \E( \|\bdelta\|_2^2 \mathbbm{1}\{ \|\bdelta\|_2 >\cU_2 \} ) &= 2
  \frac{ \exp{(-\cU_2/2)} \sum_{i = 0}^{d_a/2}
    (\frac{\cU_2}{2})^{(d_a/2) -i} \frac{(d_a/2)!}{ ((d_a/2)-i)!} }{
    ((d_a/2) -1)! } \\
& = \exp{(-\cU_2/2)} d_a \sum_{i = 0}^{d_a/2}
  (\frac{\cU_2}{2})^{(d_a/2) -i} \frac{1}{ ((d_a/2)-i)!}\\ & <
  \exp{(-\cU_2/2)} (\frac{\cU_2}{2})^{(d_a/2)}
  \frac{1}{(d_a/2)!}\frac{d_a}{ 1 - ( d_a/\cU_2 )} ,
\end{align*}
whereas when $d_a$ is odd, we have
\begin{align*}
\E( \|\bdelta\|_2^2 \mathbbm{1}\{ \|\bdelta\|_2 >\cU_2 \} ) & =\frac{
  \exp{(-\frac{\cU_2}{2})} 2^{\frac{d_a}{2}} \Big(
  \frac{(\frac{d_a}{2})!}{ (1/2)!} \int_{\frac{\cU_2}{2}}^{\infty}
  \sqrt{t}\exp{(\frac{\cU_2}{2}-t)} \dd t + \sum_{i = 0}^{(d_a-3)/2}
  (\frac{\cU_2}{2})^{\frac{d_a}{2}-i} \frac{(\frac{d_a}{2})!}{
    (\frac{d_a}{2}-i)!}  \Big) }{ 2^{\frac{d_a-2}{2}} \sqrt{\pi}
  \prod_{i=0}^{(d_a-5)/2}(d_a/2-1-i)} \\
& = \exp{(-\frac{\cU_2}{2})} \frac{d_a}{\sqrt{\pi}} \Big(
\int_{\frac{\cU_2}{2}}^{\infty} \sqrt{t}\exp{(\frac{\cU_2}{2}-t)} \dd
t + \sum_{i = 0}^{(d_a-3)/2} (\frac{\cU_2}{2})^{\frac{d_a}{2}-i}
\frac{(1/2)!}{ (\frac{d_a}{2}-i)!}  \Big) \\
& < \exp{(-\frac{\cU_2}{2})} (\frac{\cU_2}{2})^{(d_a/2)}
\frac{d_a}{\sqrt{\pi}} \Big( \frac{1}{1- (d_a/\cU_2)}
\frac{(1/2)!}{(d_a/2)!} + \sqrt{\pi} (\frac{\cU_2}{2})^{(-d_a/2)}
\Big).
\end{align*}
Plugging in $\cU_2 \geq 2d_a + 3\log{T} + d_a \log{\log{T}} $ and
using Stirling formula, we have that for $d_a\geq 1, T\geq 2$,
\begin{align*}
\E( \|\bdelta\|_2^2 \mathbbm{1}\{ \|\bdelta\|_2 >\cU_2 \} ) <
\frac{2d_a}{T^{3/2}}.
\end{align*}
Therefore, we have the bound \mbox{$\varsigma_2 < h^2 T^{-(11/6)}$.}
Similarly, for $\varsigma_1$, as $\cU_1 \geq 2 d_x + 3\log{LT} +
\log{\log{(LT)}}$, we have \mbox{$\varsigma_1 \leq \frac{h^2}{T^{1/3}}
  \frac{1}{(TL)^{3/2}}$.}

Therefore, we have shown that
\begin{align*}
Z_2(T) \leq \frac{2h^2}{T^{11/6}} + \mathbbm{1}\{\cOevent^c\} \cdot
\sup_{\fronorm{\dTheta}=1} | \dterm_{2,T}(\dTheta)
-\cdterm_{2,T}(\dTheta) |.
\end{align*}

Combining the analysis of $Z_1(T)$ and $Z_2(T)$ with the
decomposition~\eqref{eq:truncation_decomposition} yields
\begin{align*}
\Prob \bigg( \Big\{ \inf_{\fronorm{\dTheta}>0} \fluctterm(T;\dTheta) <
- 2 h^2 \sqrt{\cU_1\cU_2} \frac{1}{T^{2/3}} \sqrt{d_a d_x} -
\frac{2}{T^{2/3}} \cU_1\cU_2 h^2 \log{T} - \frac{2h^2}{T^{11/6}}
\Big\} \cap \cOevent \bigg) \leq \frac{1}{T}.
\end{align*}
Plugging in $\cU_1$ and $\cU_2$ from~\Cref{def:checkvariables}, and
setting
{\small
\begin{align}
\label{def:kappatwo}
    \kappa_2 \defn -\frac{4h^2}{T^{2/3}}(2d_a + 6\log{T} + d_a
    \log\log{T})( 2d_x + 6\log{TL} + d_x \log\log{TL} )\log{T} ,
    \log{(d_a+d_x+2)}+2)^2 \log{T},
\end{align}
}
we find that
\begin{align}
\label[ineq]{eq:bound_fluctterm}
    \Prob\Big( \big\{  \inf_{\fronorm{\dTheta}>0} \fluctterm(T;\dTheta)  < \kappa_2 \big\} \cap \cOevent \Big) < \frac{1}{T}.
\end{align}

Next we combine the high-probability bounds for
$\inf_{\fronorm{\dTheta}>0} \fluctterm(T;\dTheta)
$~\Cref{eq:bound_fluctterm} and $\inf_{\fronorm{\dTheta}>0}
\interterm(T;\dTheta)$~\Cref{eq:bound_interterm} to prove lower bound
for $\ErrTerm(\dTheta)$.

Recalling the inclusion $\cOevent \subset \Oevent$, we have
\begin{align*}
 \Prob \left( \inf_{\fronorm{\dTheta}>0} \interterm(T;\dTheta) < \kappa_1
 \text{ or } \inf_{\fronorm{\dTheta}>0} \fluctterm(T;\dTheta) <
 \kappa_2 \right) < \frac{1}{T} + \frac{1}{T} + \Prob(\Oevent^c) \leq
 \frac{3}{T} + \frac{1}{LT},
\end{align*}
where $\kappa_1$ and $\kappa_2$ are defined in~\Cref{def:kappaone}
and~\Cref{def:kappatwo}, respectively.  Therefore, going back
to~\Cref{eq:Errterm_main_decomposition}, we have
\begin{align*}
    \ErrTerm(\dTheta) & \stackrel{\qquad }{\ge}
    \frac{\floor{T^{\frac{2}{3}}}}{2T} h^2 \fronorm{\dTheta}^2 +
    \kappa_1\fronorm{\dTheta}^2 + \kappa_2\fronorm{\dTheta}^2\\
    & \stackrel{\substack{\text{plug in } \\ \kappa_1, \kappa_2}}{=}
    \frac{\floor{T^{\frac{2}{3}}}}{2T} h^2 \fronorm{\dTheta}^2 \\ &
    \qquad \kappaone \fronorm{\dTheta}^2\\ & \qquad \kappatwo
    \fronorm{\dTheta}^2 \\
    & \stackrel{\substack{\text{simplify}\\ \text{terms}}}{\ge}
    \frac{\floor{T^{2\over 3}}}{2T}h^2\fronorm{\dTheta}^2 -
    \simplifykappa{T} \fronorm{\dTheta}^2,
\end{align*}
with probability at least $1-\frac{3}{T} - \frac{1}{LT}$.  This
concludes the proof of the lemma.


\section{Details on Simulation and Real Case Studies}
\label{app:sim-case-study}

In this section, we provide more details on the tuning parameters of
different algorithms in Simulation II in~\Cref{app:sim-imp} and on the
two case studies in Sections~\ref{app:sim}
and~\ref{app:case-study-ii}.


\subsection{Details on Simulation II}
\label{app:sim-imp}

In this section, we detail the tuning parameters of each algorithm we
used for the simulation study.

\paragraph{\hicoab.} There are three tuning parameters for \hicoab: we
set the initialization steps as \mbox{$\tinit = 100$}; the initial
penalization parameter \mbox{$\lambda_0= \opnorm{{1\over 2\tinit L}
    \sum_{i=1}^{\tinit} \sum_{\ell = 1}^L |\ba_i^\top\hTheta_{\tinit}
    \bx_{i,\ell} - \reward_{i,\ell} | \bx_{i,\ell}\ba_i^\top }$;} and
the exploration parameter \mbox{$h = 0.1$.}

\paragraph{LinUCB~\citep{li2010contextual}.} 
We apply the LinUCB algorithm with disjoint linear models and set
multiplier for the upper confidence bound
$\alpha=1+\sqrt{\log(2/\delta)/2}$ with $\delta = .05$ as suggested in
the paper.

\paragraph{Lasso Bandit \citep{bastani2020online}.}
There are several tuning parameters in the original algorithm
including $h$ for the set of ``near-optimal arms'', $q$ for the
force-sample set, and $\lambda_1$ and $\lambda_{2,0}$ as the
regularization parameters for the ``forced sample estimate'' and
``all-sample estimate''.  We follow the original paper and set $h =
5$, $\lambda_1 = \lambda_{2,0} = 0.05$.  We set $q = 2$ so that the
size of initialized forced sample set is close to that we used for
$\hicoab$.

\paragraph{NeuralUCB \citep{zhou2020neural}.}

The tuning parameters of NeuralUCB include the confidence parameter as
in all UCB-based algorithm, the size of neural network, as well as the
step size, regularization parameter for gradient descent to train the
neural network.  We adapted the code from
\url{https://github.com/uclaml/NeuralUCB} and used the default
settings.

\paragraph{EE-Net \citep{ban2021ee}.}

EE-Net involves tuning parameters for gradient descent to train the
exploitation network, exploration network, and the decision-maker
network.  We adapted the code from
\url{https://github.com/banyikun/EE-Net-ICLR-2022} and used the
default settings.

\paragraph{G-ESTT \citep{kang2022efficient}.}
We implement the algorithm sketched in Appendix H of
\cite{kang2022efficient} as a potential extension of their main
algorithm to contextual setting.  Their Theorem 4.3 suggests a choice
of $T_1$ for which they documented good performance using their main
algorithm, but not for the contextual bandits of interest here.  To
best implement their idea in our setting, we note that our moderate
dimensions are already relatively large for their algorithm and our
typical time scope $T=1000$ is far too small for what is required in
their algorithm.  With this issue in mind, we set $T_1 =
\sqrt{rT\log((d_1+d_2)/\delta} / D_{rr}$ instead where $d_1 = d_x$,
$d_2 = d_a$, $\delta = 0.01$ as in their setup, $D_{rr} = 0.5$ is the
smallest non-zero singular value.


\begin{figure}[h]
\centering
\widgraph{\linewidth}{\figdir/fig_lin-non-sparse-sparse_avg_regret_app}
  \caption{Expected average regret.}
  \label{fig:sim-app}
\end{figure}

In \cite{kang2022efficient}, the main algorithm with theoretical guarantees
is designed for low-rank matrix bandit but not for contextual bandit. 
In Appendix H, the author sketched an extension to
the contextual setting, but their theory and numerical validation only 
apply to non-contextual bandits. 
We applied the modified G-ESTT to the contextual bandit setting where $d_a = 10, d_x = 100$
and the sparse setting with $d_a = 10, d_x = 100, s_0 = 2$ as in \Cref{Subsec-bandit}.
\Cref{fig:sim-app} shows that the modified G-ESTT does not perform well
compared to other contextual bandit algorithms. 
One reason can be the following.
They adopt the explore-then-commit algorithm and their initialization step
is required to be of the order $\sqrt{d_1 d_2 r T}$. 
In our simulation setting, the dimension of covariate $d_2 = d_x$ is high and therefore
their algorithm will require a large $T_1$ to perform well. 
Therefore, the modified G-ESTT does not perform well in high-dimensional settings (note that the dimensions in their simulations is of the order 10), especially when $T$ is relatively small.
In addition, the computational complexity of the modified G-ESTT is high compared to other methods.
We tried to run the modified G-ESTT for other settings as in \Cref{Subsec-bandit} 
with larger $d_x$ but it would have taken too long and 
we do not expect a better performance of the modified G-ESTT in higher dimension settings
given the above-mentioned claim.

\subsection{More details on Case Study I}
\label{app:sim}

In this section, we provides more background information on Case
Study I and additional numerical results.

\Cref{fig:sku-sales} shows the daily sales by product and each
color represents one product (only products that appeared more than
95\% of the days are colored; the rest are colored as grey). The days
corresponding to the vertical dashed grey lines are days with
promotion. The two red vertical lines correspond to the annual sales
events. The variation between products was large and one product
dominated the rest most of the time. The sales were also driven by the
promotion -- the sales went up when there is a promotion. 
\Cref{fig:price} shows the median unit price across time with the 25th
and 75th quantiles as the boundaries of the grey area. The median unit
price was around 3.2 RMB and there were variations in unit price among
products.  \Cref{fig:num-flav} shows the number of single-flavor
and multi-flavor products. Three-quarters of the products were
single-flavored. Note that products with the same flavor can have
different package sizes.  \Cref{fig:size} shows the number of
products with different package sizes. The package size of about 60\%
of the products is larger than 20 with 30\% having package sizes
between 10 and 20 and the rest less than 10.

\begin{figure}
\centering
\begin{subfigure}{.5\textwidth}
  \centering
  \widgraph{\linewidth}{\figdir/fig_sales_by_SKU_daily_sm}
  \caption{Total sales by product in 1k RMB.}
  \label{fig:sku-sales}
\end{subfigure}%
\begin{subfigure}{.5\textwidth}
  \centering
  \widgraph{\linewidth}{\figdir/fig_pricing}
  \caption{Median price of products.}
  \label{fig:price}
\end{subfigure}
\begin{subfigure}{.5\textwidth}
  \centering
  \widgraph{\linewidth}{\figdir/fig_num_flavor}
  \caption{Number of products with various number of flavors.}
  \label{fig:num-flav}
\end{subfigure}%
\begin{subfigure}{.5\textwidth}
  \centering
  \widgraph{\linewidth}{\figdir/fig_package_size}
  \caption{Number of products with various package sizes.}
  \label{fig:size}
\end{subfigure}
\caption{Summary of the products.}
\label{fig:test}
\end{figure}

\begin{figure}
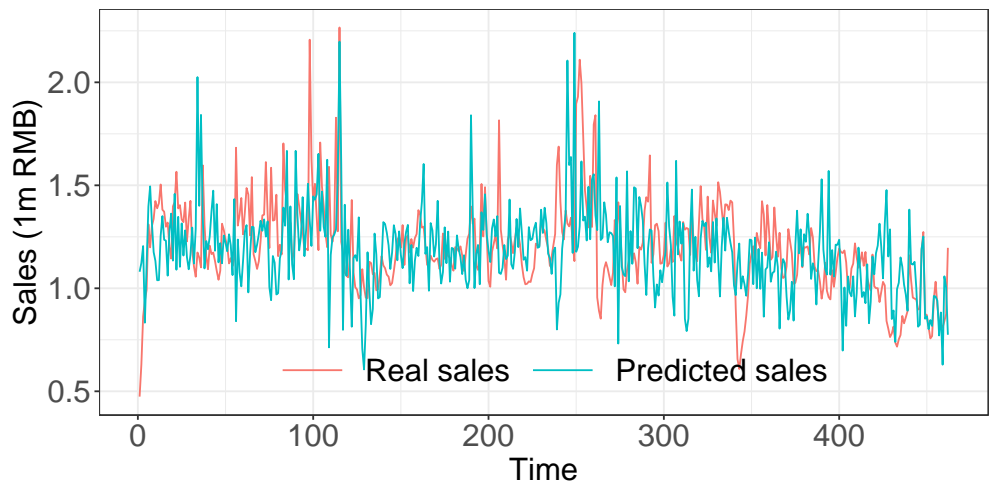

  \centering
  \widgraph{0.8\linewidth}{\figdir/fig_real_vs_predicted_sales_nopromo}
  \caption{Real sales vs predicted sales.}
\label{fig:real-sim-sales}
\end{figure}

To check our model assumption \eqref{eq:reward} on the data,  
\Cref{fig:real-sim-sales} shows the hold-out-sample prediction of the sales
versus the real sales. The predicted sale at each time point $t$ is the
predicted total sales across $L = 31$ locations based on 
$\hTheta_{-t}$ estimated from all the data except for data at time $t$, i.e.,
\begin{align*}
\hat{y}_{t,\ell} = \ba_{t}^\top \widehat{\bTheta}_{-t} \bx_{t,\ell},
\end{align*}
where $\widehat{\bTheta}_{-t} = \arg \min_{\bTheta}\sum_{i=1, i\neq
  t}^T \sum_{\ell=1}^L \left( \ba_i^\top \bTheta \bx_{i,\ell} - r_{i,\ell}
\right )^2 + \lambda \nucnorm{\bTheta}$.  As shown in
\Cref{fig:real-sim-sales}, the real sales and the out-of-sample
predicted sales follow quite closely across time and the out-of-sample
prediction error rate $ \sum_{i=1}^T \sum_{\ell=1}^L (y_{t, \ell} - \hat{y}_{t,\ell})^2/\sum_{i=1}^T \sum_{\ell=1}^L y_{t, \ell}^2 = 0.07$, which indicates that
both our model and estimation are reasonable.


\paragraph{Further simulation results.}

We first detail how we ran the simulation and then provide more
simulation results.  

We run 100 trials, in each of which we set $\tinit = 100$ for the
initialization step and $\lambda_0$ according to \Cref{algo:high-arm} 
to estimate $\hTheta_{\tinit}$; and then at each time
$t = \tinit +1, \ldots, T$, we follow \Cref{algo:high-arm} to
make an assortment-pricing decide $\ba_{t}$ given covariate
$\bx_{t}$.  After determining $\ba_{t}$, we generate the sales $y_{t,\ell}$ 
for $\ell = 1, \ldots, L = 31$ locations according to model
\eqref{eq:reward} based on the pseudo-truth-model with $(\bTheta, \sigma)$.  We
further compare the performance of the assortment-pricing policy with
exploration and without exploration and with different initialization
time $\tinit$. Each setup is simulated 100 times.

\begin{figure}
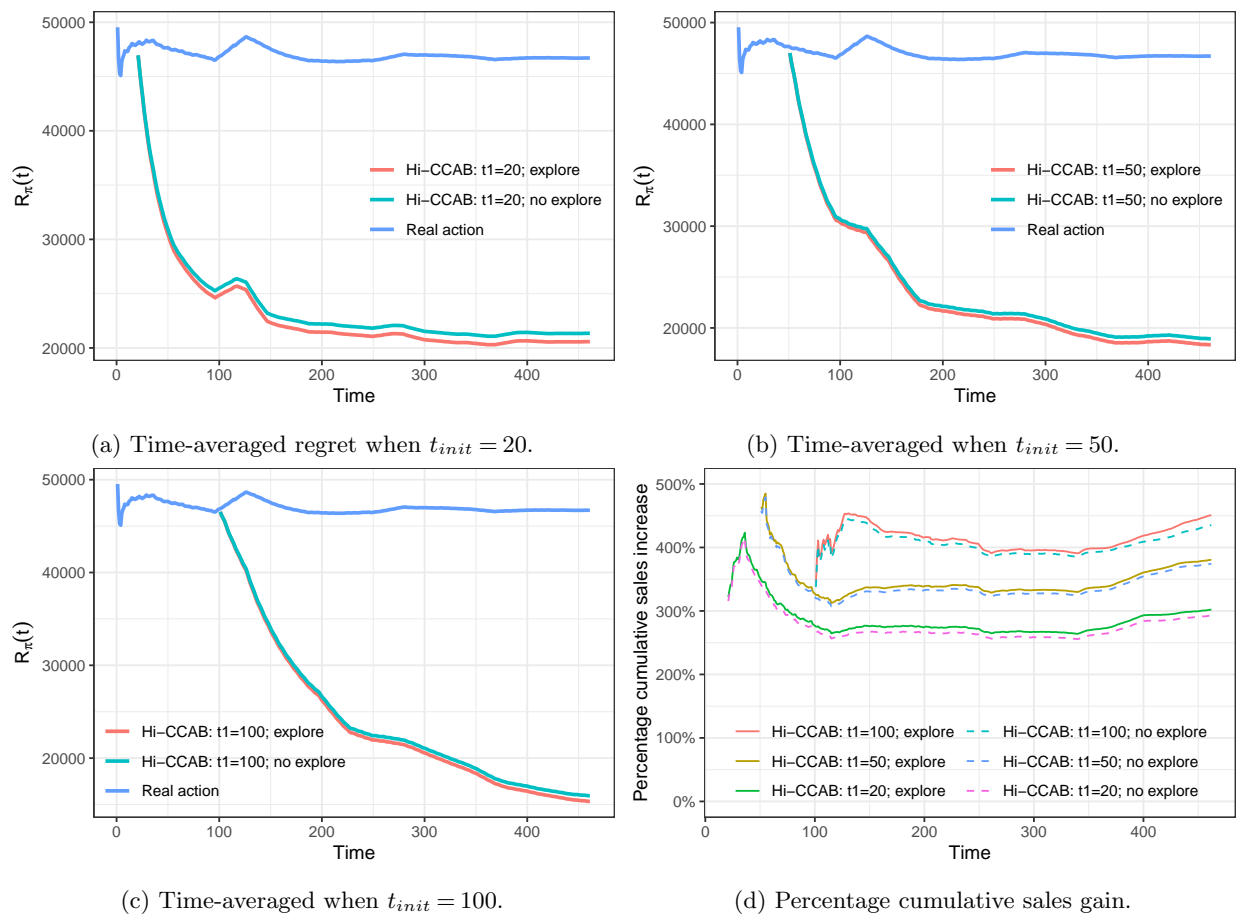

\centering
\begin{subfigure}{.5\textwidth}
  \centering
  \widgraph{\linewidth}{\figdir/fig_time_avg_regret_with_naive_t1_20}
  \caption{Time-averaged regret when $\tinit = 20$.}
  \label{fig:regret-20}
\end{subfigure}%
\begin{subfigure}{.5\textwidth}
  \centering
  \widgraph{\linewidth}{\figdir/fig_time_avg_regret_with_naive_t1_50}
  \caption{Time-averaged when $\tinit = 50$.}
  \label{fig:regret-50}
\end{subfigure}
\begin{subfigure}{.5\textwidth}
  \centering
  \widgraph{\linewidth}{\figdir/fig_time_avg_regret_with_naive}
    \caption{Time-averaged when $\tinit = 100$.}
  \label{fig:regret-100}
\end{subfigure}%
\begin{subfigure}{.5\textwidth}
  \centering
  \widgraph{\linewidth}{\figdir/fig_cum_sales_increase_diff_t0}
  \caption{Percentage cumulative sales gain.}
  \label{fig:sales-increase-diff-t0}
\end{subfigure}
\caption{Performance of \hicoab~with different initialization times
  $\tinit$ and with and without exploration.}
\label{fig:more-sim}
\end{figure}

 Figures \ref{fig:regret-20}-\ref{fig:regret-50} show the time-averaged
 regret and \Cref{fig:sales-increase-diff-t0} show percentage
 gain in cumulative sales when $\tinit = 20, 50, 100$ with exploration
 and without exploration.  \hicoab~with exploration performs better
 then without exploration.  As expected, longer initialization steps
 provide a better initial estimation of the $\bTheta$ and thus helps
 with the performance in a short time windows. As time goes by, all of
 the time-averaged regrets converge to zero and the percentage gain in
 cumulative sales should converge.

\subsection{More details on Case Study II}
\label{app:case-study-ii}

In this section, we provides more background information on Case
Study II and additional numerical results.

\begin{figure}
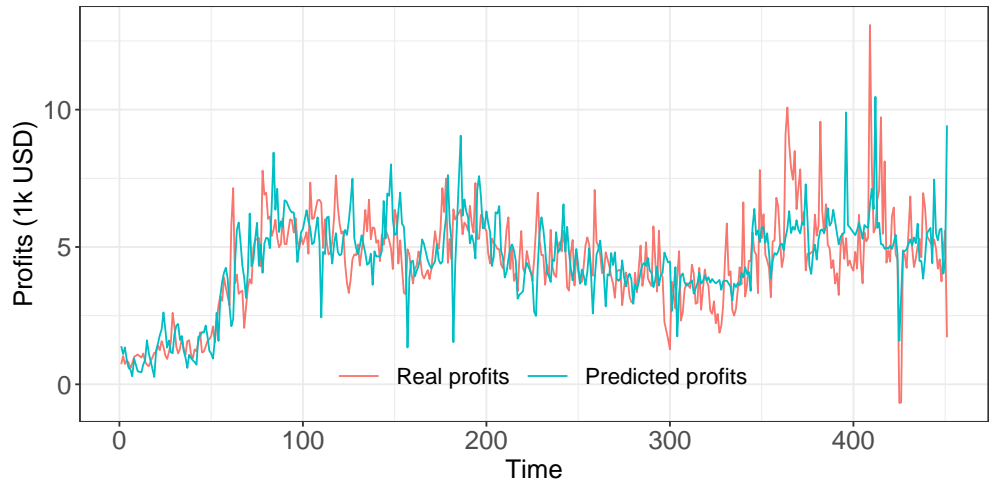

  \centering
  \widgraph{0.8\linewidth}{\figdir/ManiMe/fig_real_vs_predicted_profits}
  \caption{Real profits vs predicted profits.}
\label{fig:manime-profit}
\end{figure}

\Cref{fig:manime-color} shows the total counts of manicures featuring
various colors. Note that one manicure can potentially use multiple
colors.  Red was the most prevalent color, followed by white, gray,
black, blue, yellow, and orange.  The ranking, apart from red, was
determined by sales volumes from previous periods.
\Cref{fig:manime-style} shows the style of the manicures, i.e.,
percentages of designer, glossy, and transparent manicures
respectively.  The count of designer manicures surged in around June
2020 after the total profits increased and then plateaued as shown in
\Cref{fig:manime-profit}.  \Cref{fig:manime-followers} shows the total
number of followers on Instagram of the designers and
\Cref{fig:manime-discount} presents the discount rate, calculated as
the percentage of total daily discount amounts.  Notably, discount
peaks are observed around Thanksgiving, New Year's, and April Fool's
Day.

\begin{figure}
\centering
\begin{subfigure}{.5\textwidth}
  \centering
  \widgraph{\linewidth}{\figdir/ManiMe/fig_color}
  \caption{Counts of manicures with different colors.}
  \label{fig:manime-color}
\end{subfigure}%
\begin{subfigure}{.5\textwidth}
  \centering
  \widgraph{\linewidth}{\figdir/ManiMe/fig_style}
  \caption{Style of the manicures.}
  \label{fig:manime-style}
\end{subfigure}
\begin{subfigure}{.5\textwidth}
  \centering
  \widgraph{\linewidth}{\figdir/ManiMe/fig_followers}
  \caption{Total number of follows of designers.}
  \label{fig:manime-followers}
\end{subfigure}%
\begin{subfigure}{.5\textwidth}
  \centering
  \widgraph{\linewidth}{\figdir/ManiMe/fig_discount}
  \caption{Discount rate.}
  \label{fig:manime-discount}
\end{subfigure}
\caption{Summary of the products.}
\label{fig:test}
\end{figure}

We check our model assumption \eqref{eq:reward} on the data 
similar to Case Study I as detailed in \Cref{app:sim}.
\Cref{fig:real-sim-sales} shows the hold-out-sample prediction of the profits
versus the real profits. 
The real sales and the out-of-sample
predicted sales follow quite closely over time and the out-of-sample prediciton error rate is around 8\%, 
which again indicates that
both our model and estimation are reasonable.

\end{document}